\documentclass{article}

\usepackage{microtype}
\usepackage{graphicx}
\usepackage{subfigure}
\usepackage{booktabs} 

\usepackage{hyperref}

\usepackage[arXiv]{icml2025}

\usepackage{amsmath}
\usepackage{amssymb}
\usepackage{mathtools}
\usepackage{amsthm}

\usepackage[capitalize,noabbrev]{cleveref}

\theoremstyle{plain}
\newtheorem{theorem}{Theorem}[section]

\newtheorem{lemma}[theorem]{Lemma}

\theoremstyle{definition}
\newtheorem{definition}[theorem]{Definition}
\newtheorem{assumption}[theorem]{Assumption}
\theoremstyle{remark}

\usepackage[textsize=tiny]{todonotes}

\icmltitlerunning{Understanding Generalization in Transformers}

\begin{document}
\twocolumn[
\icmltitle{Understanding Generalization in Transformers: Error Bounds and Training Dynamics Under Benign and Harmful Overfitting}



\begin{icmlauthorlist}
\icmlauthor{Yingying Zhang}{sdu}
\icmlauthor{Zhenyu Wu}{bnu}
\icmlauthor{Jian Li}{bnu}
\icmlauthor{Yong Liu}{ruc}
\end{icmlauthorlist}

\icmlaffiliation{sdu}{Zhongtai Securities Institute for Financial Studies, Shandong University}
\icmlaffiliation{bnu}{School of Artificial Intelligence, Beijing Normal University}
\icmlaffiliation{ruc}{Gaoling School of Artificial Intelligence, Renmin University of China}

\icmlcorrespondingauthor{Jian Li}{jli@bnu.edu.cn}

\icmlkeywords{Machine Learning, ICML}

\vskip 0.3in
]



\printAffiliationsAndNotice{}  



\begin{abstract}
Transformers serve as the foundational architecture for many successful large-scale models, demonstrating the ability to overfit the training data while maintaining strong generalization on unseen data, a phenomenon known as benign overfitting. However, research on how the training dynamics influence error bounds within the context of benign overfitting has been limited. This paper addresses this gap by developing a generalization theory for a two-layer transformer with labeled flip noise. Specifically, we present generalization error bounds for both benign and harmful overfitting under varying signal-to-noise ratios (SNR), where the training dynamics are categorized into three distinct stages, each with its corresponding error bounds. Additionally, we conduct extensive experiments to identify key factors that influence test errors in transformers. Our experimental results align closely with the theoretical predictions, validating our findings.
\end{abstract}

\section{Introduction}

In recent years, the phenomenon of benign overfitting \cite{Bartlett_2020} has provided new insights into the over-parameterization of deep neural networks. Traditionally, it was believed that overfitting is more likely when there are more model parameters than training samples, which typically leads to a decrease in the model's generalization ability on new data.
However, modern deep neural networks, despite having numerous parameters, can generalize well, fitting well on training data while maintaining low test error levels. This seemingly contradictory phenomenon has spurred researchers to explore how these models can avoid the pitfalls of overfitting. Despite the fact that deep learning models often have more parameters than training samples and overfit the training data, they can still make accurate predictions on new test data \cite{zhang2017understandingdeeplearningrequires,neyshabur2018understandingroleoverparametrizationgeneralization} . This phenomenon, known as benign overfitting \cite{Bartlett_2020}, has altered our traditional understanding of statistical learning and garnered significant attention from both the statistical and machine learning communities \cite{belkin2018understanddeeplearningneed,Belkin_2019,Belkin_2020,neyshabur2018understandingroleoverparametrizationgeneralization,hastie2020surpriseshighdimensionalridgelesssquares}. This finding has prompted researchers to delve into the potential mechanisms by which these models avoid overfitting.

\textbf{Our contributions.}
In this work, we study the training dynamics, convergence, and generalization for a two-layer transformer with labeled-flipping noise. Our contributions are summarized as bellow:
\begin{itemize}
    \item We develop the generalization theory of a two-layer transformer,  a linear layer behind the attention layer. Under different SNR conditions, we analyze the generalization dynamics benign overfitting and harmful overfitting. 
    \item The training dynamics of benign overfitting are devided into three distinct stages: \textbf{\textit{initialization, signal learning, and convergence}}, while that of harmful overffiting are divided into \textbf{\textit{initialization, noise learning, and divergence}}. We provide the exact generalization error bounds for each stage. 
    \item We take label-flipping noise into consideration, and  remove the orthogonal assumption between the signal patch and the noise patch.  
    \item Our research directly starts from the test loss level, separately calculating the specific test losses for each training phase, making the loss calculation for each phase more precise and reasonable.
    \item We conduct extensive experiments to identify key factors that affact test errors in transformers. 
\end{itemize}

\section{Related Work}
\textbf{The traditional theory of benign overfitting}   
The traditional theory of benign overfitting has obtained a series of research results aimed at understanding this phenomenon in linear, kernel, and random feature models. \cite{Bartlett_2020} set risk bounds for the minimum norm interpolator, stressing overparameterization. \cite{zou2021benignoverfittingconstantstepsizesgd} provided an excess risk bound for constant stepsize SGD. \cite{Liao_2021} extended the analysis to random Fourier feature regression with fixed ratio sample size, data dimension, and number of random features. \cite{adlam2021randommatrixperspectivemixtures} incorporated bias terms into the model. \cite{tsigler2022benignoverfittingridgeregression} extended findings to ridge regression and determined optimal regularization. \cite{mallinar2024benigntemperedcatastrophictaxonomy} found tempered overfitting in certain kernels. \cite{JMLR:v25:22-1389}  extended this to the “multiple random feature model” and discovered multiple descent. 

\textbf{Benign overfitting in transformer}   Benign overfitting in transformers has attracted significant attention in the research field. Some studies utilize the feature learning framework to explore this phenomenon within transformer architectures. \cite{li2024optimizationgeneralizationtwolayertransformers} delved into symbolic gradient descent in two layer transformers, but the research was only concerned with harmful overfitting. \cite{jiang2024unveilbenignoverfittingtransformer} explored the training process of visual transformers. Their work uncovered the training dynamics and clarified the difference between benign and harmful overfitting.  Additionally, there has been new theoretical research on benign overfitting.\cite{jelassi2022visiontransformersprovablylearn,tarzanagh2024transformerssupportvectormachines,tian2023scansnapunderstandingtraining} have focused their studies on the training dynamics of transformers. Analyzing these dynamics helps to accurately describe the optimization process, providing valuable insights into how the model converges and generalizes. \cite{jin2024provableincontextlearningmixture} carried out a theoretical analysis of a Transformer for MoR context learning.\cite{huang2023incontextconvergencetransformers}  explored the training of single layer transformers. \cite{li2024nonlineartransformerslearngeneralize}  examined transformers with nonlinear self-attention. \cite{frei2024trainedtransformerclassifiersgeneralize} investigated the performance of linear Transformers in random linear classification tasks and found that benign overfitting can occur under certain conditions.  

\section{Problem Setup}
In this section, we denote the data generation model, two-layer transformer model, and the gradient descent-based training algorithm. 

\textbf{Notions.} We define two sequences $\{a_n\}$ and $\{b_n\}$, which have the following relationship. We say $a_n = O(b_n)$ and $b_n=\Omega(a_n)$if there exist $|a_n|\leq c_1|b_n|$ for some positive constant $c_1$. At the same time, we define $a_n=\Theta(b_n)$ if $a_n = O(b_n)$ and $a_n=\Omega(b_n)$ hold.
\begin{definition}[Data Generation Model]
    \label{def.Data Generation Model}
     Let $\boldsymbol{\mu}_+, \boldsymbol{\mu}_- \in \mathbb{R}^d$ be fixed vectors which represent the signals contained in each data point $(\mathbf{X}, y)$, where $\|\boldsymbol{\mu}_+\|_2=\|\boldsymbol{\mu}_-\|_2 = \|\boldsymbol{\mu}\|_2$ and $\langle\boldsymbol{\mu}_+, \boldsymbol{\mu}_-\rangle = 0$. Then we define each data point $(\mathbf{X}, y)$ with  the input features $X = (\boldsymbol{x}_{1}, \boldsymbol{x}_{2})^T\in\mathbb{R}^{2 d}$, and $y \in \{ \pm 1\}$ is generated from the following model:
\begin{itemize}
    \item The true data label $\hat{y}\in\{\pm1\}$ is generated as a Rademacher random variable. The variable satisfies $\mathbb{P}[\hat{y}=1]=\mathbb{P}[\hat{y} = - 1]=\frac{1}{2}$. The observed label $y$ is subsequently generated by flipping $\hat{y}$ with a probability $\alpha$, where $\alpha$ satisfies the condition $\alpha\in[0,1/2)$. i.e. $\mathbb{P}[y = \hat{y}]=1 - \alpha$ and  $\mathbb{P}[y=-\hat{y}]=\alpha$. 
    \item We set $\boldsymbol{x}_1$= $y \cdot \boldsymbol{\mu}$, which represents the signal. Note that, when $y = 1$, $\boldsymbol{x}_1$ is set to $\boldsymbol{\mu}_+$, and  $\boldsymbol{x}_1$ is set to $\boldsymbol{\mu}_-$ when $y=-1$, where $\boldsymbol{\mu}_+$ and $\boldsymbol{\mu}_-$ both represent signals.
    \item We set $x_2$=$\boldsymbol{\xi}$, which represents noise and satisfies $\boldsymbol{\xi} \sim \mathcal{N}(0,\sigma_{p}^{2} \boldsymbol{I}_d)$.
\end{itemize}
\end{definition}
We consider each data point as a vector of two tokens, \(\mathbf{X} = (\boldsymbol{x}_1, \boldsymbol{x}_2)^T \in \mathbb{R}^{2d}\). The first token, \(\boldsymbol{x}_1\), represents the component of the signal that is intrinsically linked to the training data label and the signals \(\boldsymbol{\mu}_+\) and \(\boldsymbol{\mu}_-\), while the second token, \(\boldsymbol{x}_2\), serves as noise and is irrelevant to both the label and the signal. Building on Definition \ref{def.Data Generation Model} from \cite{jiang2024unveilbenignoverfittingtransformer}, we further refine the data distribution to enhance its practical applicability. Specifically, we introduce label-flipping noise to the true label \(\hat{y}\) and relax the orthogonality condition between the signal vector \(\boldsymbol{\mu}\) and the noise vectors \(\boldsymbol{\xi}\) (for comparison, see Definition 3.1 in \cite{jiang2024unveilbenignoverfittingtransformer}). Unlike the original approach, our data distribution is not restricted to image data in the context of vision transformers (ViT), and thus it is applicable to all transformer models.

\textbf{Signal-to-Noise Ratio (SNR):} From \cite{cao2022benignoverfittingtwolayerconvolutional}, when the dimension \(d\) is large, the norm of the noise vector satisfies \(\lVert \boldsymbol{\xi} \rVert_2 \approx \sigma_p \sqrt{d}\) based on standard concentration bounds. Therefore, the signal-to-noise ratio (SNR) can be expressed as \(\rm{SNR} = \frac{\lVert \boldsymbol{\mu} \rVert_2}{\sigma_p \sqrt{d}}\), which is approximately equal to \(\frac{\lVert \boldsymbol{\mu} \rVert_2}{\lVert \boldsymbol{\xi} \rVert_2}\). Hence, we use the expression \(\rm{SNR} = \frac{\lVert \boldsymbol{\mu} \rVert_2}{\sigma_p \sqrt{d}}\) to represent the signal-to-noise ratio.

\textbf{Two-layer Transformers:} We denote a two-layer transformer model. Specifically, the first layer is a self-attention layer with a softmax activations, and the second layer is a linear layer, where $\mathcal{S}$ represents the softmax function. The two-layer transformer model is defined as:
\begin{equation*}
f(\mathbf{X},\upsilon) =\upsilon^T\mathbf{X}^T \mathcal{S}(\mathbf{X}\mathbf{W}_Q\mathbf{W}_K\mathbf{X}^T)\mathbf{X}\mathbf{W}_V.
\end{equation*}
The parameters of the linear layer is denoted as $\upsilon \in \mathbb{R}^{m_v}$. Note that, the linear layer is fixed in \cite{jiang2024unveilbenignoverfittingtransformer}, while in this paper $\upsilon$ is learnable. The parameters of the attention layer are defined as $\mathbf{W}_Q, \mathbf{W}_K, \mathbf{W}_{V,j}$, where $\mathbf{W}_Q, \mathbf{W}_K \in \mathbb{R}^{m_k\times 2d}$ and $\mathbf{W}_{V,j} \in \mathbb{R}^{m_v\times d}$ for $j \in \{\pm 1\}$, which represents the query matrix, the key matrix, and the value matrix respectively. 
We use \(\theta\) to represent all the weights of the attention  in the model, which is defined as \(\theta = (\mathbf{W}_Q, \mathbf{W}_K, \mathbf{W}_{V,j}\)). We overwrite the model in specific form for $j \in \{\pm 1\}$:
\begin{equation}
\begin{aligned}
f(\theta, \mathbf{X}, \upsilon) = \frac{1}{m_v} \sum_{r \in [m_v]} \Big( & \upsilon^T \mathbf{x}_{1} (S_{11} + S_{21}) \mathbf{W}_{Vj,r}+ \\
& \upsilon^T \mathbf{x}_{2} (S_{12}+S_{22}) \mathbf{W}_{Vj,r} 
\Big).\label{eq:formula1}
\end{aligned}
\end{equation}
In the above formula, we divide the output of the softmax function into four types of vectors, which are the softmax outputs of the pairwise inner products of the query signal, query noise, key signal, and key noise. Specifically, signal-to-signal output $S_{11}$, signal-to-noise output $S_{12}$, noise-to-signal output $S_{21}$,   and noise-to-noise output $S_{22}$ have been defined in the appendix. For example, signal-to-signal output $S_{11}$ can be written as:
\begin{equation*}
\begin{aligned}
S_{11} &= \text{Softmax}(\langle q_{\pm}^{(t)}, k_{\pm}^{(t)} \rangle) \\
&= 
\begin{cases}
\frac{\exp(\langle q_{+}, k_{+} \rangle)}{\exp(\langle q_{+}, k_{+} \rangle) + \exp(\langle q_{+}, k_{\xi,i} \rangle)} & \text{for } i \in [S_+], \\
\frac{\exp(\langle q_{-}, k_{-} \rangle)}{\exp(\langle q_{-}, k_{-} \rangle) + \exp(\langle q_{-}, k_{\xi,i} \rangle)} & \text{for } i \in [S_-].
\end{cases}
\end{aligned}
\end{equation*}
$S_+$ is the set of $i$ in $[N]$ with $y_i = 1$ and $S_-$ is the set of $i$ in $[N]$ with $y_i=-1$. 
Note that,
$q_{+}$, $k_{+}$, $q_{-}$, $k_{-}$, $k_{\xi,i}$ are related to the query with $+1$ label, the key with $+1$ label, the query with $-1$ label, the key with $-1$ label, and the key with noise, respectively.
 
\textbf{Training Algorithm:}
We use a training dataset generated from the distribution \(D\) in Definition \ref{def.Data Generation Model} called \(S := \{(\mathbf{X}_i, y_i)\}_{i = 1}^N\), where \(i\) is the number of samples in the training data. Our transformer model is trained through the minimization of the empirical cross-entropy loss function.
\begin{equation*}
L_S(\theta) = \frac{1}{N} \sum_{i=1}^{N} \ell(y_i f(\theta, \mathbf{X}, \upsilon)),
\end{equation*}
where \(\ell(z) = \log(1 + \exp(-z))\). We also defined test loss \(L_{\mathcal{D}}(\theta) := \mathbb{E}_{(X, y) \sim {\mathcal{D}}} \ell(y f(\theta, \mathbf{X}, \upsilon))\).

 We train the transformer model starting from random Gaussian initialization \(W_Q, W_K \sim \mathcal{N}(0, \sigma_k^2)\), and \(W_V\) is sampled from \(\mathcal{N}(0, \sigma_v^2)\) at initialization. We use GD optimization to minimize training loss \(L_S(\theta)\), and have the following update rules:
\begin{equation*}
\begin{aligned}
\mathbf{W}_V^{(t + 1)} &= \mathbf{W}_{V}^{(t)} - \eta \, \left(\nabla_{\mathbf{W}_{V}} L_S(\mathbf{W}^{(t)})\right), \\
\mathbf{W}_{Q}^{(t + 1)} &= \mathbf{W}_{Q}^{(t)} - \eta \, \left(\nabla_{\mathbf{W}_{Q}} L_S(\mathbf{W}^{(t)})\right), \\
\mathbf{W}_{K}^{(t + 1)} &= \mathbf{W}_{K}^{(t)} - \eta \, \left(\nabla_{\mathbf{W}_{K}} L_S(\mathbf{W}^{(t)})\right),
\end{aligned}
\end{equation*}
 We write the detailed gradient formulas and update rules  in Appendix D.
\section{Main Results}
In this section, we present our main theoretical findings. These finds are based on several key conditions: the dimensionality \(d\), the sample size \(N\), the variances of the Gaussian initialization \(\sigma_h^2\) and \(\sigma_V^2\), the weights of the linear layer \(\upsilon\), the learning rate \(\eta\), and the noise rate \(p\). Collectively, these factors determine the convergence and generalization capabilities of the label-flipping transformer model. The specific conditions are as follows.

\begin{assumption}
    \label{Definition.4.1} 
    Assume that there exists a sufficiently small failure probability $\delta > 0$, a large constant $C$, and a certain training loss $\epsilon > 0$ such that the following hold:
    \begin{enumerate}
        \item \label{Definition:1} The dimension $d$ satisfies
        \[
        d\geq
        \begin{cases}
            \text{SNR}^{4}N^{4}\epsilon^{-4}, & \text{if } \|\mathbf{\mu}\|\geq\sigma_p\sqrt{d},\\[0.5em]
            \text{SNR}^{-4}N^{4}\epsilon^{-4}, & \text{if } \|\mathbf{\mu}\|<\sigma_p\sqrt{d}.
        \end{cases}
        \]
        \item \label{Definition:2} The training sample size $N$ satisfies $N \geq C\cdot\text{polylog}(d)$.
        \item \label{Definition:3} The noise rate $\alpha$ satisfies $\alpha\in[0, 1/2)$.
        \item \label{Definition:4} The linear layer weights satisfy $\left\|\upsilon\right\|_2 = \Theta(1)$.
        \item \label{Definition:5} The learning rate $\eta$ satisfies:
        \[
        \eta \leq\min\left\{\sigma_p^{2}d, \|\boldsymbol{\mu}\|_2^{2}\right\}N^{2}\epsilon^{-2}.
        \]
        \item \label{Definition:6} The standard deviation of Gaussian initialization $\sigma_v$ satisfies:
        \[
        \sigma_v\leq
        \begin{cases}
            \frac{\sigma_p\sqrt{d}\epsilon}{\|\upsilon\|N}, & \text{if } \|\mathbf{\mu}\|\geq\sigma_p\sqrt{d},\\[0.5em]
            \frac{\|\mathbf{\mu}\|\epsilon}{\|\upsilon\|N}, & \text{if } \|\mathbf{\mu}\|<\sigma_p\sqrt{d}.
        \end{cases}
        \]
        \item \label{Definition:7} The variance of Gaussian initialization $\sigma_k^{2}$ satisfies:
        \[
        \sigma_k^{2}\leq\min\left\{\sigma_p^{2}d, \|\boldsymbol{\mu}\|_2^{2}\right\}N^{2}\epsilon^{-2}.
        \]
    \end{enumerate}
\end{assumption}

The first two assumptions in Assumption \ref{Definition.4.1} are to ensure that the learning model is in the over-parameterized setting. The assumption \ref{Definition:3} is to ensure that we do not add too much noise to avoid seriously affecting the learning results of the transformer. The assumption \ref{Definition:4} is intended to simplify the calculation. The parameter $\sigma_v$ bounded setting in the assumption \ref{Definition:6} is used to study the effect on error under benign overfitting conditions. The assumptions \ref{Definition:5} and \ref{Definition:7} on $\eta$ and $\sigma_k^2$ are to ensure that gradient descent can effectively minimize the training loss.

\begin{theorem}[\textbf{Benign overfitting in transformers}]
    \label{thm:4.1}
    When \(N \cdot \text{SNR}^2 = \Omega(1)\), for any \(\epsilon > 0\), under Assumption \ref{Definition.4.1}, with probability at least \(1-\delta\): 
\begin{itemize}
    \item (\textbf{Phase 1: Initialization}) There exists \(T_1 = \frac{140N}{\eta d^4(N\|\mu\|_2^2 - 120^2C_p^2\sigma_p^2d)\|\upsilon\|_2^2}\), and for \(t\in(0,T_1]\), the test error is:
    \(L_{\mathcal{D}}(\mathbf{W}^{(t)})\leq\alpha + O(1)\).
    \item (\textbf{Phase 2: Signal learning}) There exists \(T_2=\Theta\left(\frac{1}{\eta\|\mu\|_2^2\|\upsilon\|_2^2}\right)\), for \(t\in(T_1,T_2]\), the test error is:
    \begin{align*}
    &L_{\mathcal{D}}(\mathbf{W}^{(t)})\leq\alpha + \\
    &\exp\left(\frac{c_4}{2\pi}-c_{10}\eta^4\|\mu\|_2^8(t - T_1)^2(t - T_1 - 1)^2\text{SNR}^2\right).
    \end{align*}
    \item (\textbf{Phase 3: Convergence}) There exists \(T_3=\Theta\left(n^{-1}\epsilon^{-1}\|\mu\|_2^{-2}\|\upsilon\|_2^{-2}\right)\), and for \(t\in(T_2,T_3]\):
        \begin{itemize}
            \item The training loss converges to \(\epsilon\): \(L_S(\mathbf{W}^{(t)})\leq\epsilon\).
            \item The upper bound of the test error is close to the noise rate \(\alpha\):
        \end{itemize}
    \end{itemize}
    \begin{align*}
        L_{\mathcal{D}}(\mathbf{W}^{(t)})\leq\alpha + 
        \exp\left(\frac{c_{12}}{2\pi}-\frac{c_{14}\eta^4(t - T_2)^4\|\boldsymbol{\mu}\|_2^6\cdot\text{SNR}^2}{2\sigma_v^2}\right).
    \end{align*}
\end{theorem}

Theorem \ref{thm:4.1} analyzes the scenario in which a transformer satisfies the condition for benign overfitting, i.e., a large SNR where \( N \cdot \text{SNR}^2 = \Omega(1) \). This condition suggests that the model training dynamics can be divided into three stages:
\begin{enumerate}
    \item \textbf{Initialization stage}: At the beginning, when all model parameters are initialized according to Assumption \ref{Definition.4.1}, the transformer parameters have not been sufficiently trained, resulting in large test errors. During this stage, the test loss is influenced by the label-flipping probability $\alpha$ and a large constant.
    \item \textbf{Signal learning stage}: In this stage, the attention mechanism increasingly focuses on the signal, leading to a reduction in test errors. The test loss is proportional to the number of iterations \( t \), the learning rate \( \eta \), and \( \text{SNR}^2 \). At the beginning of this stage (\( t = T_1 \)), the test loss drops to a value similar to the initialization stage, given by 
   $
   L_{\mathcal{D}}(\mathbf{W}^{(t)}) \leq \alpha + \exp\left(\frac{c_{4}}{2\pi}\right).`
   $
    \item \textbf{Convergence stage}: During the final stage, the training loss converges to a small error \( \epsilon \), while the upper bound of the test error converges exponentially in terms of key factors. Specifically, the test loss is proportional to the number of iterations \( t \), the learning rate \( \eta \), the signal strength \( \|\boldsymbol{\mu}\| \), and \( \text{SNR}^2 \), but it is also inversely proportional to the initialization variance \( \sigma_v^2 \).
    At the beginning of this stage, the test loss is upper bounded by    
    $
   L_{\mathcal{D}}(\mathbf{W}^{(t)}) \leq \alpha + \exp\left(\frac{c_{12}}{2\pi}\right).`
   $
   Note that, the constant $\frac{c_{12}}{2\pi}$ when $t=T_2$ is much smaller than $\frac{c_{4}}{2\pi}$ when $t=T_1$ due to the signal learning in the second stage.
\end{enumerate}

\begin{theorem}[\textbf{Harmful overfitting in transformers}]
    \label{thm:4.2}
    When \(N^{-1} \cdot \text{SNR}^{-2} = \Omega(1)\),  for any \(\epsilon > 0\), under Assumption \ref{Definition.4.1}, with probability at least \(1-\delta\) :
    \begin{itemize}
        \item (\textbf{Phase 1: Initialization}) There exists \(T_1 = \frac{56N}{\eta {\frac{1}{4}}\sigma_p^2d\lVert\upsilon\rVert_2^2}\), for \(t\in(0,T_1]\), such that the test error is:
        \[
        L_{\mathcal{D}}(\mathbf{W}^{(t)}) \leq \alpha + O(1).
        \]
        \item (\textbf{Phase 2: Noise learning}) There exists \(T_2 = \Theta\left(\frac{N}{\eta\sigma_p^2d\lVert\upsilon\rVert_2^2\log(6N^2M^2/\delta)}\right)\). For \(t\in(T_1,T_2]\), the test error is:
        \begin{align*}
&L_D(\mathbf{W}^{(t)})\leq (\frac{1}{2} - \alpha)\left(O\left(\frac{d^{\frac{1}{2}}(\log(N^2/\delta))^3}{m_v\|\mu\|_2^2\|v\|_2^2}\right)\right.\\
&\left.+O\left(\eta d^{\frac{1}{2}}(\log(N^2/\delta))^3(t - T_1)\left(\frac{1}{m_v\|\mu\|_2^2N\|v\|_2^2}+\frac{1}{N}\right)\right)\right.\\
&\left.+O\left(\frac{\eta d\|\ v\|_2^2(t - T_1)}{N}\right)\right)
   \end{align*}
        \item (\textbf{Phase 3: Divergence}) There exists \(T_3 = \Theta\left(\frac{N}{\eta\epsilon\sigma_p^2d\lVert\upsilon\rVert_2^2}\right)\), \(t\in(T_2,T_3]\) such that:
        \begin{itemize}
            \item The training loss converges to \(\epsilon\): \(L_S(\mathbf{W}^{(t)}) \leq \epsilon\).
            \item The test loss is higher:
        \end{itemize}
        \begin{equation*}
            L_{\mathcal{D}}(\mathbf{W}^{(t)}) \geq (1 - \alpha) \log\left(\frac{1 + 2e^{\frac{1}{2}}}{1 + e^{\frac{1}{2}}}\right) \\
            +\alpha\log\left(2 + e^{-\frac{1}{2}}\right).
        \end{equation*}
    \end{itemize} 
\end{theorem}

Theorem \ref{thm:4.2} describes the training dynamics under harmful overfitting, where the signal-to-noise ratio (SNR) is small, satisfying \( \text{SNR} \leq O(1/\sqrt{N}) \). The training process is divided into three stages: 

\begin{itemize}
    \item \textbf{Initialization Stage}: The upper bound on the test loss in this stage is similar to that in the initialization stage of benign overfitting.
    \item \textbf{Noise Learning Stage}: As training progresses, the upper bound on the test loss increases with respect to the number of iterations \( t \), since the transformer model starts learning the noise instead of the signal. The longer the training, the greater the accumulation of errors.
    \item \textbf{Divergence Stage}: Once the model begins fitting the noise, the test error diverges and is larger than a large constant. The lower bound of the test loss exceeds $0.483$ when \( \alpha = 0 \) and $0.721$ when \( \alpha = 1/2 \).
\end{itemize}

\subsection{Comparison with Prior Works}
Compared with \cite{kou2023benignoverfittingtwolayerrelu,cao2022benignoverfittingtwolayerconvolutional,shang2024initializationmattersbenignoverfitting}, we extend the feature learning framework from CNN to the transformer and provide the test error for each stage.
Compared with \cite{magen2024benignoverfittingsingleheadattention}, who investigated benign overfitting in single head attention models without label noise, our setup is more sophisticated. We incorporate a linear layer weighted by $\upsilon$ after the attention layer and relax the assumption of orthogonality between noise and signal.
In contrast to \cite{li2024nonlineartransformerslearngeneralize}, who studied the training dynamics of two-layer transformers using the softmax activation function and sign gradient descent, our findings are more comprehensive. We address both benign and harmful overfitting, adjust the linear layer differently, and provide conditions for the phase transition between the two types of overfitting.
When compared to \cite{jiang2024unveilbenignoverfittingtransformer}, who explored the training dynamics of two-layer transformers without considering label noise, our approach is more practical. We take label-flipping noise into account and calculate the test losses for each stage of training more precisely.

\section{Proof Techniques}
\label{submission}
In this section, we present the main proof techniques for studying the training dynamics and the specific test loss in both benign and harmful overfitting scenarios. The complete proofs of Theorem \ref{thm:4.1} and Theorem \ref{thm:4.2} are provided in the appendix.

\subsection{Key Technique 1: Distinct Test Errors between Benign and Harmful Overfitting}
We are aware that due to the existence of label-flipping noise, the Bayes optimal test error is at least \(\alpha\). So the gap between the test error and the training error is at least \(\alpha\). This circumstance precludes us from applying commonly used standard bounds based on uniform convergence \cite{bartlett2017spectrallynormalizedmarginboundsneural,neyshabur2018understandingroleoverparametrizationgeneralization} or stability \cite{zhang2017understandingdeeplearningrequires,mou2017generalizationboundssgldnonconvex,chen2018stabilityconvergencetradeoffiterative}.

To simplify the theoretical analysis, we use two forms with distinct test errors. In the analysis of benign overfitting, we will conduct an algorithm-dependent test error analysis. Firstly, we can decompose the test error as follows:
\begin{equation}
    \begin{aligned}
        L_{\mathcal{D}}(\mathbf{W}^{(t)})
        = &\mathbb{P}(y\neq \text{sign}(f(\theta, \mathbf{X}, \upsilon)))\\
        \leq & \alpha+\mathbb{P}(\widehat{y}f(\theta, \mathbf{X}, \upsilon)\leq 0).
    \end{aligned}
    \label{eq:formula2}
\end{equation}
Using \eqref{eq:formula2}, the analysis of test error can be reduced to bounding the wrong prediction probability
$\mathbb{P}(\widehat{y}f(\theta, \mathbf{x}, \upsilon)\leq 0)$. 

In the analysis with  harmful overfitting, we use the ordinary representation  and the full expectation formula
\[
\mathbb{E}[Z]=\mathbb{E}[Z|A]P(A)+\mathbb{E}[Z|\overline{A}]P(\overline{A}).
\]
which can split the test loss into two parts: label not flipped and label flipped respectively, so the test loss is as follows:
\begin{equation}
    \begin{aligned}
        L_{\mathcal{D}}(\mathbf{W}^{(t)})&=\mathbb{E}\ell[yf(\theta, \mathbf{X}, \upsilon)]\\
        &=(1 - \alpha)\mathbb{E}_{y\in\widehat{y}}\ell(yf(\theta, \mathbf{X}, \upsilon))\\
        & \quad + \alpha\mathbb{E}_{y\in - \widehat{y}}\ell(-yf(\theta, \mathbf{X}, \upsilon)).
    \end{aligned}
    \label{eq:formula3}
\end{equation}
\subsection{Key Technique 2: Splitting $V_{+}^{(t)}, V_{-}^{(t)}$ and $V_{\xi }^{(t)}$}
{\small
\[
\begin{aligned}
    &\widehat{y}f(\theta,\mathbf{x},\upsilon)\\
    =&\frac{1}{m_v}\sum_{r\in[m_v]}(\upsilon^Tx_1(S_{11} + S_{21})W_{V,r}^{(t)}+\upsilon^Tx_2(S_{12}+S_{22})W_{V,r}^{(t)})\\
    =&\sum_{j}\sum_{r\in[m_v]}[(S_{11}+S_{21})\langle W_{Vj,r}^{(t)},x_1\rangle \upsilon+(S_{12}+S_{22})\langle W_{Vj,r}^{(t)},x_2\rangle \upsilon]\\
    =&\sum_{j,r}[(S_{11}+S_{21})(\mu_+^TW_{vj,r}^{(t)}\upsilon+\mu_-^TW_{Vj,r}^{(t)}\upsilon) +(S_{12}+S_{22})\xi^TW_{Vj}^{(t)}\upsilon]\\
    =&\sum_{j,r}[(S_{11}+S_{21})(V_{+}^{(t)}+V_{-}^{(t)})+(S_{12}+S_{22})V_{\xi }^{(t)}].
\end{aligned}
\]
}
Because it involves label flipping, the model output value can be divided into real label $\widehat{y}$ and flipped label $-\widehat{y}$. We consider:
\begin{definition}[Splitting of V vector]
    \label{def:Splitting_of_V_vector}
    \begin{align*}
        V_{+\widehat{y}}^{(t)}&=\sum_{r} a_r V_+^{(t)},\\
        V_{+(-\widehat{y})}^{(t)}&=\sum_{r} b_r V_+^{(t)},\\
        V_{+}^{(t)}&=V_{+\widehat{y}}^{(t)} + V_{+(-\widehat{y})}^{(t)}.
    \end{align*}
\end{definition}

Similarly, we split $V_{-}^{(t)}$ and $V_{\xi}^{(t)}$ for $j = \pm\widehat{y}$. 
\begin{lemma}[The test loss of benign overfitting]
    \label{lem_the_test_loss_benign_overfitting}
    When analyzing the second  phase of benign overfitting, by substituting Key Technique 2 into the calculation of Key Technique 1, we have 
    \begin{equation*}
    \begin{aligned}
        &P(\widehat{y}f(\theta,\mathbf{x},\upsilon) \leq 0) \\
        =&P\Bigg(\sum_{r}(S_{11} + S_{21})(V_{+,\widehat{y}}^{(t)} +  V_{-,\widehat{y}}^{(t)})+ (S_{12} + S_{22}) V_{\xi,\widehat{y}}^{(t)} \\
        \leq &\sum_{r} \Big((S_{11} + S_{21})(V_{+}^{(t)} + V_{-}^{(t)}) + (S_{12} + S_{22}) V_{\xi,(-\widehat{y})}^{(t)} \Big)\Bigg)\\
        \leqslant & P\left(\sum_{r}\frac{S_{11}+S_{21}}{S_{12}+S_{22}}\left(V^{(t)}_{+,\widehat{y}}+V^{(t)}_{-,\widehat{y}}\right)\leq V^{(t)}_{\xi,(-\widehat{y})}+O(1)\right).
    \end{aligned}
\end{equation*}
\end{lemma}

\subsection{Key Technique 3: Update Rules of $\mathbf{W}_Q$, $\mathbf{W}_K$,$\mathbf{W}_V$}
Another key problem we encountered in theoretical analysis was handling label-flipping noise. Numerous cases in practical applications demonstrate that certain deep learning models can still exhibit excellent prediction performance when dealing with datasets contaminated by label noise. This phenomenon starkly contradicts traditional machine learning theory. To effectively learn from data that contain both correct signals and flip noise, the model is compelled to fit the noise induced by label flips. Undoubtedly, this will have a negative impact on the model's generalization ability on new data. Moreover, label-flip noise is highly likely to steer the learning algorithm towards an incorrect optimization direction, making it arduous to capture valuable signals within the data.

Now, let's delve into the update strategy of parameters $\rho_{V,+}^{(t)}$, $\rho_{V,-}^{(t)}$, $\rho_{V,\xi,i}^{(t)}$.According to \cite{jiang2024unveilbenignoverfittingtransformer}, we can get the following two lemmas. In the following two lemmas, we denote $S_+ := \{i \in [N]: y_i = 1\}$, $S_- := \{i \in [N]: y_i = -1\}$ in order to analyze the dynamic changes of training internal mechanism under label flipping.
\renewcommand{\thelemma}{5.4} 
\begin{lemma}[ Update rules of $\mathbf{W}_V$]
\label{lem_update_rules_V}
Let $W_V^{(t)}$ represent the $V$ matrix of the Transformer during the $t$-th iteration of gradient descent. The coefficients $\rho_{V,+}^{(t)}$, $\rho_{V,-}^{(t)}$, $\rho_{V,\xi,i}^{(t)}$ specified in this lemma adhere to the following iterative equations:
\begin{align*}
\rho_{V,+}^{(0)} &= 0, \quad
\rho_{V,-}^{(0)} = 0,\quad
\rho_{V,\xi,i}^{(0)} = 0,\\
V_+^{(t)}&=\mu_+^{\top}W_V^{(t)}\upsilon=\mu_+^{\top}W_V^{(0)}\upsilon + \rho_{V,+}^{(t)}\|\upsilon\|_2^2,\\
V_-^{(t)}&=\mu_-^{\top}W_V^{(t)}\upsilon=\mu_-^{\top}W_V^{(0)}\upsilon + \rho_{V,-}^{(t)}\|\upsilon\|_2^2,\\
V_{\xi,i}^{(t)}&=\xi_{i}^{\top}W_V^{(t)}\upsilon=\xi_{i}^{\top}W_V^{(0)}\upsilon + \rho_{V,\xi,i}^{(t)}\|\upsilon\|_2^2
\end{align*}
for  $i \in \{1,\ldots,N\}$.
\end{lemma}
\section{Experiments}

In this section, the theoretical results are verified by a series of experiments, and the relevant experimental design and code implementation refer to the literature \cite{jiang2024unveilbenignoverfittingtransformer}. The data sets and models are constructed in strict accordance with Definition \ref{def.Data Generation Model}. The specific design is as follows:
\begin{enumerate}
 \item Dataset: We generate the training set and the test set according to the theoretical definition. Specifically, each data point is divided into two parts (M=2): signal and noise. The signal is generated from two mutually orthogonal vectors,  $\|\boldsymbol{\mu}\|_2 \cdot \mathbf{e}_1$ and $\|\boldsymbol{\mu}\|_2 \cdot \mathbf{e}_2$., with equal probability, where \( \mathbf{e}_1 = [1, 0, \dots, 0]^\top \) and \( \mathbf{e}_2 = [0, 1, \dots, 0]^\top \). The noise is generated from Gaussian noise \( \mathcal{N}(0, \sigma_p^2 \mathbf{I}) \).
   The data size \( N \) is not fixed in our experiment.  
   In Sections 6.1, 6.2, and 6.3, to investigate the impact of \( N \) variation on test loss, we vary \( N \) from 2 to 20. In the remaining experiments, to allow the model to fully learn the data, we fix \( N \) at 100.
   For the composition parameters of the signal-to-noise ratio (\( \text{SNR
   } \)), we vary the signal \( \mu \) from 1 to 100, while fixing the noise variance at 4, to explore the effect of \( \text{SNR} \) on benign overfitting.
\item Model: We define the model as a simple two-layer transformer model used to explore the relevant factors that affect benign overfitting, consisting of an attention layer and multiple layers of perceptrons. The dimensions of the weight matrix are all 512. Initialize parameters using PyTorch's default initialization method. We set the target loss to 0.01 and conducted experiments using the full batch descent method.
\end{enumerate}
 All experimental results were the average of 20 repeated experiments. And all experiments were conducted on NVIDIA \( \text{A100} \)  GPU. 

\subsection{Three Stages of Training Process}

In this section, we conduct an in-depth analysis of the benign and harmful overfitting behavior of the loss function, attention weight, and $\mathbf{W}_V$  during training, and study the influence of different label inversion probabilities $\alpha$ on them.
Label flipping is applied to the signal with reversal probabilities \(\alpha \in \{0.2,0.1, 0.01, 0.001\}\), and the model is trained to examine the changes in training loss, test loss, signal attention weight (atten signal), noise attention weight (atten noise), as well as the projections of the $\mathbf{W}_V$  onto the signal (V signal) and onto the noise (V noise) throughout the training process. Previous studies have identified three distinct stages in this process, and we conduct experiments to carefully analyze the changes in each stage. The specific results are presented in the following figure.

\begin{figure}[t]
    \centering
    \textbf{Three stage analysis of losses}
    \par\vspace{0.5em} 
    \subfigure[Benign overfitting]{
       \includegraphics[width=0.46\linewidth]{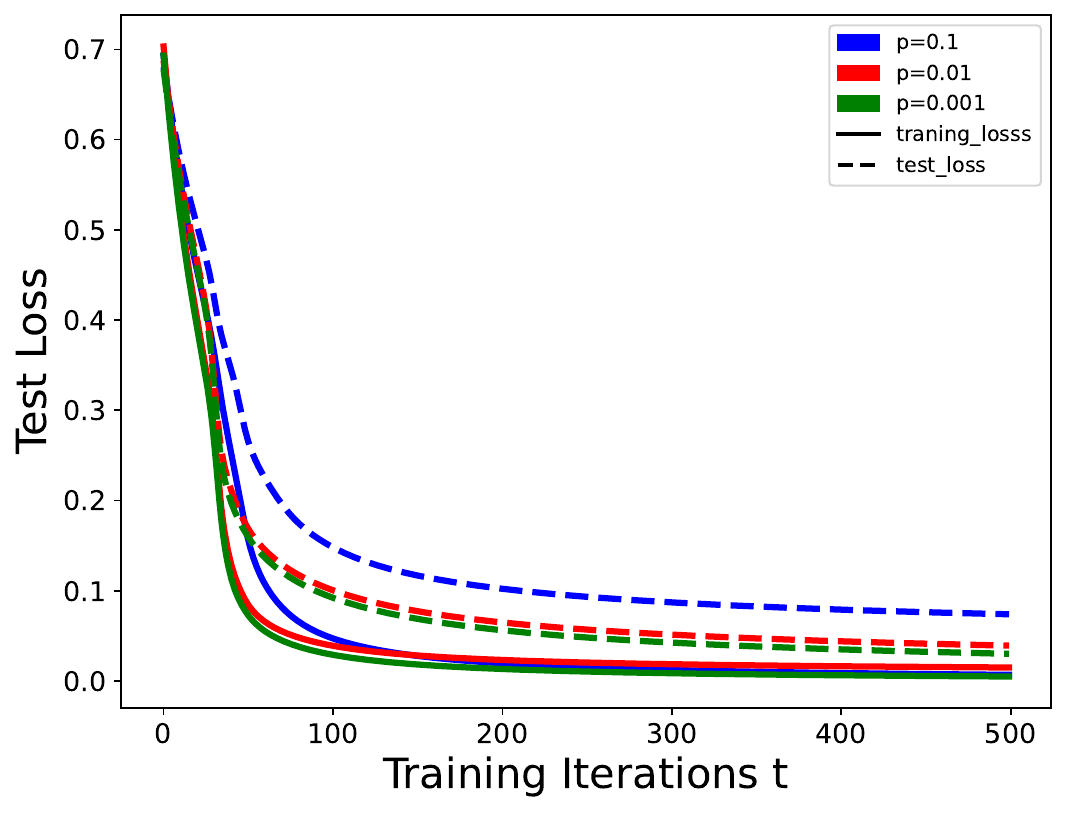}
    }
    \hfill
    \subfigure[Harmful overfitting]{
       \includegraphics[width=0.46\linewidth]{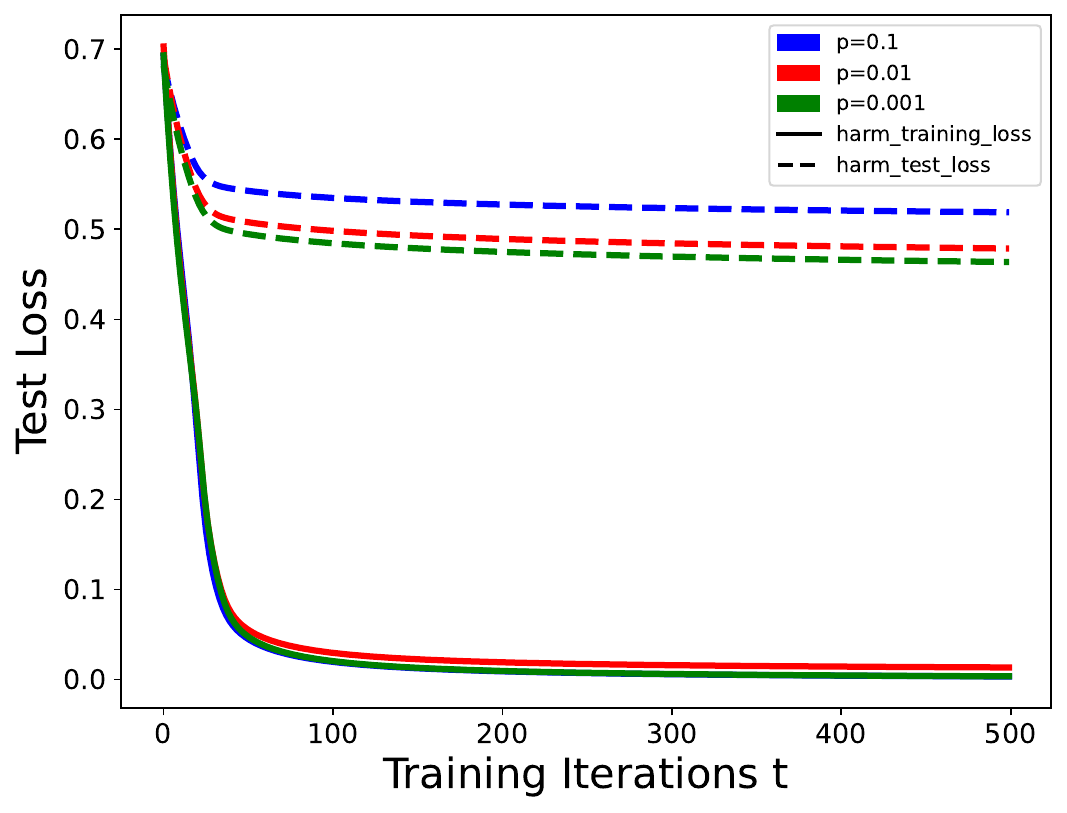}
    }
    \par\vspace{1em} 

    \textbf{Three-stage analysis of atten }
    \par\vspace{0.5em} 
    \subfigure[Benign overfitting]{
       \includegraphics[width=0.46\linewidth]{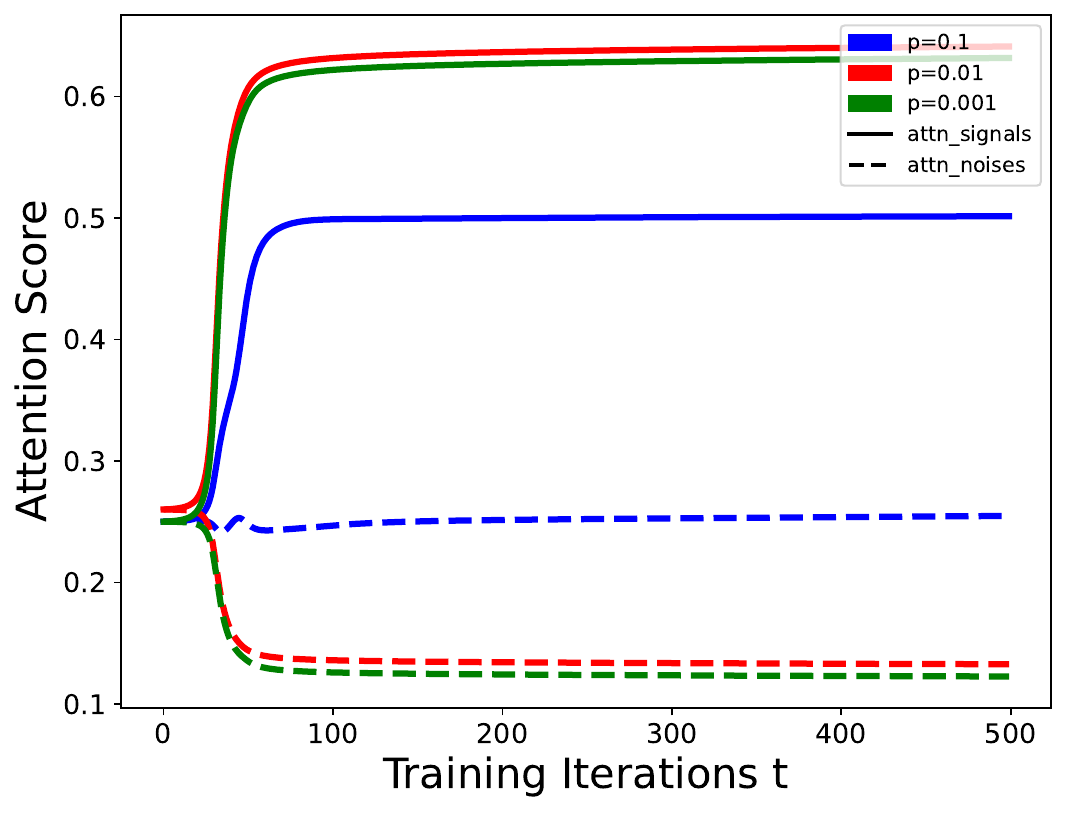}
    }
    \hfill
    \subfigure[Harmful overfitting]{
       \includegraphics[width=0.46\linewidth]{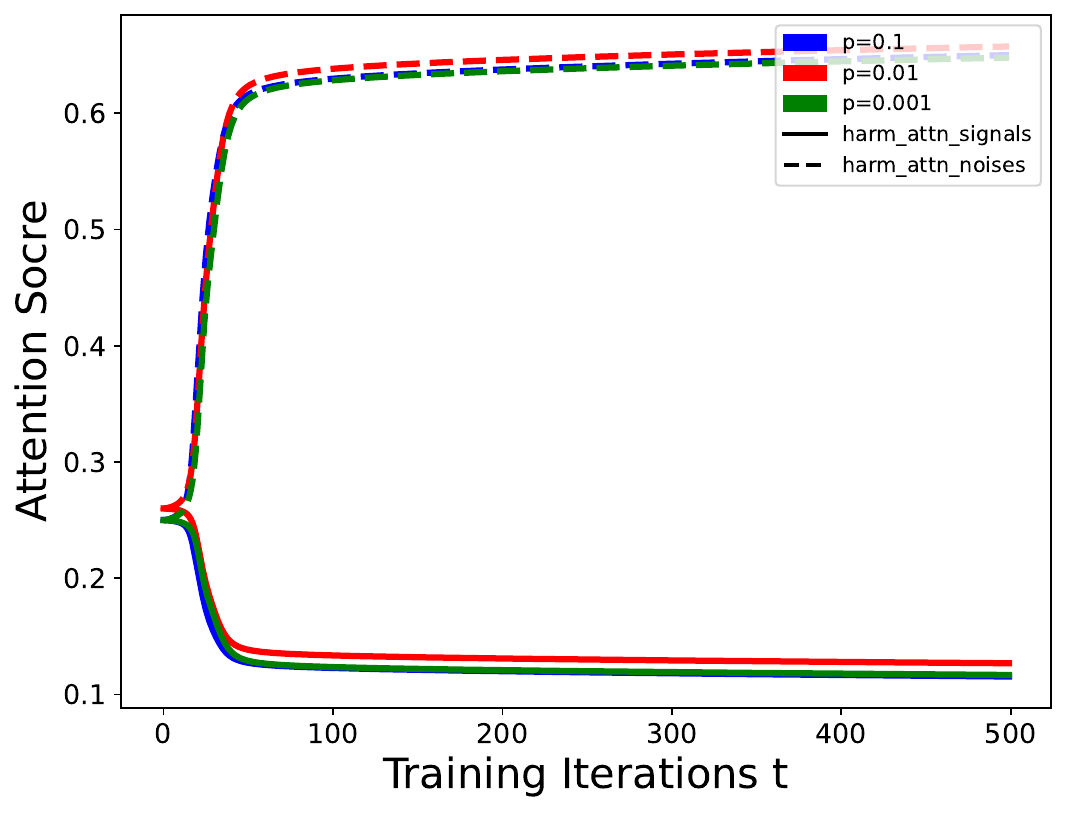}
    }
    \par\vspace{1em} 

    \textbf{Three stage analysis of weight v matrix}
    \subfigure[Benign overfitting]{
       \includegraphics[width=0.46\linewidth]{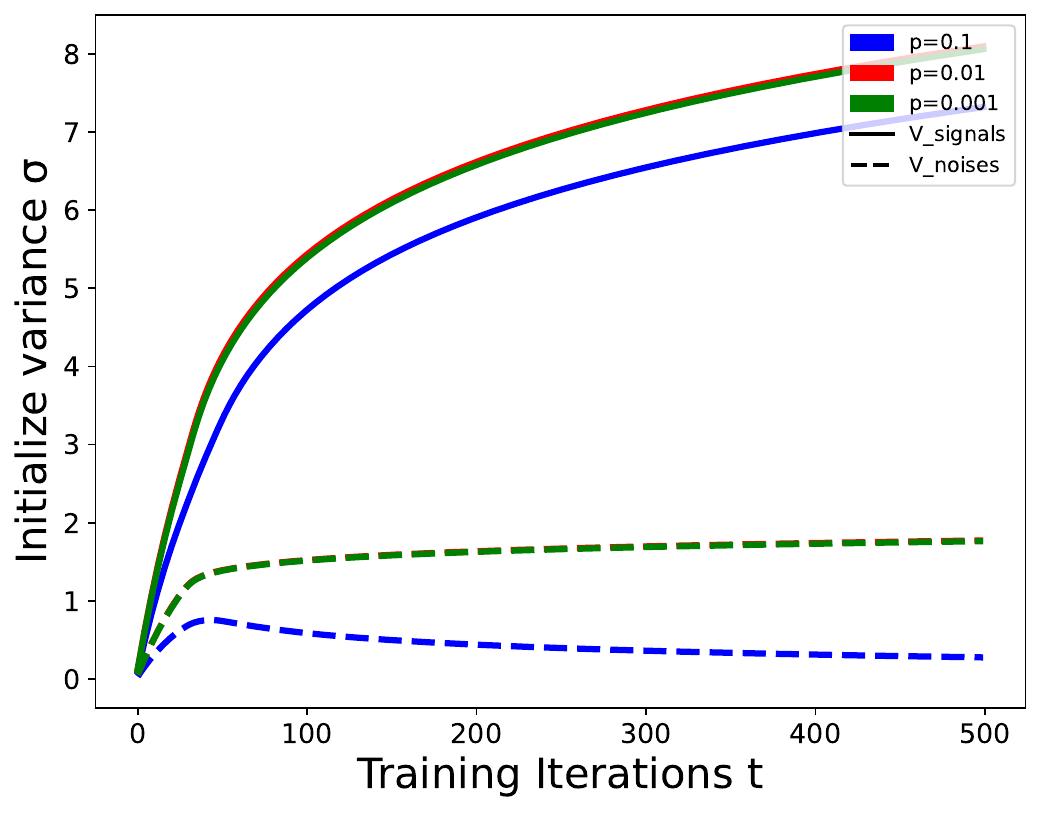}
    }
    \hfill
    \subfigure[Harmful overfitting]{
       \includegraphics[width=0.46\linewidth]{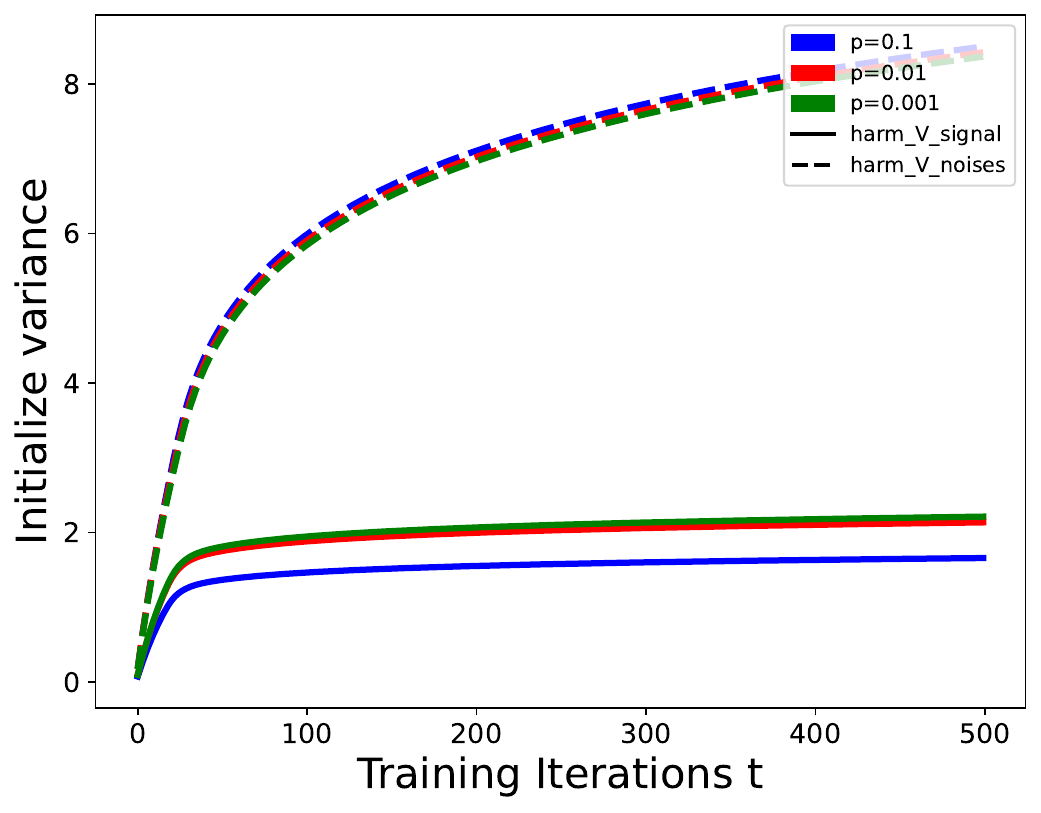}
    }

    \caption{Training stage analysis of benign overfitting and harmful overfitting under label flipping conditions (Experimental Design Reference \cite{jiang2024unveilbenignoverfittingtransformer}), (a)(b): Test loss and training loss vary over time; (c)(d): Signal atten and noise atten vary over time; (e)(f): Signal V and noise V vary over time. }
    \label{Three_stage}
\end{figure}
\cref{Three_stage} (a) and (b) show the training and test losses of Transformers under benign and harmful overfitting. Initially, loss reduction is minimal but quickly transitions to a second stage where test loss drops significantly before stabilizing. Higher label-flipping probabilities lead to increased final test loss due to added noise interfering with signal discrimination.

\cref{Three_stage} (c) and (d) illustrate attention dynamics for signals and noise. In benign overfitting, early-stage attention is undifferentiated but quickly shifts toward signals, improving discrimination. Eventually, attention stabilizes, allowing the model to distinguish signals from noise. However, at a label-flipping probability of 0.1, attention to signals weakens, reducing discriminative ability.

In harmful overfitting, the model initially struggles to distinguish signals from noise but gradually shifts focus toward noise, leading to poor learning performance. In both cases, excessive noise attention degrades the model's ability to learn meaningful patterns.

\subsection{Phase Transition between Benign Overfitting and Harmful Overfitting}

In this section, we will explore the overall impact of SNR and N on test loss. To further investigate the boundary of benign overfitting, we will add different label flips to the data set. The specific results are shown in the following figure:

\begin{figure}[t]
    \centering
    \subfigure[\( \alpha = 0.001 \)]{
       \includegraphics[width=0.46\linewidth]{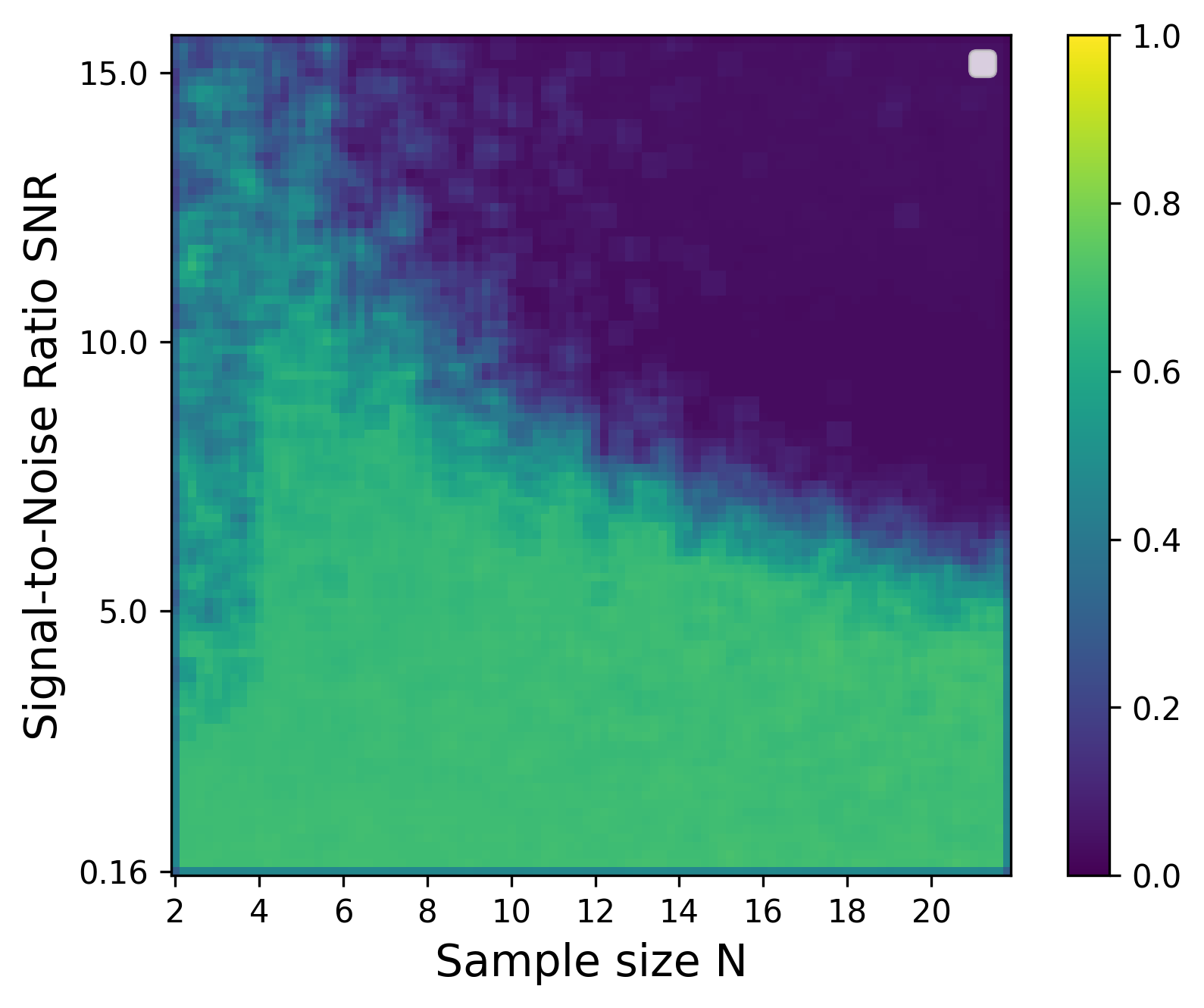}
    }
    \hfill
    \subfigure[\( \alpha = 0.01 \)]{
       \includegraphics[width=0.46\linewidth]{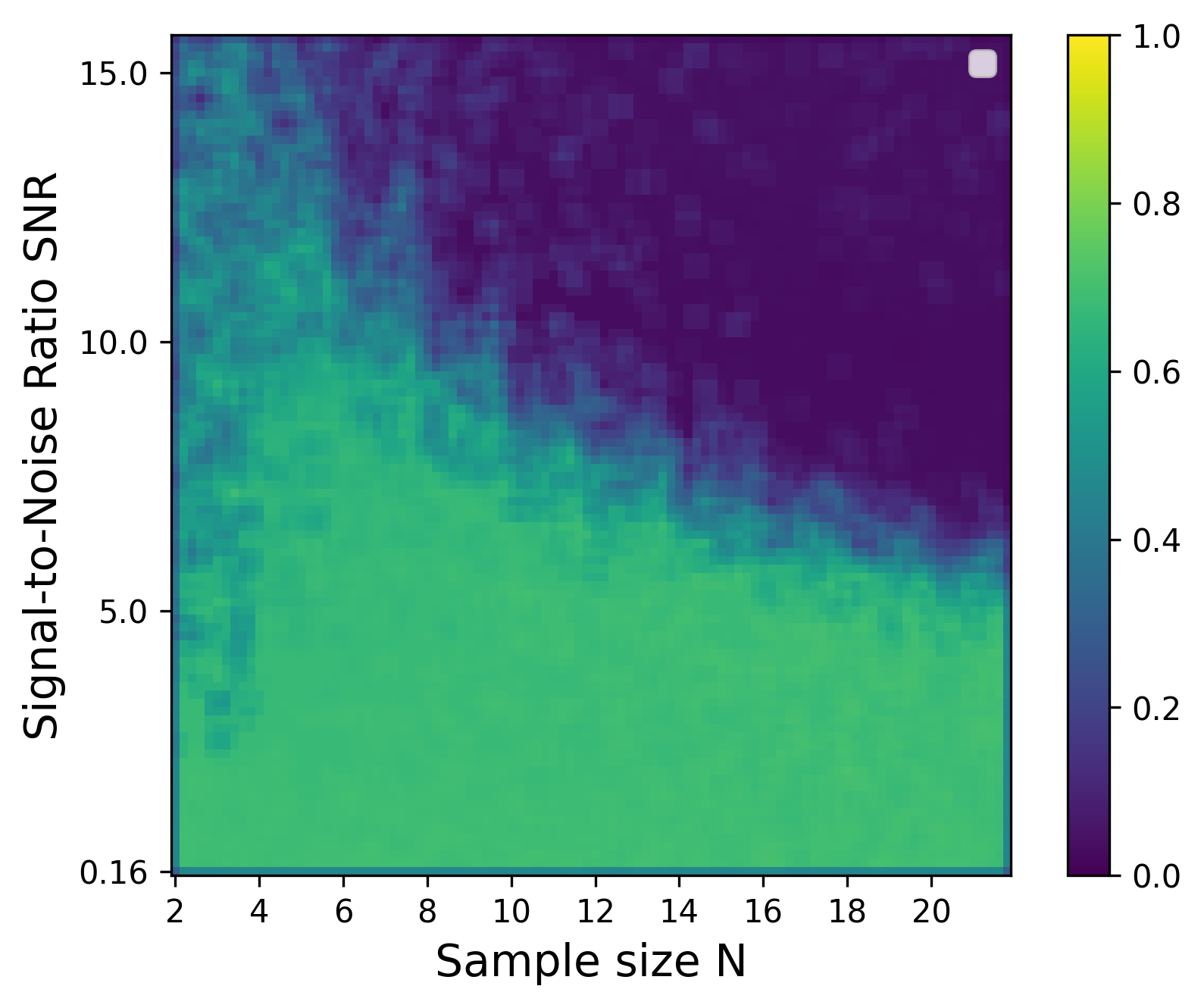}
    }
    \par\vspace{1em} 
    \subfigure[\( \alpha = 0.1 \)]{
       \includegraphics[width=0.46\linewidth]{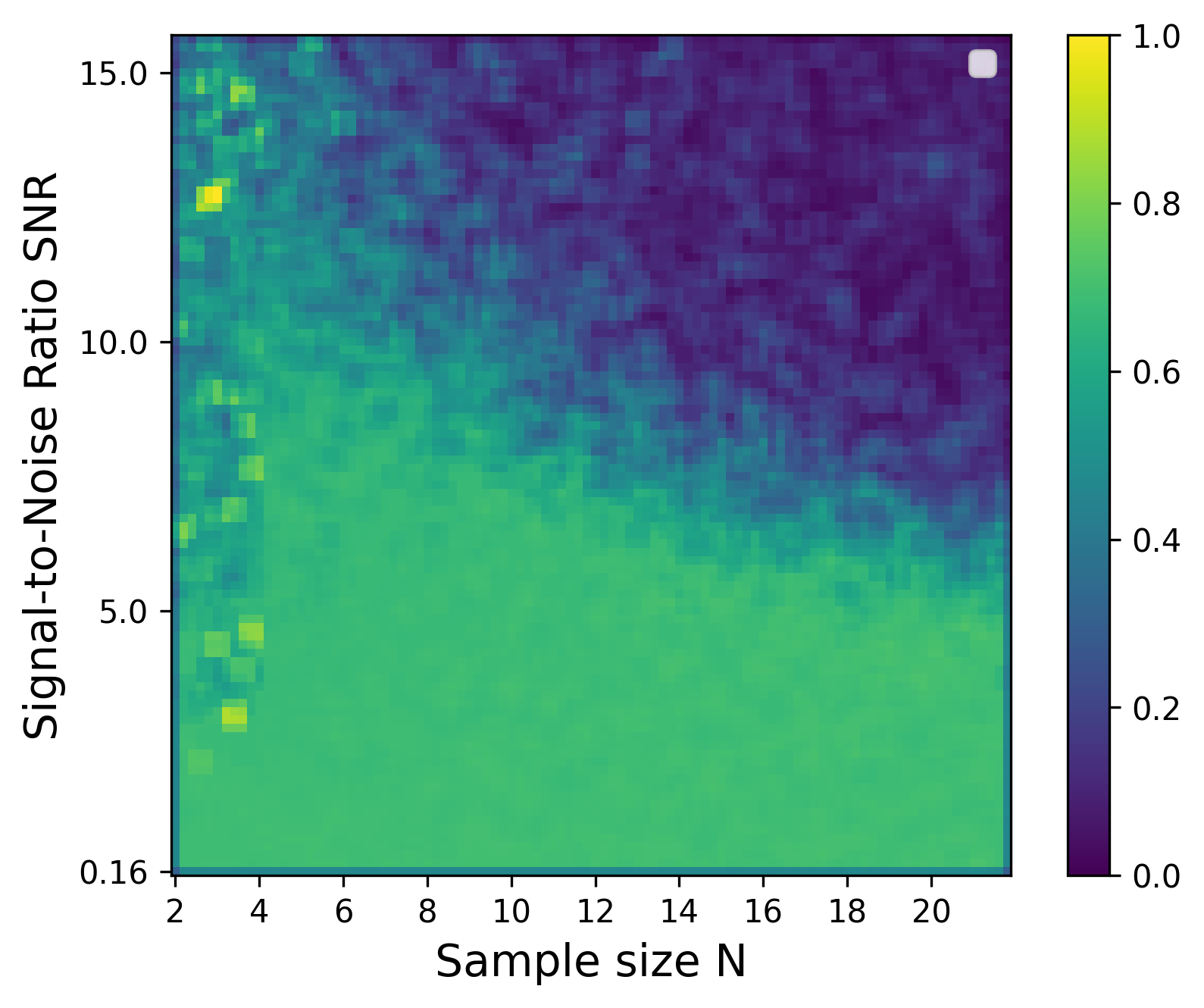}
    }
    \hfill
    \subfigure[\( \alpha = 0.2 \)]{
       \includegraphics[width=0.46\linewidth]{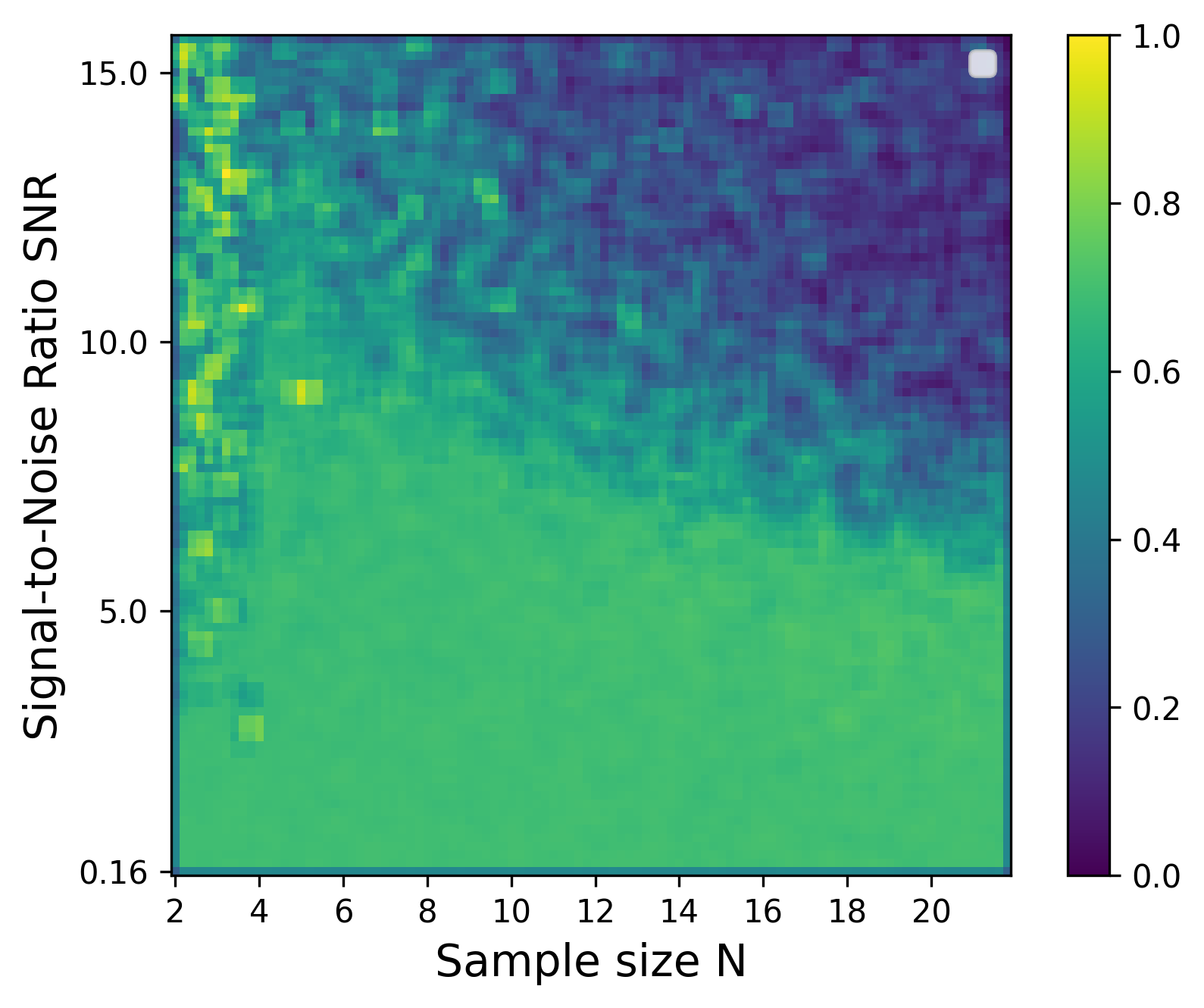}
    }
    \caption{
    The test loss w.r.t. different label-flipping probability $\alpha$. Figure (a) represents the heat map drawn at a label flipping probability of \( \alpha = 0.001 \), Figure (b) represents the heat map drawn at a label flipping probability of \( \alpha = 0.01 \), Figure (c) represents the heat map drawn at a label flipping probability of \( \alpha = 0.1 \), and Figure (d) represents the heat map drawn at a label flipping probability of \( \alpha = 0.2 \)(Experimental design originated from \cite{jiang2024unveilbenignoverfittingtransformer}). 
    }
    \label{heatmap}
\end{figure}

\begin{figure}[t]
    \centering
    \subfigure[\( \alpha = 0.001 \)]{
       \includegraphics[width=0.46\linewidth]{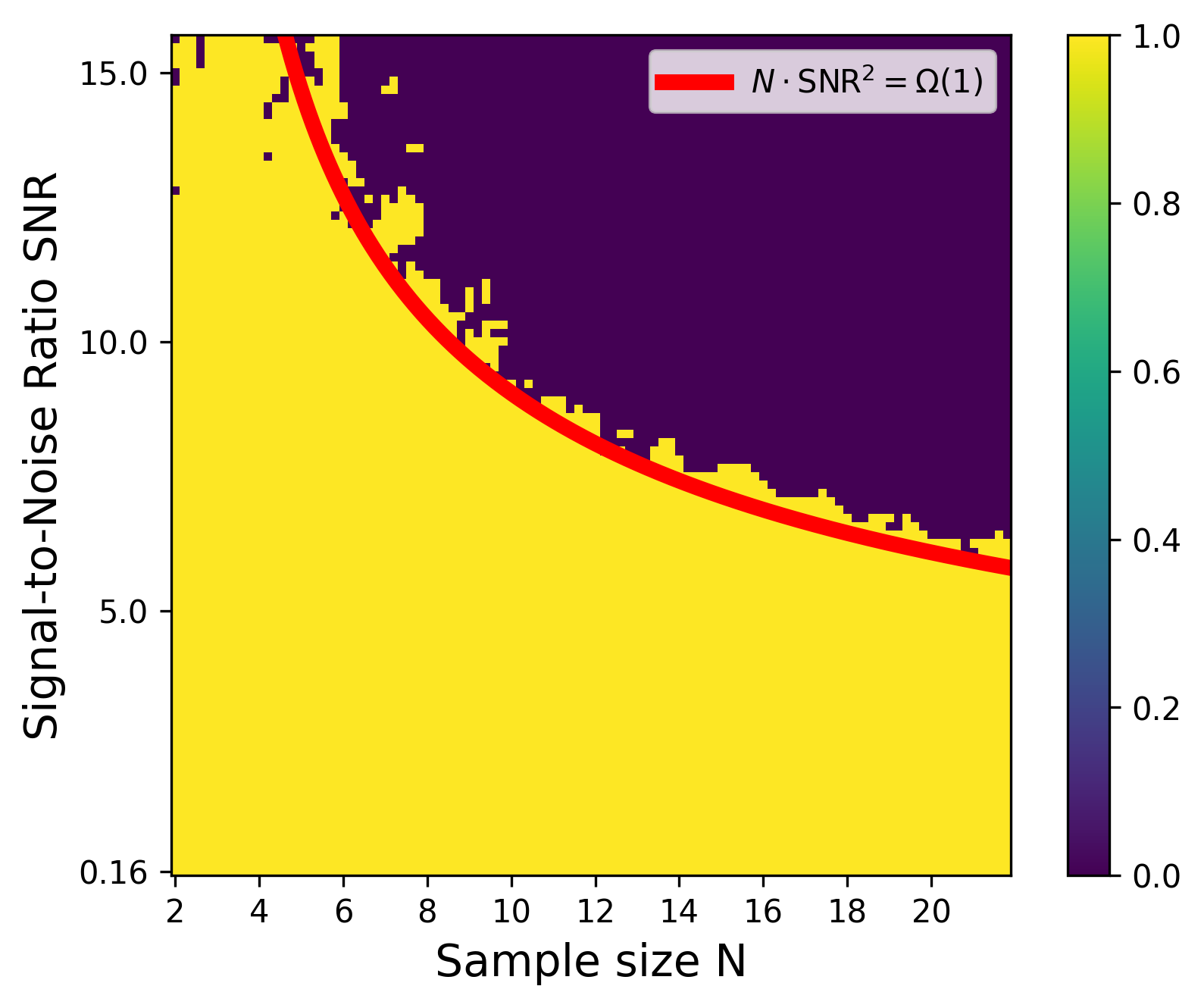}
    }
    \hfill
    \subfigure[\( \alpha = 0.01 \)]{
       \includegraphics[width=0.46\linewidth]{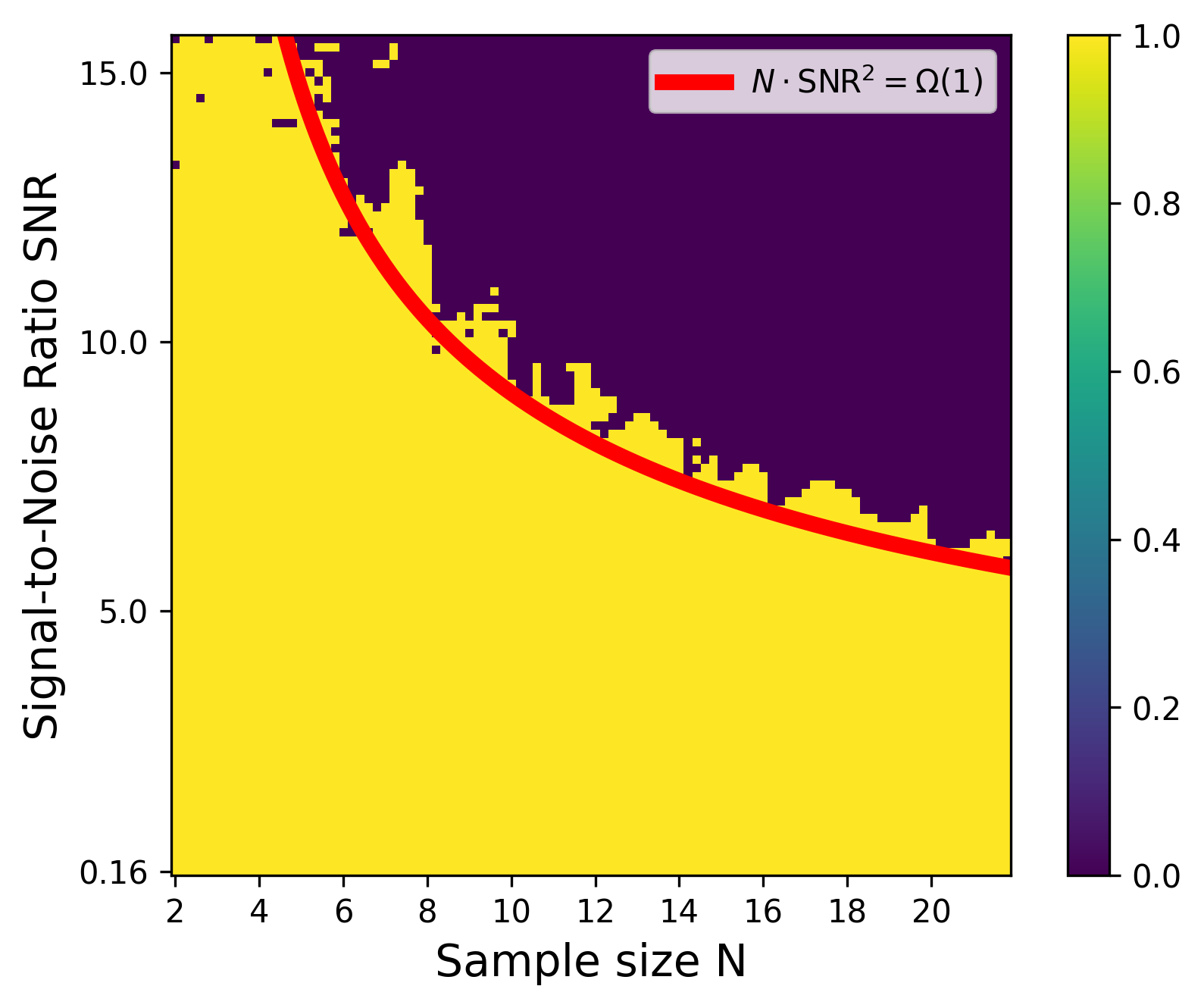}
    }
    \par\vspace{1em} 
    \subfigure[\( \alpha = 0.1 \)]{
       \includegraphics[width=0.46\linewidth]{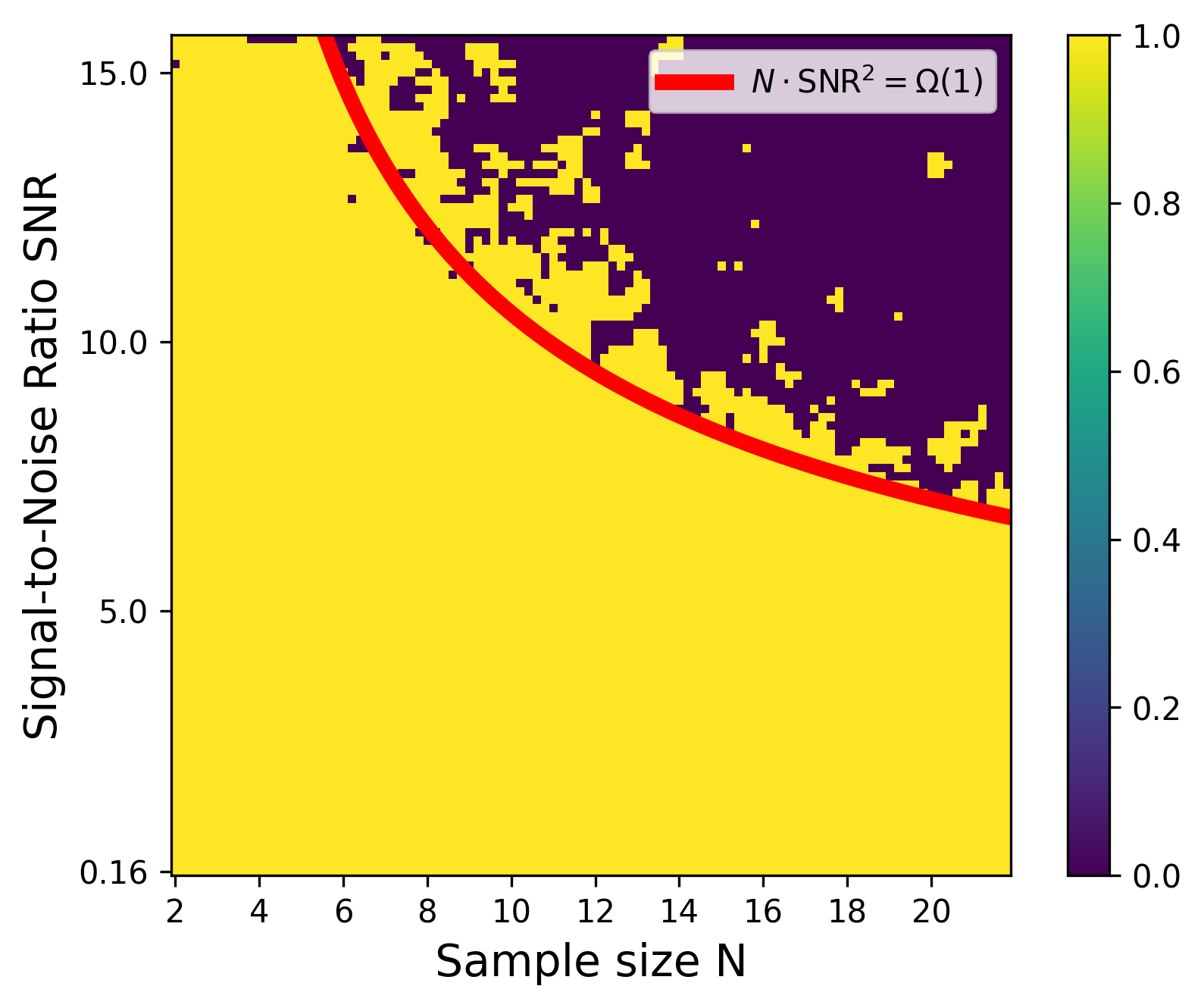}
    }
    \hfill
    \subfigure[\( \alpha = 0.2 \)]{
       \includegraphics[width=0.46\linewidth]{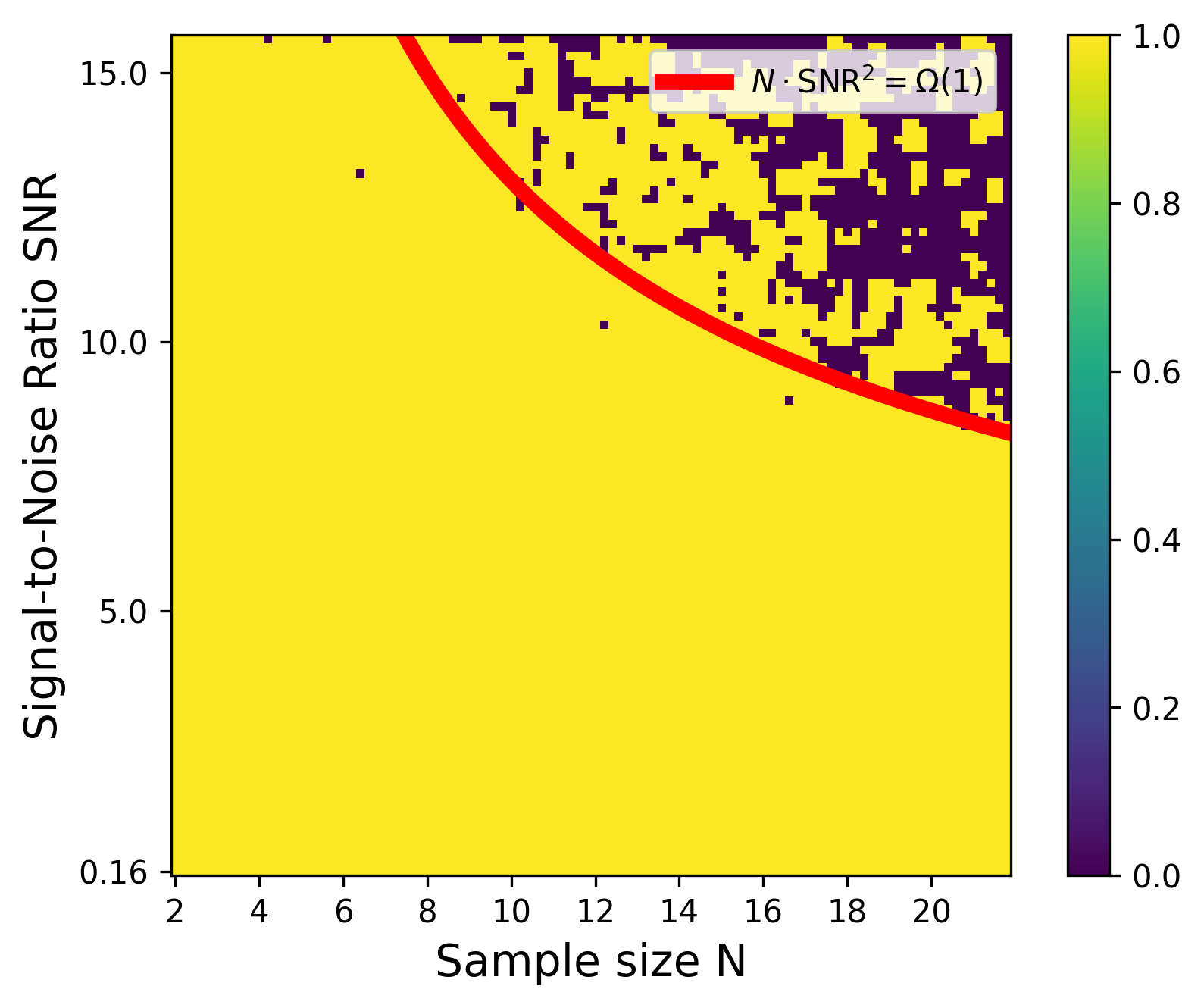}
    }
    \caption{Phase transition between benign and harmful overfitting based on Figure \ref{heatmap}, Map the parts with benign overfitting and the parts without benign overfitting to opposite colors. }
    \label{_heatmap}
\end{figure}

As illustrated in \cref{heatmap}, the test loss demonstrates a consistent upward trend as both the data volume and signal-to-noise ratio increase. Across various flipping probability conditions, while the mean test loss escalates with higher flipping probabilities, the overall loss distribution remains stable. This phenomenon can be attributed to the fact that label flipping alters the signal labels, effectively reducing the sample size \( \text{N} \) , thereby leading to an increase in average loss. However, since the reduction in \( \text{N} \)  does not impact the signal distribution, the loss distribution remains unchanged. This observation will be further substantiated in subsequent experimental validations.
 \cref{_heatmap} is the heat map after further processing of \cref{heatmap}, specifically highlighting the parts of benign overfitting. We have specifically highlighted the regions corresponding to benign overfitting and overlaid the theoretically derived formula as a fitted curve. Notably, as the flipping probability \( \alpha \) increases, The fitted curve has not undergone any spatial deformation. This observation aligns with our theoretical derivation, which indicates that the boundary of benign overfitting is solely determined by the signal-to-noise ratio (\( \text{SNR} \) ) and the sample size \( N \), independent of \( \alpha \). This consistency between empirical results and theoretical predictions further validates our analysis.

\subsection{The Impact of Critical Line}

In section 6.2 , we performed a preliminary analysis of the changes of the heat map in the loss of the test under the combined influence of \( \text{SNR} \), \( \text{N} \), and \(\alpha\). To eliminate the variable as a primary contributing factor, a systematic analysis of individual variables was conducted as detailed in Appendix \ref{a}. Furthermore, the experiment described in Section 6.2 was extended and enhanced through the inclusion of additional control parameters and validation tests.

\begin{figure}[t]
    \centering
    \subfigure[The test loss w.r.t. the critical line \( N \cdot \text{SNR}^2 = C \).]{
       \includegraphics[width=0.46\linewidth]{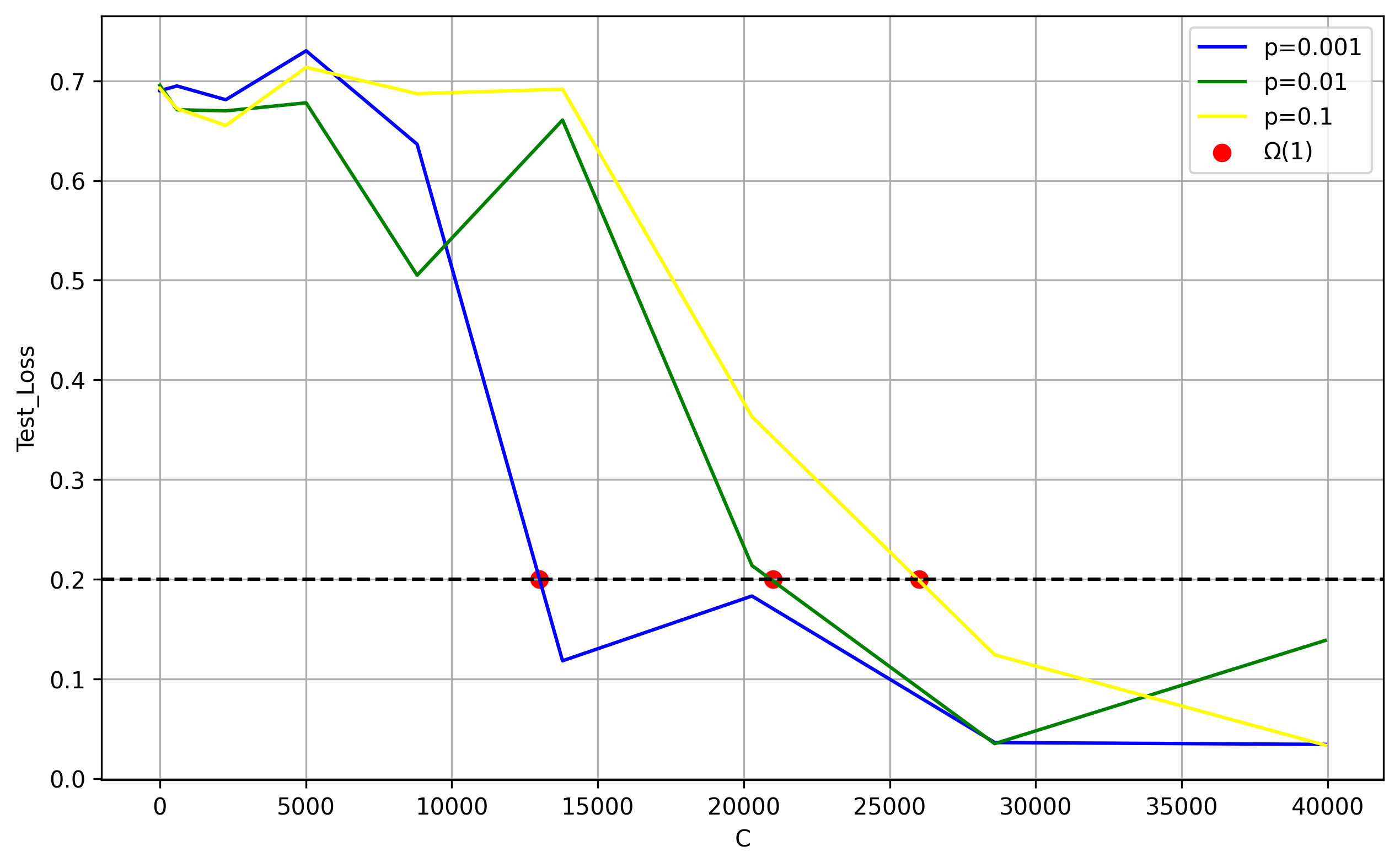}
    }
    \hfill
    \subfigure[Similarity analysis of \( \alpha=0.1 \) and \( \alpha =0.001 \).]{
       \includegraphics[width=0.46\linewidth]{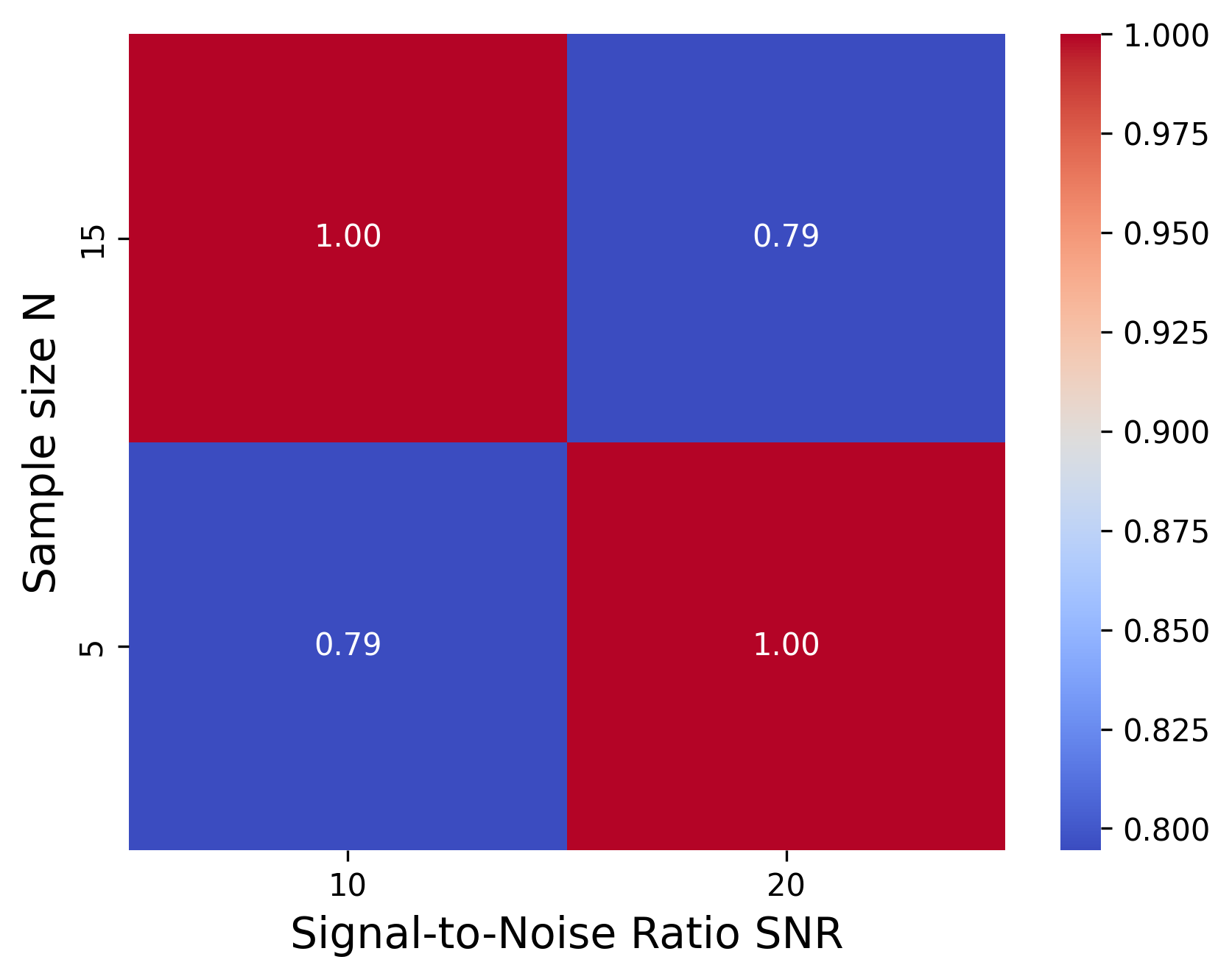}
    }
    \vspace{-0.2cm}
    
    \caption{(a) shows the variation of $C$ value with \(\alpha\), while (b) calculates the similarity between the two images \(\alpha=0.1\) and \(\alpha=0.001\), with higher scores indicating higher similarity.}
    \label{oumiga}
\end{figure}

 In theory, it is mentioned that \( N \cdot \text{SNR}^2 = \Omega(1) \), which is the minimum value at which benign overfitting occurs. We plotted the graph of this value in \cref{oumiga} under different conditions, and it is not difficult to see from that the probability of convergence to  \(\Omega(1)\))  varies for different \(\alpha\). \(\alpha\) increases this minimum value. And it is not difficult to find in \cref{_heatmap} that the curve has not undergone any deformation, indicating that \(\alpha\) does not affect the shape of the curve, but only causes the curve to shift in space, which has an impact on the constant C. From the (b) of the overall correlation score drawn \cref{oumiga}, it can be seen that \( \alpha = 0.1 \) and \( \alpha = 0.001 \) have a very high degree of similarity, which is sufficient to prove our conclusion that $\alpha$. does not change the distribution of the data.

\subsection{Learning Rate}

In the following section, we will explore some influencing factors that are rarely mentioned in other papers. In this section, we will investigate the impact of learning rate on test loss.
\begin{figure}
    \centering
    \includegraphics[width=0.45\linewidth]{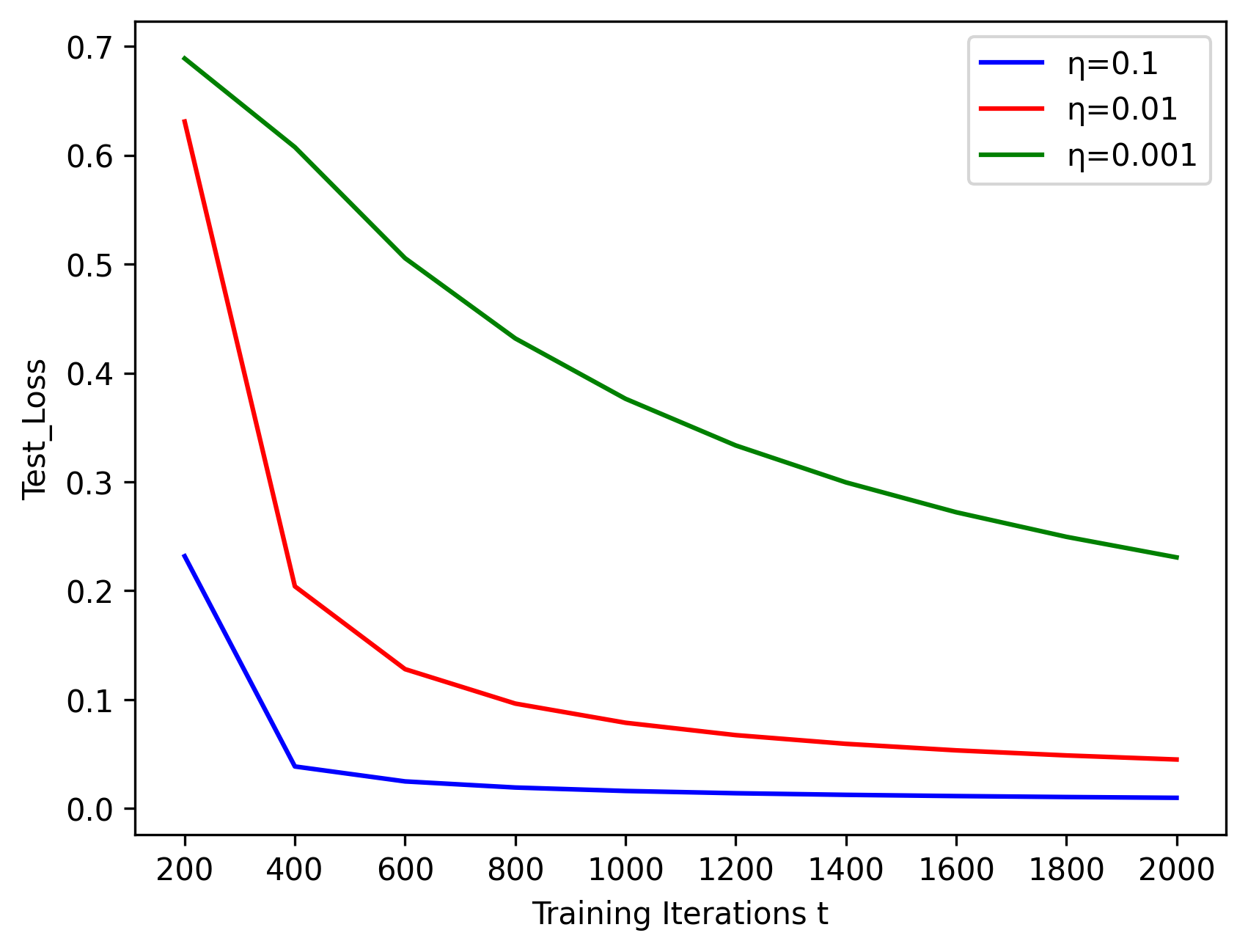}
    \caption{The test loss w.r.t. different leaning rates.}
    \label{learning rate}
 \end{figure}
In theory, there is an upper limit \textbf{\textit{T}} for the number of \textbf{\textit{t}}. In the experiment, we used 2000 iterations as \textbf{\textit{T}}. From the graph, it can be seen that the three learning rate curves have become flat, but there is a significant gap. Reducing the learning rate will increase the testing loss of the model, which is consistent with our theory.

\subsection{Initialization of Weight V Matrix}

\begin{figure}[t]
    \centering
    \subfigure[The impact of $\mathbf{W}_V$ Matrix Initialization.]{
       \includegraphics[width=0.46\linewidth]{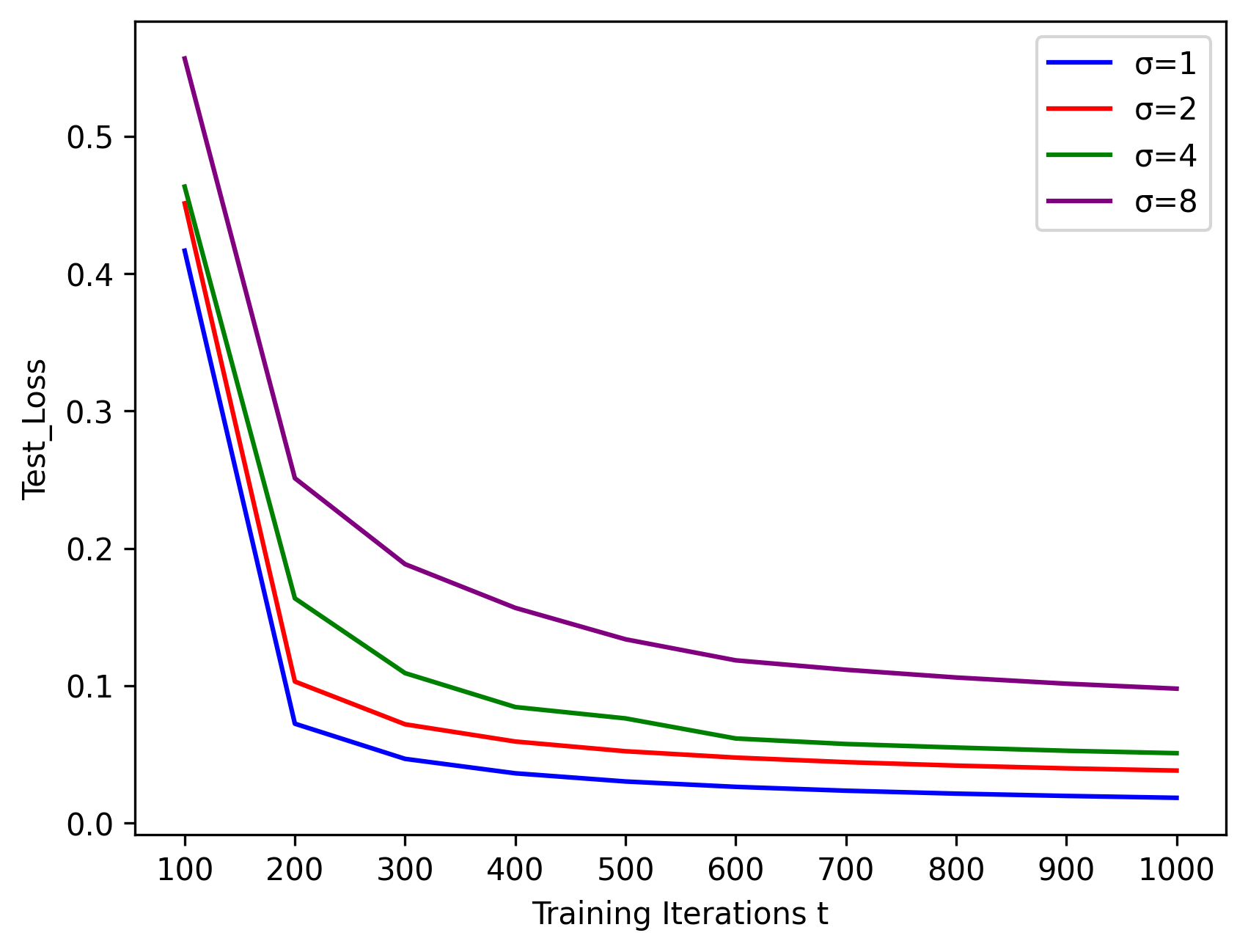}
    }
    \hfill
    \subfigure[Analysis of Optimal Initialization of $\mathbf{W}_V$ Matrix.]{
       \includegraphics[width=0.46\linewidth]{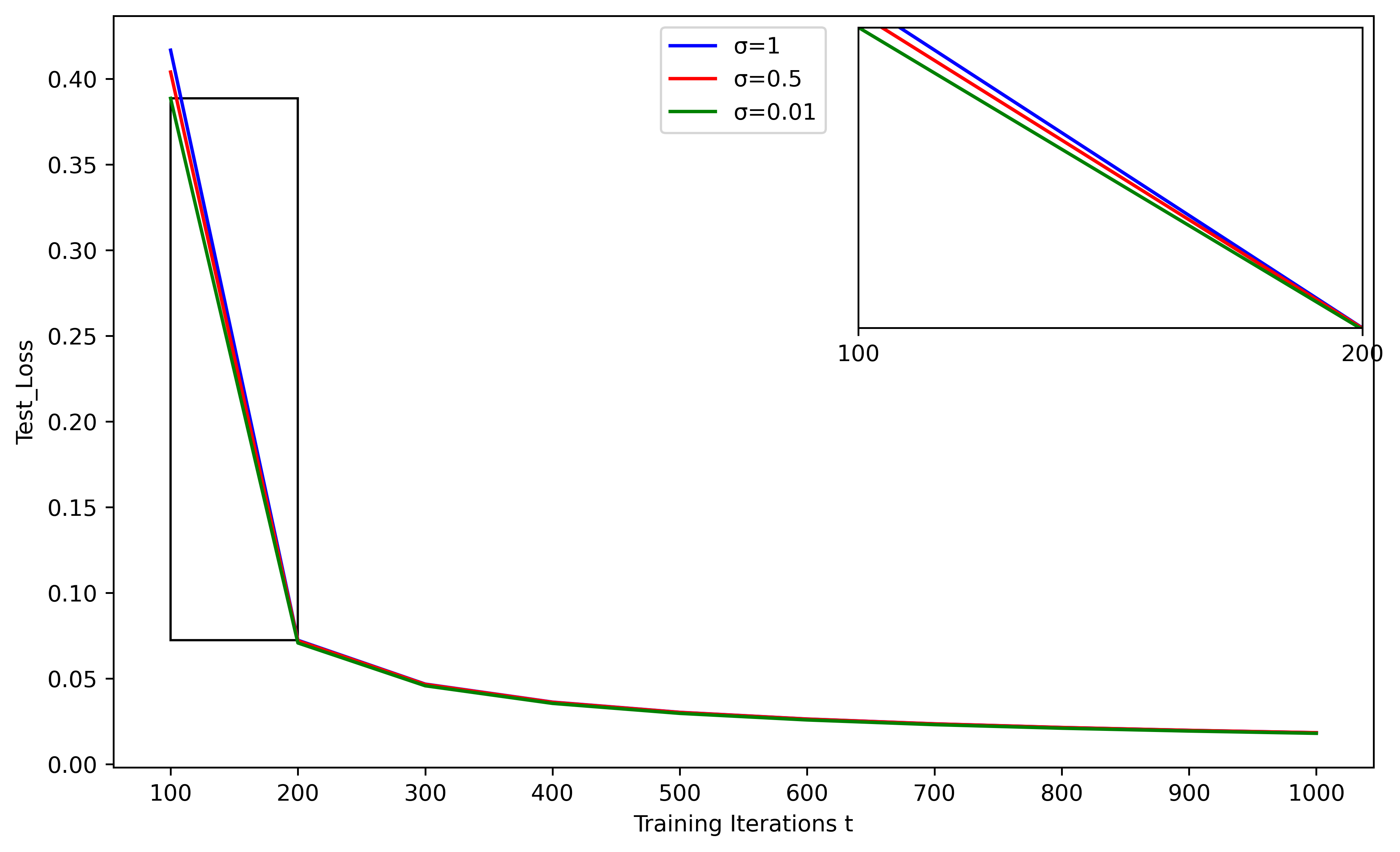}
    }
    \caption{The impact of the initilization of $\mathbf{W}_V$. (a) explores the test loss in terms of different initialization $\mathbf{W}_V$ matrices, and (b) further studies the optimal value of initialization.}
    \label{V}
\end{figure}

In this section, we examine the impact of the initialization of the value matrix \( \mathbf{W}_V \) on the test loss. From \cref{V} (a), we observe that as the initialization variance of the weight matrix decreases, the test loss shows a downward trend. To further investigate the effect of the weight matrix \( \mathbf{W}_V \) on the training loss, we designed a second experiment with a lower initialization variance, as shown in (b). It is clear that the variance stabilizes when it is no less than 1, suggesting that the optimal initialization is located near these curves.

\section*{Conclusion}
This paper studies the training dynamics, convergence, and generalization for a two-layer transformer with labeled
flip noise. Firstly, we present generalization
error bounds for both benign and harmful overfit
ting under varying signal-to-noise ratios (SNR),
Seconly, the training dynamics are categorized into
three distinct stages, each with its corresponding
error bounds. Overall, the techniques presented here pave the way for understanding the generalization of transformer-based models.

\section*{Acknowledgements}
The work of Jian Li is supported partially by National Natural Science Foundation of China (No. 62106257), and supported by “the Fundamental Research Funds for the Central Universities”. The work of Yong Liu is supported partially by National Natural Science Foundation of China (No.62076234), Beijing Outstanding Young Scientist Program (No.BJJWZYJH012019100020098), the Unicom Innovation Ecological Cooperation Plan, and the CCF-Huawei Populus Grove Fund.

\section*{Impact Statement}
This paper presents work whose goal is to advance the field of 
Machine Learning. There are many potential societal consequences 
of our work, none which we feel must be specifically highlighted here.

\bibliographystyle{plainnat}
\bibliography{main}

\begin{thebibliography}{29}
\providecommand{\natexlab}[1]{#1}
\providecommand{\url}[1]{\texttt{#1}}
\expandafter\ifx\csname urlstyle\endcsname\relax
  \providecommand{\doi}[1]{doi: #1}\else
  \providecommand{\doi}{doi: \begingroup \urlstyle{rm}\Url}\fi

\bibitem[Adlam et~al.(2021)Adlam, Levinson, and Pennington]{adlam2021randommatrixperspectivemixtures}
Ben Adlam, Jake Levinson, and Jeffrey Pennington.
\newblock A random matrix perspective on mixtures of nonlinearities for deep learning, 2021.
\newblock URL \url{https://arxiv.org/abs/1912.00827}.

\bibitem[Bartlett et~al.(2017)Bartlett, Foster, and Telgarsky]{bartlett2017spectrallynormalizedmarginboundsneural}
Peter Bartlett, Dylan~J. Foster, and Matus Telgarsky.
\newblock Spectrally-normalized margin bounds for neural networks, 2017.
\newblock URL \url{https://arxiv.org/abs/1706.08498}.

\bibitem[Bartlett et~al.(2020)Bartlett, Long, Lugosi, and Tsigler]{Bartlett_2020}
Peter~L. Bartlett, Philip~M. Long, Gábor Lugosi, and Alexander Tsigler.
\newblock Benign overfitting in linear regression.
\newblock \emph{Proceedings of the National Academy of Sciences}, 117\penalty0 (48):\penalty0 30063–30070, April 2020.
\newblock ISSN 1091-6490.
\newblock \doi{10.1073/pnas.1907378117}.
\newblock URL \url{http://dx.doi.org/10.1073/pnas.1907378117}.

\bibitem[Belkin et~al.(2018)Belkin, Ma, and Mandal]{belkin2018understanddeeplearningneed}
Mikhail Belkin, Siyuan Ma, and Soumik Mandal.
\newblock To understand deep learning we need to understand kernel learning, 2018.
\newblock URL \url{https://arxiv.org/abs/1802.01396}.

\bibitem[Belkin et~al.(2019)Belkin, Hsu, Ma, and Mandal]{Belkin_2019}
Mikhail Belkin, Daniel Hsu, Siyuan Ma, and Soumik Mandal.
\newblock Reconciling modern machine-learning practice and the classical bias–variance trade-off.
\newblock \emph{Proceedings of the National Academy of Sciences}, 116\penalty0 (32):\penalty0 15849–15854, July 2019.
\newblock ISSN 1091-6490.
\newblock \doi{10.1073/pnas.1903070116}.
\newblock URL \url{http://dx.doi.org/10.1073/pnas.1903070116}.

\bibitem[Belkin et~al.(2020)Belkin, Hsu, and Xu]{Belkin_2020}
Mikhail Belkin, Daniel Hsu, and Ji~Xu.
\newblock Two models of double descent for weak features.
\newblock \emph{SIAM Journal on Mathematics of Data Science}, 2\penalty0 (4):\penalty0 1167–1180, January 2020.
\newblock ISSN 2577-0187.
\newblock \doi{10.1137/20m1336072}.
\newblock URL \url{http://dx.doi.org/10.1137/20M1336072}.

\bibitem[Cao et~al.(2022)Cao, Chen, Belkin, and Gu]{cao2022benignoverfittingtwolayerconvolutional}
Yuan Cao, Zixiang Chen, Mikhail Belkin, and Quanquan Gu.
\newblock Benign overfitting in two-layer convolutional neural networks, 2022.
\newblock URL \url{https://arxiv.org/abs/2202.06526}.

\bibitem[Chen et~al.(2018)Chen, Jin, and Yu]{chen2018stabilityconvergencetradeoffiterative}
Yuansi Chen, Chi Jin, and Bin Yu.
\newblock Stability and convergence trade-off of iterative optimization algorithms, 2018.
\newblock URL \url{https://arxiv.org/abs/1804.01619}.

\bibitem[Frei and Vardi(2024)]{frei2024trainedtransformerclassifiersgeneralize}
Spencer Frei and Gal Vardi.
\newblock Trained transformer classifiers generalize and exhibit benign overfitting in-context, 2024.
\newblock URL \url{https://arxiv.org/abs/2410.01774}.

\bibitem[Hastie et~al.(2020)Hastie, Montanari, Rosset, and Tibshirani]{hastie2020surpriseshighdimensionalridgelesssquares}
Trevor Hastie, Andrea Montanari, Saharon Rosset, and Ryan~J. Tibshirani.
\newblock Surprises in high-dimensional ridgeless least squares interpolation, 2020.
\newblock URL \url{https://arxiv.org/abs/1903.08560}.

\bibitem[Huang et~al.(2023)Huang, Cheng, and Liang]{huang2023incontextconvergencetransformers}
Yu~Huang, Yuan Cheng, and Yingbin Liang.
\newblock In-context convergence of transformers, 2023.
\newblock URL \url{https://arxiv.org/abs/2310.05249}.

\bibitem[Jelassi et~al.(2022)Jelassi, Sander, and Li]{jelassi2022visiontransformersprovablylearn}
Samy Jelassi, Michael~E. Sander, and Yuanzhi Li.
\newblock Vision transformers provably learn spatial structure, 2022.
\newblock URL \url{https://arxiv.org/abs/2210.09221}.

\bibitem[Jiang et~al.(2024)Jiang, Huang, Zhang, Suzuki, and Nie]{jiang2024unveilbenignoverfittingtransformer}
Jiarui Jiang, Wei Huang, Miao Zhang, Taiji Suzuki, and Liqiang Nie.
\newblock Unveil benign overfitting for transformer in vision: Training dynamics, convergence, and generalization, 2024.
\newblock URL \url{https://arxiv.org/abs/2409.19345}.

\bibitem[Jin et~al.(2024)Jin, Balasubramanian, and Lai]{jin2024provableincontextlearningmixture}
Yanhao Jin, Krishnakumar Balasubramanian, and Lifeng Lai.
\newblock Provable in-context learning for mixture of linear regressions using transformers, 2024.
\newblock URL \url{https://arxiv.org/abs/2410.14183}.

\bibitem[Kou et~al.(2023)Kou, Chen, Chen, and Gu]{kou2023benignoverfittingtwolayerrelu}
Yiwen Kou, Zixiang Chen, Yuanzhou Chen, and Quanquan Gu.
\newblock Benign overfitting for two-layer relu convolutional neural networks, 2023.
\newblock URL \url{https://arxiv.org/abs/2303.04145}.

\bibitem[Li et~al.(2024{\natexlab{a}})Li, Huang, Han, Zhou, Suzuki, Zhu, and Chen]{li2024optimizationgeneralizationtwolayertransformers}
Bingrui Li, Wei Huang, Andi Han, Zhanpeng Zhou, Taiji Suzuki, Jun Zhu, and Jianfei Chen.
\newblock On the optimization and generalization of two-layer transformers with sign gradient descent, 2024{\natexlab{a}}.
\newblock URL \url{https://arxiv.org/abs/2410.04870}.

\bibitem[Li et~al.(2024{\natexlab{b}})Li, Wang, Lu, Cui, and Chen]{li2024nonlineartransformerslearngeneralize}
Hongkang Li, Meng Wang, Songtao Lu, Xiaodong Cui, and Pin-Yu Chen.
\newblock How do nonlinear transformers learn and generalize in in-context learning?, 2024{\natexlab{b}}.
\newblock URL \url{https://arxiv.org/abs/2402.15607}.

\bibitem[Liao et~al.(2021)Liao, Couillet, and Mahoney]{Liao_2021}
Zhenyu Liao, Romain Couillet, and Michael~W Mahoney.
\newblock A random matrix analysis of random fourier features: beyond the gaussian kernel, a precise phase transition, and the corresponding double descent*.
\newblock \emph{Journal of Statistical Mechanics: Theory and Experiment}, 2021\penalty0 (12):\penalty0 124006, December 2021.
\newblock ISSN 1742-5468.
\newblock \doi{10.1088/1742-5468/ac3a77}.
\newblock URL \url{http://dx.doi.org/10.1088/1742-5468/ac3a77}.

\bibitem[Magen et~al.(2024)Magen, Shang, Xu, Frei, Hu, and Vardi]{magen2024benignoverfittingsingleheadattention}
Roey Magen, Shuning Shang, Zhiwei Xu, Spencer Frei, Wei Hu, and Gal Vardi.
\newblock Benign overfitting in single-head attention, 2024.
\newblock URL \url{https://arxiv.org/abs/2410.07746}.

\bibitem[Mallinar et~al.(2024)Mallinar, Simon, Abedsoltan, Pandit, Belkin, and Nakkiran]{mallinar2024benigntemperedcatastrophictaxonomy}
Neil Mallinar, James~B. Simon, Amirhesam Abedsoltan, Parthe Pandit, Mikhail Belkin, and Preetum Nakkiran.
\newblock Benign, tempered, or catastrophic: A taxonomy of overfitting, 2024.
\newblock URL \url{https://arxiv.org/abs/2207.06569}.

\bibitem[Meng et~al.(2024)Meng, Yao, and Cao]{JMLR:v25:22-1389}
Xuran Meng, Jianfeng Yao, and Yuan Cao.
\newblock Multiple descent in the multiple random feature model.
\newblock \emph{Journal of Machine Learning Research}, 25\penalty0 (44):\penalty0 1--49, 2024.
\newblock URL \url{http://jmlr.org/papers/v25/22-1389.html}.

\bibitem[Mou et~al.(2017)Mou, Wang, Zhai, and Zheng]{mou2017generalizationboundssgldnonconvex}
Wenlong Mou, Liwei Wang, Xiyu Zhai, and Kai Zheng.
\newblock Generalization bounds of sgld for non-convex learning: Two theoretical viewpoints, 2017.
\newblock URL \url{https://arxiv.org/abs/1707.05947}.

\bibitem[Neyshabur et~al.(2018)Neyshabur, Li, Bhojanapalli, LeCun, and Srebro]{neyshabur2018understandingroleoverparametrizationgeneralization}
Behnam Neyshabur, Zhiyuan Li, Srinadh Bhojanapalli, Yann LeCun, and Nathan Srebro.
\newblock Towards understanding the role of over-parametrization in generalization of neural networks, 2018.
\newblock URL \url{https://arxiv.org/abs/1805.12076}.

\bibitem[Shang et~al.(2024)Shang, Meng, Cao, and Zou]{shang2024initializationmattersbenignoverfitting}
Shuning Shang, Xuran Meng, Yuan Cao, and Difan Zou.
\newblock Initialization matters: On the benign overfitting of two-layer relu cnn with fully trainable layers, 2024.
\newblock URL \url{https://arxiv.org/abs/2410.19139}.

\bibitem[Tarzanagh et~al.(2024)Tarzanagh, Li, Thrampoulidis, and Oymak]{tarzanagh2024transformerssupportvectormachines}
Davoud~Ataee Tarzanagh, Yingcong Li, Christos Thrampoulidis, and Samet Oymak.
\newblock Transformers as support vector machines, 2024.
\newblock URL \url{https://arxiv.org/abs/2308.16898}.

\bibitem[Tian et~al.(2023)Tian, Wang, Chen, and Du]{tian2023scansnapunderstandingtraining}
Yuandong Tian, Yiping Wang, Beidi Chen, and Simon Du.
\newblock Scan and snap: Understanding training dynamics and token composition in 1-layer transformer, 2023.
\newblock URL \url{https://arxiv.org/abs/2305.16380}.

\bibitem[Tsigler and Bartlett(2022)]{tsigler2022benignoverfittingridgeregression}
A.~Tsigler and P.~L. Bartlett.
\newblock Benign overfitting in ridge regression, 2022.
\newblock URL \url{https://arxiv.org/abs/2009.14286}.

\bibitem[Zhang et~al.(2017)Zhang, Bengio, Hardt, Recht, and Vinyals]{zhang2017understandingdeeplearningrequires}
Chiyuan Zhang, Samy Bengio, Moritz Hardt, Benjamin Recht, and Oriol Vinyals.
\newblock Understanding deep learning requires rethinking generalization, 2017.
\newblock URL \url{https://arxiv.org/abs/1611.03530}.

\bibitem[Zou et~al.(2021)Zou, Wu, Braverman, Gu, and Kakade]{zou2021benignoverfittingconstantstepsizesgd}
Difan Zou, Jingfeng Wu, Vladimir Braverman, Quanquan Gu, and Sham~M. Kakade.
\newblock Benign overfitting of constant-stepsize sgd for linear regression, 2021.
\newblock URL \url{https://arxiv.org/abs/2103.12692}.

\end{thebibliography}
\newpage

\appendix
\onecolumn


\section{Symbol Notions}
\begin{table}[h]
    \centering
    \caption{Notions related to Query (Q), Key (K), and Value (V). }
    \begin{tabular}{|l|l|}
        \hline
        Symbols & Notions \\
        \hline
        $q_+^{(t)},q_-^{(t)},q_{\xi,i}^{(t)}$ & vectorized Q, defined as \\
        & $q_+^{(t)}=\mu_+^{\top}W_Q^{(t)},q_-^{(t)}=\mu_-^{\top}W_Q^{(t)},q_{\xi,i}^{(t)}=\xi_{i}^{\top}W_Q^{(t)}$ \\
        \hline
        $k_+^{(t)},k_-^{(t)},k_{\xi,i}^{(t)},k_{\xi,i'}^{(t)}$ & vectorized K, defined as \\
        & $k_+^{(t)}=\mu_+^{\top}W_K^{(t)},k_-^{(t)}=\mu_-^{\top}W_K^{(t)},k_{\xi,i}^{(t)}=\xi_{i}^{\top}W_K^{(t)},k_{\xi,i'}^{(t)}=\xi_{i'}^{\top}W_K^{(t)}$ \\
        \hline
        $V_+^{(t)},V_-^{(t)},V_{\xi,i}^{(t)}$ & scalarized V, defined as \\
        & $V_+^{(t)}=\mu_+^{\top}W_V^{(t)}\upsilon,V_-^{(t)}:=\mu_-^{\top}W_V^{(t)}\upsilon,V_{\xi,i}^{(t)}:=\xi_{i}^{\top}W_V^{(t)}\upsilon$ \\
        \hline        $\Lambda_{\xi,\pm,i}^{(t)},\Lambda_{\xi,i,\pm,i'}^{(t)}$ & $\Lambda_{\xi,\pm,i}^{(t)}:=\langle\boldsymbol{q}_{\pm}^{(t)},\boldsymbol{k}_{\pm}^{(t)}\rangle - \langle\boldsymbol{q}_{\pm}^{(t)},\boldsymbol{k}_{\xi,i}^{(t)}\rangle$, \\
        & $\Lambda_{\xi,i,\pm,i'}^{(t)}:=\langle\boldsymbol{q}_{\xi,i}^{(t)},\boldsymbol{k}_{\pm}^{(t)}\rangle - \langle\boldsymbol{q}_{\xi,i}^{(t)},\boldsymbol{k}_{\xi,i'}^{(t)}\rangle$ \\
        \hline
    \end{tabular}
\end{table}
\begin{table}[h]
    \centering
    \caption{Notions related to softmax. }
    \begin{tabular}{|l|l|}
        \hline
        Symbols & Notions \\
        \hline
        $S_{11}$ & a general reference to $\frac{\exp(\langle q_+^{(t)},k_+^{(t)}\rangle)}{\exp(\langle q_+^{(t)},k_+^{(t)}\rangle)+\exp(\langle q_+^{(t)},k_{\xi,i}^{(t)}\rangle)}$ for $i \in S_+$, and \\
        & $\frac{\exp(\langle q_-^{(t)},k_+^{(t)}\rangle)}{\exp(\langle q_-^{(t)},k_+^{(t)}\rangle)+\exp(\langle q_-^{(t)},k_{\xi,i}^{(t)}\rangle)}$ for $i \in S_-$ \\
        \hline
        $S_{21}$ & a general reference to $\frac{\exp(\langle q_{\xi,i}^{(t)},k_+^{(t)}\rangle)}{\exp(\langle q_{\xi,i}^{(t)},k_+^{(t)}\rangle)+\exp(\langle q_{\xi,i}^{(t)},k_{\xi,i'}^{(t)}\rangle)}$ for $i,i' \in S_+$, and \\
        & $\frac{\exp(\langle q_{\xi,i}^{(t)},k_-^{(t)}\rangle)}{\exp(\langle q_{\xi,i}^{(t)},k_-^{(t)}\rangle)+\exp(\langle q_{\xi,i}^{(t)},k_{\xi,i'}^{(t)}\rangle)}$ for $i,i'\in S_-$ \\
        \hline
        $S_{12}$ & a general reference to $\frac{\exp(\langle q_+^{(t)},k_{\xi,i}^{(t)}\rangle)}{\exp(\langle q_+^{(t)},k_+^{(t)}\rangle)+\exp(\langle q_+^{(t)},k_{\xi,i}^{(t)}\rangle)}$ for $i \in S_+$, and \\
        & $\frac{\exp(\langle q_-^{(t)},k_{\xi,i}^{(t)}\rangle)}{\exp(\langle q_-^{(t)},k_-^{(t)}\rangle)+\exp(\langle q_-^{(t)},k_{\xi,i}^{(t)}\rangle)}$ for $i \in S_-$ \\
        \hline
        $S_{22}$ & a general reference to $\frac{\exp(\langle q_{\xi,i}^{(t)},k_{\xi,i'}^{(t)}\rangle)}{\exp(\langle q_{\xi,i}^{(t)},k_+^{(t)}\rangle)+\exp(\langle q_{\xi,i}^{(t)},k_{\xi,i'}^{(t)}\rangle)}$ for $i,i' \in S_+$, and \\
        & $\frac{\exp(\langle q_{\xi,i}^{(t)},k_{\xi,i'}^{(t)}\rangle)}{\exp(\langle q_{\xi,i}^{(t)},k_-^{(t)}\rangle)+\exp(\langle q_{\xi,i}^{(t)},k_{\xi,i'}^{(t)}\rangle)}$ for $i,i' \in S_-$ \\
        \hline
    \end{tabular}
\end{table}
\section{Detailed Proof}
\subsection{Detailed Proof in 5.1.}
We can decompose the test error as follows in bengin overfitting:
\begin{equation*}
    \begin{aligned}
        &\mathbb{P}(y\neq \text{sign}(f(\theta, \mathbf{X}, \upsilon)))\\
        \leq & \alpha+\mathbb{P}(\widehat{y}f(\theta, \mathbf{X}, \upsilon)\leq 0).
    \end{aligned}
    \tag{2}
\end{equation*}
\textit{Proof} of \eqref{eq:formula2}.
We can write out the test error as
\begin{align*}
L_{\mathcal{D}}(\mathbf{W}^{(t)}) 
=& P_{(\mathbf{X},\widehat{y}, y)\sim\mathcal{D}}(y \neq \text{sign}(f(\theta,\mathbf{X},\upsilon))) \\
=& P_{(\mathbf{X},\widehat{y}, y)\sim\mathcal{D}}(yf(\theta,\mathbf{X},\upsilon) \leq 0) \\
=& P_{(\mathbf{X},\widehat{y}, y)\sim\mathcal{D}}(yf(\theta,\mathbf{X},\upsilon) \leq 0, y \neq \widehat{y}) + P_{(\mathbf{X},\widehat{y},y)\sim\mathcal{D}}(yf(\theta,\mathbf{X},\upsilon) \leq 0, y = \widehat{y}) \\
=& \alpha \cdot P_{(\mathbf{X},\widehat{y},y)\sim\mathcal{D}}(\widehat{y}f(\theta,\mathbf{X},\upsilon)\geq 0) + (1-\alpha) \cdot P_{(\mathbf{X},\widehat{y},y)\sim\mathcal{D}}(\widehat{y}f(\theta,\mathbf{X},\upsilon) \leq 0) \\
\leq& \alpha + P_{(\mathbf{X},\widehat{y},y)\sim\mathcal{D}}(\widehat{y}f(\theta,\mathbf{X},\upsilon) \leq 0),
\end{align*}
In the second and third equation, we used the definition of $\mathcal{D}$ in Definition \ref{def.Data Generation Model}.

\subsection{Proof of The Theorem}
\begin{theorem}
    \label{thm_1}
    Under Assumption \ref{Definition.4.1}, in the theoretical analysis of test error in the second and third stages of benign overfitting, we define \(g(\xi)\) as \(V_{\xi,(-\widehat{y})}^{(t)} = \sum_{r} \left\langle \upsilon W_{-\widehat{y}, r}^{(t)}, \xi\right\rangle\), where \(W_{-\widehat{y}, r}^{(t)}\) refers to the row vector in parameter matrix \(W\) with label \(-\widehat{y}\) and index \(r\) at the \(t\)-th training round. Then, we know that for any \(x \geq 0\), if \(g:\mathbb{R}^n\rightarrow\mathbb{R}\) is a Lipschitz function and \(c\) is a constant, the following inequality holds for the test loss. 
    \begin{equation*}
        \mathbb{P}(g(\xi) - \mathbb{E}g(\xi) \geq x) \leq \exp\left(-\frac{cx^2}{\sigma_p^2 \sum_{r = 1}^{m_v} \left\| W_{-\widehat{y},r}^{(t)} \upsilon  \right\|_{2}^2}\right).
    \end{equation*}
\end{theorem}
\textit{Proof.} According to Theorem 5.2.2 in Vershynin (2018), we know that for any \(x \geq 0\), if \(g:\mathbb{R}^n\rightarrow\mathbb{R}\) is a Lipschitz function, it holds that
    \begin{equation}
        \mathbb{P}(g(\xi) - \mathbb{E}g(\xi) \geq x) \leq \exp\left(-\frac{cx^2}{\sigma_p^2 \|g\|_{\text{Lip}}^2}\right),
        \label{eq:formula4}
    \end{equation}
    Since \(g(\xi)\) is defined as \(V_{\xi,(-\widehat{y})}^{(t)} = \sum_{r} \left\langle \upsilon W_{-\widehat{y}, r}^{(t)}, \xi\right\rangle\), and since \(\langle W_{-\widehat{y}, r}^{(t)}\upsilon , \xi \rangle \sim \mathcal{N}(0, \|W_{-\widehat{y}, r}^{(t)}\|_2^2 \| \upsilon \|_2^2 \sigma_p^2)\), we have
    \begin{align*}
        \left| g(\xi) - g(\xi') \right| 
        &= \left| \sum_{r=1}^{m_v} \left\langle \upsilon W_{-\widehat{y},r}^{(t)}, \xi \right\rangle - \sum_{r=1}^{m_v} \left\langle \upsilon W_{-\widehat{y},r}^{(t)}, \xi' \right\rangle \right| \\
        &\leq \sum_{r=1}^{m_v} \left| \left\langle \upsilon W_{-\widehat{y},r}^{(t)}, \xi \right\rangle - \left\langle\upsilon W_{-\widehat{y},r}^{(t)}, \xi' \right\rangle \right| \\
        &= \sum_{r=1}^{m_v} \left| \left\langle \upsilon W_{-\widehat{y},r}^{(t)}, \xi - \xi' \right\rangle \right| \\
        &\leq \sum_{r=1}^{m_v} \| W_{-\widehat{y},r}^{(t)} \|_2 \| \upsilon\|_2 \| \xi - \xi' \|_2
    \end{align*}
So, we can get
\begin{equation}
\|g\|_{\text{Lip}} \leq \sum_{r=1}^{m_v} \left\| W_{-\widehat{y},r}^{(t)} \ \upsilon \right\|_{2}, 
\label{eq:formula5}
\end{equation}
By plugging \eqref{eq:formula5} into \eqref{eq:formula4},  we get:
\begin{equation*}
    \mathbb{P}(g(\xi) - \mathbb{E}g(\xi) \geq x) \leq \exp\left(-\frac{cx^2}{\sigma_p^2 \sum_{r = 1}^{m_v} \left\| W_{-\widehat{y},r}^{(t)} \upsilon  \right\|_{2}^2}\right).
\end{equation*}
\subsection{Proof of The Lemma}
\begin{lemma}[The test loss of benign overfitting in Lemma \ref{lem_the_test_loss_benign_overfitting}]
\label{lem_The_Test_Loss_in_Benign_Overfitting}
\begin{align}
P(\widehat{y}(f(\theta,\mathbf{X},\upsilon)) \leq 0) 
&= P\left(\sum_{r} \left[(S_{11} + S_{21})V_{+r,\widehat{y}}^{(t)} + (S_{11} + S_{21}) V_{-r,\widehat{y}}^{(t)} + (S_{12} + S_{22}) V_{\xi,ir,\widehat{y}}^{(t)}\right] \right. \notag \\
&\quad\left. \leq \sum_{r} \left[(S_{11} + S_{21})(V_{+r}^{(t)} + V_{-r}^{(t)}) + (S_{12} + S_{22}) V_{\xi,ir,(-\widehat{y})}^{(t)}\right]\right) \notag \\
&\leq P\left(\sum_{r}\frac{S_{11}+S_{21}}{S_{12}+S_{22}}\left(V_{+r,\widehat{y}}^{(t)}+V_{-r,\widehat{y}}^{(t)}\right)\leq V_{\xi,ir,(-\widehat{y})}^{(t)}+o(1)\right)
\end{align}
\end{lemma}
\textit{Proof.} where the first equality is by the conversion relationship between $V_{+}^{(t)}, V_{+,\widehat{y}}^{(t)}, V_{+,(-\widehat{y})}^{(t)}$; the second inequality is because it is a benign overfitting stage, the signal memory is higher than the noise memory.  Here we see that the left side of the inequality is
dominated by the signal memory, and the right side of the inequality satisfies \eqref{eq:formula16} and \eqref{eq:formula19}.
\begin{lemma}[Relationship of constants]
\label{lem_Relationship_of_Constants}
To ensure the continuity of the test error function, we need to verify that when $t = T_1$, the first - stage test error equals the initial test error of the second stage. When $t = T_2$, the second - stage test error is equal to the initial test error of the third stage. It is evident that this holds true when $t = T_1$. We only verify this for $t = T_2$.\\
\textit{Proof.}
\begin{align*}
\alpha + \exp\left[\frac{c_4}{2\pi}- c_{10}\eta^4\|\mu\|^8(T_2 - T_1)^2(T_2 - T_1 - 1)^2\text{SNR}^2\right]=&\alpha+\exp\left(\frac{c_{12}}{2\pi}\right)\\
\frac{c_4 - c_{12}}{2\pi\cdot c_{10}}=\eta^4\|\mu\|^8(T_2 - T_1)^2(T_2 - T_1 - 1)^2\text{SNR}^2&
\end{align*}
\text{It is sufficient to ensure that the above relations are satisfied among the three constants. }
\end{lemma}
\section{Basic Inequality}
\subsection{Benign Overfitting}
\text{According to \cite{jiang2024unveilbenignoverfittingtransformer}, we know  the following inequality holds:}
when benign overfitting occurs in the second stage, for \(t\in(T_1,T_2]\) , the following inequality holds:
\begin{align}
&\frac{\exp(\langle q_+^{(t)},k_{\xi,i}^{(t)}\rangle)}{\exp(\langle q_+^{(t)},k_+^{(t)}\rangle)+\exp(\langle q_+^{(t)},k_{\xi,i}^{(t)}\rangle)}\notag\\
&\leq \frac{\exp(\langle q_+^{(t)},k_{\xi,i}^{(t)}\rangle)}{\exp(\langle q_+^{(t)},k_+^{(t)}\rangle)} \notag\\
&= \frac{1}{c_5\exp(\Lambda_{\xi,\pm,i}^{(t)})}\label{eq:formula7}
\end{align}
\begin{align}
&\frac{\exp(\langle q_{\xi,i}^{(t)},k_{\xi,i'}^{(t)}\rangle)}{\exp(\langle q_{\xi,i}^{(t)},k_+^{(t)}\rangle)+\exp(\langle q_{\xi,i}^{(t)},k_{\xi,i'}^{(t)}\rangle)}\notag\\
&\leq \frac{1}{c_7\exp(\Lambda_{\xi,i,\pm,i'}^{(t)})} \label{eq:formula8}
\end{align}
\begin{align}
V_{+}^{(t)} &\geq \eta c_1\lVert\boldsymbol{\mu}\rVert_2^2\lVert\upsilon\rVert_2^2(t - T_1) \label{eq:formula9}\\
|V_{\pm}^{(t)}| &\leq O(d^{-\frac{1}{4}}) + \eta c_4\lVert\boldsymbol{\mu}\rVert_2^2\lVert\upsilon\rVert_2^2(t - T_1) \label{eq:formula10}
\end{align}
\begin{align}
    \Lambda_{\xi,\pm,i}^{(t+1)} &\geq \log \left( \exp(\Lambda_{\xi,\pm,i}^{(T_1)}) + \frac{\eta^2 c_8 \|\boldsymbol{\mu}\|_2^4 \|\upsilon\|_2^2 d^{\frac{1}{2}}}{N (\log(24N^2 / \delta))^2} \cdot (t - T_1)(t - T_1 + 1) \right) \label{eq:formula11}
\end{align}
\begin{align}
   \Lambda_{\xi,i,\pm,i'}^{(t+1)} &\geq \log \left( \exp(\Lambda_{\xi,i,\pm,i'}^{(T_1)}) + \frac{\eta^2 c_8 \sigma_p^2 d \|\boldsymbol{\mu}\|_2^2 \|\upsilon\|_2^2 d^{\frac{1}{2}}}{N (\log(24N^2 / \delta))^2} \cdot (t - T_1)(t - T_1 + 1) \right) \label{eq:formula12}
\end{align}
\begin{align}
\left\|\boldsymbol{W}_{V}^{(t + 1)}\upsilon-\boldsymbol{W}_{V}^{(t)}\upsilon\right\|_{2}=O\left(\eta\cdot\max\left\{\left\|\boldsymbol{\mu}\right\|_{2},\sigma_{p}\sqrt{d}\right\}\cdot\left\|\upsilon\right\|_{2}\right)\label{eq:formula13}
\end{align}
\begin{align}
V_{+}^{(T_{2})}\geq6\cdot\left|V_{\xi,i}^{(T_{2})}\right|,
\label{eq:formula14}\\
V_{-}^{(T_{2})}\leq-6\cdot\left|V_{\xi,i}^{(T_{2})}\right|.
\label{eq:formula15}
\end{align}
\begin{equation}
\label{eq:formula16}
\left|V_{+}^{(T_2)}\right|,\left|V_{-}^{(T_2)}\right|,\left|V_{\xi,i}^{(T_2)}\right|=o(1),
\end{equation}
when benign overfitting occurs in the third stage, for \(t\in(T_2,T_3]\), the following inequality holds:
\begin{align}
&\log\left(\exp\left(V_+^{(T_2)}\right)+\eta c_{11}\lVert\boldsymbol{\mu}\rVert_2^2\lVert\upsilon\rVert_2^2(t - T_2)\right)\leq V_+^{(t)}\leq 2\log\left(O\left(\frac{1}{\epsilon}\right)\right),
\label{eq:formula17}\\
&-2\log\left(O\left(\frac{1}{\epsilon}\right)\right)\leq V_-^{(t)}\leq -\log\left(\exp\left(-V_-^{(T_2)}\right)+\eta c_{11}\lVert\boldsymbol{\mu}\rVert_2^2\lVert\upsilon\rVert_2^2(t - T_2)\right)
\label{eq:formula18}
\end{align}
\begin{equation}
\label{eq:formula19}
\left|V_{+}^{(t)}\right|,\left|V_{-}^{(t)}\right|,\left|V_{\xi,i}^{(t)}\right|=o(1),
\end{equation}
\subsection{Harmful Overfitting}
\text{According to \cite{jiang2024unveilbenignoverfittingtransformer}, we know  the following inequality holds:}
when harmful overfitting occurs in the second stage, for \(t\in(T_1,T_2]\), the following inequality holds:
\begin{align}
V_{\xi,i}^{(t)} &\geq \frac{\eta c_{13}\sigma_p^2 d \| \upsilon\|_2^2 (t - T_1)}{N} \label{eq:formula20}\\
V_{\xi,i}^{(t)} &\leq - \frac{\eta c_{14}\sigma_p^2 d \| \upsilon \|_2^2 (t - T_1)}{N} \label{eq:formula21}\\
|V_{\pm}^{(t)}| &\leq O(d^{-\frac{1}{4}}) + \frac{\eta c_{15}\sigma_p^2 d \|\upsilon\|_2^2 (t - T_1)}{N} \label{eq:formula22}\\
|V_{\xi,i}^{(t)}| &\leq O(d^{-\frac{1}{4}}) + \frac{\eta c_{16}\sigma_p^2 d \| \upsilon\|_2^2 (t - T_1)}{N} \label{eq:formula23}
\end{align}
\begin{align}
\mathrm{Softmax}(\langle \boldsymbol{q}_{\pm}^{(t)}, \boldsymbol{k}_{\pm}^{(t)} \rangle) &= O\left(\frac{\sigma_p^2 d (\log(24N^2 / \delta))^3}{\|\boldsymbol{\mu}\|_2^2 \|\upsilon \|_2^2 d^{\frac{1}{2}}}\right) \label{eq:formula24}\\
\mathrm{Softmax}(\langle \boldsymbol{q}_{\xi,i}^{(t)}, \boldsymbol{k}_{\xi,i'}^{(t)} \rangle) &= 1 - O\left(\frac{(\log(24N^2 / \delta))^3}{\|\upsilon\|_2^2 d^{\frac{1}{2}}}\right) \label{eq:formula25}
\end{align}
\begin{equation}
\label{eq:formula26}
\left|V_{+}^{(T_2)}\right|,\left|V_{-}^{(T_2)}\right|,\left|V_{\xi,i}^{(T_2)}\right|=o(1),
\end{equation}
When harmful overfitting occurs in the third stage, for \(t\in(T_2,T_3]\), the following inequality holds:
\begin{align}
V_{\pm}^{(T_3)} = o(1) \label{eq:formula27}
\end{align}
\begin{equation}
\label{eq:formula28}
\left|V_{+}^{(t)}\right|,\left|V_{-}^{(t)}\right|,\left|V_{\xi,i}^{(t)}\right|=o(1),
\end{equation}
\begin{align}
\widehat{y}(f(\theta,\mathbf{X},\upsilon)) &= \frac{1}{m_v}\sum_{r = 1}^{m_v} \left(\upsilon^T x_1(S_{11} + S_{21})W_{Vr}^{(t)} + \upsilon^T x_2(S_{12} + S_{22})W_{Vr}^{(t)}\right) \notag \\
&\geq \log(1 + e^{-1/2}), \text{ with probability at least } \frac{1}{2} \label{eq:formula29}
\end{align}
\section{Upper Bound of Benign Overfitting}
\subsection{Stage I Test Loss}
\begin{theorem}[\textbf{(First part of Theorem \ref{thm:4.1})}]
    Under the same conditions as Theorem \ref{thm:4.1}, when $N \cdot \text{SNR}^2 = \Omega(1)$, there exists \(T_1=\frac{140N}{\eta d^4(N\|\mu\|_2^2 - 120^2C_p^2\sigma_p^2d)\|\upsilon\|_2^2}\), and for \(t\in(0,T_1]\), the test error is:
\begin{equation*}
L_{\mathcal{D}}(\mathbf{W}^{(t)})\leq \alpha + O(1)\\
\end{equation*}
\label{thm:4.1.1}
\end{theorem}
\textit{Proof.} In the paper, the high test error in the first stage of benign overfitting, which is $\alpha + O(1)$, can be understood from the following aspects.In terms of model structure, during the initial phase of gradient optimization, the error decline is slow. After initialization, the weight matrices in the Transformer, such as $W_Q$, $W_K$, and $W_V$, have not been effectively adjusted. This makes it difficult for the model to capture the signal features of the data. From the perspective of the attention mechanism related to $Q$ and $K$ in the initial stage of the transformer, it fails to clearly distinguish between signals and noise. At the beginning, the values of the $Q$ and $K$ matrices are randomly initialized and not well-tuned. As a result, the output of the activation function assigns relatively uniform attention weights to different elements in the sequence.
Rather than concentrating more attention on the signal parts of the data, the attention mechanism spreads its focus evenly across both signal and noise. This means that the model is unable to effectively highlight the important signal features in the training data, leading to insufficient utilization of the signal information and poor data learning ability. These factors work together to make the test error high and equal to $\alpha + O(1)$.
\subsection{Stage II Test Loss}
\begin{theorem}[\textbf{(Second part of Theorem \ref{thm:4.1})}]
Under the same conditions as Theorem \ref{thm:4.1}, when $N \cdot \text{SNR}^2 = \Omega(1)$,there exists \(T_2=\Theta\left(\frac{1}{\eta\|\mu\|_2^2\|\upsilon\|_2^2}\right)\), and for \(t\in(T_1,T_2]\), the test error is:
\begin{equation*}
L_{\mathcal{D}}(\mathbf{W}^{(t)}) \leq \alpha+\exp\left(\frac{c_4}{2\pi}-c_{10}\eta^4\|\mu\|_2^8(t - T_1)^2(t - T_1 - 1)^2\text{SNR}^2\right)
\end{equation*}
\end{theorem}
\textit{Proof.} For the sake of convenience, we use $(X,\widehat{y}, y)\sim\mathcal{D}$ to denote the following: data point $(X, y)$ follows distribution $\mathcal{D}$ defined in  Assumption \ref{Definition.4.1} , and $\widehat{y}$ is its true label. We can write out the test error as \eqref{eq:formula2},
\begin{align*}
L_{\mathcal{D}}(\mathbf{W}^{(t)}) 
\leq& \alpha + P_{(X,\widehat{y},y)\sim\mathcal{D}}(\widehat{y}f(\theta,\mathbf{X},\upsilon) \leq 0),
\end{align*}
Therefore we need to calculate an upper bound for $P_{(X,\widehat{y})\sim\mathcal{D}}(\widehat{y} f(\theta,\mathbf{X},\upsilon) \leq 0)$.\\
\text{To achieve this and take into account the existence of label flipping ,we write } $X = (y\mu, \xi)$,  we can get the following inequality from lemma \ref{lem_The_Test_Loss_in_Benign_Overfitting}.
\begin{equation*}
P(\widehat{y}f(\theta,\mathbf{X},\upsilon) \leq 0)
\leqslant P\left(\sum_{r}\frac{S_{11}+S_{21}}{S_{12}+S_{22}}\left(V^{(t)}_{+r,\widehat{y}}+V^{(t)}_{-r,\widehat{y}}\right)\leq V^{(t)}_{\xi,(-r\widehat{y})}+o(1)\right).
\end{equation*}
Denote \(g(\xi)\) as \(V_{\xi,(-\widehat{y})}^{(t)} = \sum_{r} \left\langle \upsilon W_{-\widehat{y}, r}^{(t)}, \xi\right\rangle\). The following inequality holds according to the update rules of the $\mathbf{V}$  vector in Lemma \ref{lem_update_rules_V}, the first equality is derived from  \eqref{eq:formula12}, and the second equality is due to the initialization of the $\mathbf{V}$ vector. 
\begin{align}
\left\lVert W_{V}^{(t)}\upsilon\right\rVert_2 &\leq \left\lVert W_{V}^{(0)}\upsilon\right\rVert_2 + \sum_{t' = 0}^{t - 1}\left\lVert W_{V}^{(t'+1)}\upsilon - W_{V}^{(t')}\upsilon\right\rVert_2 \notag\\
&= \left\lVert W_{V}^{(0)}\upsilon\right\rVert_2 + t O\left(\eta\cdot\max\left\{\left\lVert \mu \right\rVert_2,\sigma_p\sqrt{d}\right\}\cdot\left\lVert \upsilon\right\rVert^2\right) \notag\\
&= O\left(\sigma\nu\left\lVert \upsilon\right\rVert_2\sqrt{d}+t \eta\left\lVert \upsilon\right\rVert^2\max\left\{\left\lVert \mu \right\rVert_2,\sigma_p\sqrt{d}\right\}\right) \label{eq:formula30}
\end{align}
Since \(g(\xi)\) as \(V_{\xi,(-\widehat{y})}^{(t)} = \sum_{r} \left\langle \upsilon W_{-\widehat{y}, r}^{(t)}, \xi\right\rangle\), and since $\langle W_{-\widehat{y}, r}^{(t)}\upsilon , \xi \rangle \sim \mathcal{N}(0, \|W_{-\widehat{y}, r}^{(t)}\|_2^2 \| \upsilon \|_2^2 \sigma_p^2)$, in the benign overfitting stage, from \eqref{eq:formula30} the strength of the signal is higher than that of the noise, so we can get:
\begin{align}
\mathbb{E}g(\xi)&=\sum_{r = 1}^{m_v}\mathbb{E}\langle\upsilon W_{-\widehat{y},r}^{(t)},\xi\rangle \notag\\
&=\sum_{r = 1}^{m_v}\frac{\|W_{-\widehat{y},r}^{(t)}\|_2\sigma_p\upsilon}{\sqrt{2\pi}}\notag\\
&=\frac{\sigma_p}{\sqrt{2\pi}}\sum_{r = 1}^{m_v}\|W_{-\widehat{y},r}^{(t)}\upsilon\|_2 \label{eq:formula31}\\
&\leq O\left(\sigma_V\sigma_p\left\lVert\upsilon\right\rVert_2\sqrt{\frac{d}{2\pi}}+t\eta\sigma_p\left\lVert\upsilon\right\rVert^2\frac{\left\lVert\mu\right\rVert}{\sqrt{2\pi}}\right)\notag
\end{align}
The first  inequalities are derived from \ref{lem_The_Test_Loss_in_Benign_Overfitting}, and we drop the constant $o (1)$ to simplify the analysis. The first equality is due to Theorem \ref{thm_1} and \eqref{eq:formula31}. The last inequality is due to the fact that $(s - t)^2\geq\frac{s^2}{2}-t^2$ for all $s,t\geq0$.
\begin{align}
P(\widehat{y}(f(\theta,\mathbf{X},\upsilon)) \leq 0) &\leqslant P\left(V_{\xi,ir,(-\vec{y})}^{(t)} \geq \sum_{r} \frac{S_{11}+S_{21}}{S_{12}+S_{22}}\left(V_{+r,\vec{y}}^{(t)} + V_{-r,\vec{y}}^{(t)}\right)\right)\notag\\
&= P\left(g(\xi) - E\{g(\xi)\} \geq \sum_{r} \frac{S_{11}+S_{21}}{S_{12}+S_{22}}\left(V_{+r,\vec{y}}^{(t)} + V_{-r,\vec{y}}^{(t)}\right) - \frac{\sigma_p}{\sqrt{2\pi}} \sum_{r = 1}^{m_v} \left\|W_{-r\vec{y},r}^{(t)} \upsilon\right\|_2\right)\notag\\
&\leqslant \exp\left[-\frac{c_2\left(\sum_{r} \frac{S_{11}+S_{21}}{S_{12}+S_{22}}\left(V_{+r,\vec{y}}^{(t)} + V_{-r,\vec{y}}^{(t)}\right) - \frac{\sigma_p \sum_{r = 1}^{m_v} \left\|W_{-\vec{y},r}^{(t)} \upsilon\right\|_2}{\sqrt{2\pi}}\right)^2}{\sigma_p^2\left(\sum_{r = 1}^{m_v} \left\|W_{-\vec{y},r}^{(t)} \upsilon\right\|_2\right)^2}\right]\notag\\
&= \exp\left[-c_3\left(\frac{\sum_{r} \frac{S_{11}+S_{21}}{S_{12}+S_{22}}\left(V_{+r,\vec{y}}^{(t)} + V_{-r,\vec{y}}^{(t)}\right)}{\sigma_p \sum_{r = 1}^{m_v} \left\|W_{-\vec{y},r}^{(t)} \upsilon\right\|_2} - \frac{1}{\sqrt{2\pi}}\right)^2\right]\notag\\
&\leqslant \exp(c_4 / 2\pi) \cdot \exp\left[-\frac{c_5}{2}\left(\frac{\sum_{r} \frac{S_{11}+S_{21}}{S_{12}+S_{22}}\left(V_{+r,\vec{y}}^{(t)} + V_{-r,\vec{y}}^{(t)}\right)}{\sigma_p \sum_{r =1}^{m_v} \left\|W_{-\vec{y},r}^{(t)} \upsilon\right\|_2}\right)^2\right] \label{eq_formula32}
\end{align}
\text{When benign overfitting occurs, we know the bounds of the following terms in the second stage: }
\begin{itemize}
\item $1 - Softmax(\langle\boldsymbol{q}_{+}^{(t)},\boldsymbol{k}_{+}^{(t)}\rangle)$
\item $1 - Softmax(\langle\boldsymbol{q}_{\xi,i}^{(t)},\boldsymbol{k}_{+}^{(t)}\rangle)$
\item $Softmax(\langle\boldsymbol{q}_{+}^{(t)},\boldsymbol{k}_{\xi,i}^{(t)}\rangle)$
\item $Softmax(\langle\boldsymbol{q}_{\xi,i}^{(t)},\boldsymbol{k}_{\xi,i'}^{(t)}\rangle)$
\end{itemize}
It is noted that the following inequalities hold,so we only need to calculate the last two terms:
\begin{align*}
1 - Softmax(\langle\boldsymbol{q}_{+}^{(t)},\boldsymbol{k}_{+}^{(t)}\rangle)&=\sum_{j}Softmax(\langle\boldsymbol{q}_{+}^{(t)},\boldsymbol{k}_{\xi,i}^{(t)}\rangle)\\
1 - Softmax(\langle\boldsymbol{q}_{\xi,i}^{(t)},\boldsymbol{k}_{+}^{(t)}\rangle)&=\sum_{j}Softmax(\langle\boldsymbol{q}_{\xi,i}^{(t)},\boldsymbol{k}_{\xi,i'}^{(t)}\rangle)
\end{align*}
\text{By plugging \eqref{eq:formula11} into \eqref{eq:formula7}, and plugging \eqref{eq:formula12} into \eqref{eq:formula8}, we can get the following inequality holds:}
\begin{align}
&\frac{\exp(\langle q_+^{(t)},k_{\xi,i}^{(t)}\rangle)}{\exp(\langle q_+^{(t)},k_+^{(t)}\rangle)+\exp(\langle q_+^{(t)},k_{\xi,i}^{(t)}\rangle)}\notag\\
&\leq \frac{1}{c_5\exp\left(\Lambda_{\xi,\pm,i}^{(t)}\right)}\notag\\
&\leq \frac{1}{c_5\exp\left(\Lambda_{\xi,\pm,i}^{(T_1)}\right)+\frac{\eta^2c_8\lVert\boldsymbol{\mu}\rVert_2^4\lVert\upsilon\rVert_2^2d_h^{\frac{1}{2}}}{N(\log(24N^2/\delta))^2}\cdot(t - T_1)(t - T_1 - 1)} \notag\\
&\leq \frac{1}{c_6+\frac{\eta^2c_{13}\lVert\boldsymbol{\mu}\rVert_2^4\lVert\upsilon\rVert_2^2d_h^{\frac{1}{2}}}{N(\log(24N^2/\delta))^2}\cdot(t - T_1)(t - T_1 - 1)}.\label{eq:formula33}
\end{align}
\begin{align}
&\frac{\exp(\langle q_{\xi,i}^{(t)},k_{\xi,i'}^{(t)}\rangle)}{\exp(\langle q_{\xi,i}^{(t)},k_+^{(t)}\rangle)+\exp(\langle q_{\xi,i}^{(t)},k_{\xi,i'}^{(t)}\rangle)}\notag\\
&\leq \frac{1}{c_7\exp(\Lambda_{\xi,i,\pm,i'}^{(t)})} \notag\\
&\leq \frac{1}{c_8+\frac{\eta^2C_{13}\sigma_p^2d\lVert\boldsymbol{\mu}\rVert_2^2\lVert\upsilon\rVert_2^2d_h^{\frac{1}{2}}}{N(\log(24N^2/\delta))^2}\cdot(t - T_1)(t - T_1 - 1)}.\label{eq:formula34}
\end{align}
\text{By adding  \eqref{eq:formula33} and  \eqref{eq:formula34}, we obtain the following result:}
\begin{align}
\frac{S_{11}+S_{21}}{S_{12}+S_{22}}&=\frac{1 - S_{22}}{S_{22}}\notag\\
&=-1+\frac{1}{S_{12}}\notag\\
&\geq\frac{\eta^2c_9\left\lVert \mu \right\rVert^4\left\lVert \upsilon \right\rVert^2d_h^{\frac{1}{2}}}{N(\log(24N^2/\delta))^2}(t - T_1)(t - T_1 - 1) \label{eq:formula35}
\end{align}

\begin{align}
\frac{\sum_{r} \frac{S_{11}+S_{21}}{S_{12}+S_{22}}\left(V_{+r,\hat{y}}^{(t)} + V_{-r,\hat{y}}^{(t)}\right)}{\sigma_p \sum_{r = 1}^{m_v} \left\|W_{-\hat{y},r}^{(t)}\upsilon\right\|_2}&\geq \frac{\frac{\eta^2 c_9\left\lVert \mu \right\rVert^4\left\lVert \upsilon \right\rVert^2 d^{\frac{1}{2}}}{N(\log(24N^2/\delta))^2}(t - T_1)(t - T_1 - 1)(V_{+r\hat{y}}^{(t)} + V_{-r\hat{y}}^{(t)})}{O(\sigma_p\sigma_v\left\lVert \upsilon \right\rVert\sqrt{d} + (t - T_1)\eta\sigma_p\left\lVert \upsilon \right\rVert^2\left\lVert \mu \right\rVert)}\notag \\
&\geq\frac{\frac{\eta^3 c_9\left\lVert \mu \right\rVert^6\left\lVert \upsilon \right\rVert^4 d^{\frac{1}{2}}(t - T_1)^2(t - T_1 - 1)}{N(\log(24N^2/\delta))^2} - O(d^{\frac{-1}{4}})\cdot\frac{\eta^2 c\left\lVert \mu \right\rVert^4\left\lVert \upsilon \right\rVert^2 d^{\frac{1}{2}}(t - T_1)(t - T_1 - 1)}{N(\log(24N^2/\delta))^2}}{O(\sigma_p\sigma_v\left\lVert \upsilon \right\rVert\sqrt{d} + (t - T_1)\eta\sigma_p\left\lVert \upsilon \right\rVert^2\left\lVert \mu \right\rVert)} \notag\\
&\geq\frac{\eta^3 c_9\left\lVert \mu \right\rVert^6\left\lVert \upsilon \right\rVert^4 d^{\frac{1}{2}}(t - T_1)^2(t - T_1 - 1)}{N(\log(24N^2/\delta))^2\cdot O(\sigma_p\sigma_v\left\lVert \upsilon \right\rVert\sqrt{d} + (t - T_1)\eta\sigma_p\left\lVert \upsilon \right\rVert^2\left\lVert \mu \right\rVert)}\notag \\
&\approx\frac{\eta^3 c_9\left\lVert \mu \right\rVert^6\left\lVert \upsilon \right\rVert^4 d^{\frac{1}{2}}(t - T_1)^2(t - T_1 - 1)}{N(\log(24N^2/\delta))^2\cdot O(\eta\sigma_p\ (t - T_1)\left\lVert \upsilon \right\rVert^2\left\lVert \mu \right\rVert)} \notag\\
&\approx O\left(\frac{\eta^2 c_9\left\lVert \mu \right\rVert^5\left\lVert \upsilon \right\rVert^2 d^{\frac{1}{2}}(t - T_1)(t - T_1 - 1)}{N(\log(24N^2/\delta))^2\cdot \sigma_p}\right)\notag \\
&\approx O\left(\frac{\eta^2 c_9\left\lVert \mu \right\rVert^5\left\lVert \upsilon \right\rVert^2(t - T_1)(t - T_1 - 1)}{(\log(24N^2/\delta))^2\cdot \sigma_p}\right)\label{eq:formula36}
\end{align}
In the above formulas, we substitute \eqref{eq:formula9}, \eqref{eq:formula10}, \eqref{eq:formula35} and \eqref{eq:formula31} to obtain the first and second inequalities. By applying the idea of scaling in the last few steps, then we plug \eqref{eq:formula36} into \eqref{eq_formula32}.

\begin{align*}
P(\hat{y}f(\theta,\mathbf{X},\upsilon) \leq 0) &\leq \exp(c_4 / 2\pi) \exp\left[-\frac{c_5}{2}\left(\frac{\sum_{r} \frac{S_{11}+S_{21}}{S_{12}+S_{22}}\left(V_{+r,\hat{y}}^{(t)} + V_{-r,\hat{y}}^{(t)}\right)}{\sigma_p \sum_{r = 1}^{m_v} \left\|W_{-\hat{y},r}^{(t)} \upsilon\right\|}\right)^2\right]\\
&\leq \exp(c_4/2\pi)\exp\left[-\frac{c_5}{2}O\left(\frac{\eta^4c_9\left\lVert \mu \right\rVert^{10}\left\lVert \upsilon \right\rVert^4(t - T_1)^2(t - T_1 - 1)^2}{\sigma_p^2(\log(24N^2/\delta))^4}\right)\right] \\
&\leq \exp\left[\frac{c_4}{2\pi}-c_{10}\frac{\eta^4\left\lVert \mu \right\rVert^8\left\lVert \upsilon \right\rVert^4(t - T_1)^2(t - T_1 - 1)^2}{(\log(24N^2/\delta))^4}\cdot \text{SNR}^2\right] \\
&\leq \exp\left[\frac{c_4}{2\pi}-c_{10}\eta^4\left\lVert \mu \right\rVert^8(t - T_1)^2(t - T_1 - 1)^2\cdot \text{SNR}^2\right]
\end{align*}

\subsection{Stage III Test Loss}
\begin{theorem}[\textbf{(Third part of Theorem \ref{thm:4.1})}]
Under the same conditions as Theorem \ref{thm:4.1}, there exists a large  valued constant $C$ such that $T_3 = \Theta\left(n^{-1}\epsilon^{-1}\|\mu\|_2^{-2}\|\upsilon\|_2^{-2}\right)$. For $t\in(T_2,T_3]$, when $N\cdot\text{SNR}^2=\Omega(1)$, the upper  bound of the test error approaches the noise rate $\alpha$.
\begin{equation*}
L_{\mathcal{D}}(\mathbf{W}^{(t)}) \leq \alpha+\exp\left(\frac{c_{12}}{2\pi}-\frac{c_{14}\eta^4(t - T_2)^4\|\mu\|_2^6\cdot \text{SNR}^2}{2\sigma_v^2}\right)
\end{equation*}
\end{theorem}
\textit{Proof.}
we can get the following inequality from Lemma \ref{lem_The_Test_Loss_in_Benign_Overfitting}.
\begin{align}
&P(\widehat{y}f(\theta,\mathbf{X},\upsilon))\leq 0)\leq P\left(\sum_{r}\frac{S_{11}+S_{21}}{S_{12}+S_{22}}\left(V_{+r,\widehat{y}}^{(t)}+V_{-r,\widehat{y}}^{(t)}\right)\leq V_{\xi,ir,(-\widehat{y})}^{(t)}+o(1)\right)\label{eq:formula37}
\end{align}
In the following formulas, in the first inequality, because the signal memory in the benign overfitting stage is much higher than the noise memory, we have $S_{11}+S_{21} > S_{12}+S_{22}$. Then, by substituting \eqref{eq:formula16}, \eqref{eq:formula17}, \eqref{eq:formula18} and \eqref{eq:formula31}, we obtain the second and third inequalities. For the last inequality, we use the Taylor formula expansion of $\log(1 + x)$. When $x > 1$, the $x^2$  dominates the expansion.
\begin{align}
\frac{\sum_{r} \frac{S_{11}+S_{21}}{S_{12}+S_{22}}\left(V_{+r,\hat{y}}^{(t)} + V_{-,\hat{y}}^{(t)}\right)}{\sigma_p \sum_{r = 1}^{m_v} \left\|W_{-\hat{y},r}^{(t)} \upsilon\right\|_2} &\geqslant \frac{\sum_{r} \left(V_{+r,\hat{y}}^{(t)} + V_{-rr,\hat{y}}^{(t)}\right)}{O(\sigma_p \sigma_v \sqrt{d} \left\|\upsilon\right\|_2 + \frac{\sigma_p}{\varepsilon \left\|\mu\right\|_2})}\notag\\
&\geqslant \frac{O[\log(\exp(V_{+}^{(T_2)}) + \eta c_{10} \cdot \left\|\mu\right\|_2^2 \left\|\upsilon\right\|_2^2 (t - T_2)) - 2\log (O(1/\varepsilon))]}{O(\sigma_p \sigma_v \sqrt{d} \left\|\upsilon\right\|_2 + \frac{\sigma_p}{\varepsilon \left\|\mu\right\|_2})}\notag\ \\
&\geqslant \frac{O(\log(c_{11} + 1 + \eta c_{10} \cdot \left\|\mu\right\|^2 \left\|\upsilon\right\|^2 (t - T_2))}{O(\sigma_p \sigma_v \sqrt{d} \left\|\upsilon\right\|_2 + \frac{\sigma_p}{\varepsilon \left\|\mu\right\|_2})}\notag\\
&\geqslant \frac{O[\eta c_{10} \left\|\mu\right\|^2 \left\|\upsilon\right\|^2 (t - T_2) - \frac{1}{2} \eta^2 (c_1)^2 \left\|\mu\right\|^4 \left\|\upsilon\right\|^4 (t - T_2)^2]}{O(\sigma_p \sigma_v \sqrt{d} \left\|\upsilon\right\|_2 + \frac{\sigma_p}{\varepsilon \left\|\mu\right\|_2})}\notag\\
&\approx \frac{O(\eta^2 \left\|\mu\right\|^4 \left\|\upsilon\right\|^4 (t - T_2)^2)}{O(\sigma_p \sigma_v \sqrt{d} \left\|\upsilon\right\|_2)}\notag\\
&\approx O\left(\frac{\eta^2 \left\|\upsilon\right\|^3 (t - T_2)^2 \left\|\mu\right\|^3}{\sigma_v} \cdot \text{SNR}\right)\label{eq:formula38}
\end{align}
By plugging \eqref{eq:formula38} into \eqref{eq:formula37}, we can get :
\begin{align*}
P(\widehat{y}(f(\theta,\mathbf{X},\upsilon) \leq 0) &\leq \exp(c_{12}/ 2\pi) \exp\left[-\frac{c_{13}}{2}\left(\frac{\sum_{r} \frac{s_{11}+s_{21}}{s_{12}+s_{22}}\left(V_{+r,\hat{y}}^{(t)} + V_{-r,\hat{y}}^{(t)}\right)}{\sigma_p \sum_{r = 1}^{m} \left\|W_{-\hat{y},r}^{(t)} \upsilon\right\|}\right)^2\right]\\
&\leq \exp(c_{12} / 2\pi) \exp\left[-\frac{c_{13}}{2} O\left(\frac{\eta^4 \left\|\upsilon\right\|^6 (t - T_2)^4 \left\|\mu\right\|^6 \cdot \rm{SNR}^2}{\sigma_v^2}\right)\right]\\
&\leq\exp\left(\frac{c_{12}}{2\pi}-\frac{c_{14}\eta^{4}(t - T_{2})^{4}\left\|\mu\right\|^{6}\cdot \rm{SNR}^2}{2\sigma_{v}^{2}}\right)
\end{align*}

\section{Test Loss of Harmful Overfitting}
\subsection{Stage I Test Loss}
\begin{theorem}[\textbf{(First part of Theorem \ref{thm:4.2})}]
Under the same conditions as Theorem \ref{thm:4.2}, when $N^{-1} \cdot \text{SNR}^{-2} = \Omega(1)$, there exists \(T_1 = \frac{56N}{\eta {\frac{1}{4}}\sigma_p^2d\lVert\upsilon\rVert_2^2}\), for \(t\in(0,T_1]\), such that the test error is:
\begin{equation}
L_{\mathcal{D}}(\mathbf{W}^{(t)}) \leq \alpha + O(1)\\
\end{equation}
\end{theorem}
\textit{Proof.}The proof is the same as in the first stage of benign overfitting (Theorem \ref{thm:4.1.1}).
\subsection{Stage II Test Loss}
\begin{theorem}[\textbf{(Second part of Theorem \ref{thm:4.2}}]
 Under the same conditions as Theorem \ref{thm:4.2}, when $N^{-1} \cdot \text{SNR}^{-2} = \Omega(1)$, there exists \(T_2 = \Theta\left(\frac{N}{\eta\sigma_p^2d\lVert\upsilon\rVert_2^2\log(6N^2M^2/\delta)}\right)\). For \(t\in(T_1,T_2]\), the test error is:
\[
L_D(\mathbf{W}^{(t)}) \leq \left(\frac{1}{2} - \alpha\right) O\left(\frac{d^{\frac{1}{4}}(\log N^{2})^{3}(1 + \eta(t - T_{1}))}{m_{v}\|\mu\|^{2}\|\upsilon\|^{2}}+\frac{\eta d\|\upsilon\|^{2}(t - T_{1})}{N}\right)
\]
\end{theorem}
\textit{Proof.}
\text{we bound $f(\theta, \mathbf{X}, \upsilon)$ as follows.}
\begin{align*}
f(\theta, \mathbf{X}, \upsilon) &= \frac{1}{m_v} \sum_{r \in [m_v]} \left( \upsilon^T \mathbf{x}_{1} (S_{11} + S_{21}) \mathbf{W}_{Vj,r}+ \upsilon^T \mathbf{x}_{2} (S_{12}+S_{22}) \mathbf{W}_{Vj,r} \right)\\
&\approx \frac{1}{m_v} \cdot \frac{\exp(\langle\boldsymbol{q}_+^{(t)},\boldsymbol{k}_+^{(t)}\rangle)}{\exp(\langle\boldsymbol{q}_+^{(t)},\boldsymbol{k}_+^{(t)}\rangle)+\exp(\langle\boldsymbol{q}_+^{(t)},\boldsymbol{k}_{\xi,i}^{(t)}\rangle)}\cdot V_+^{(t)}\\
&+ \frac{1}{m_v} \cdot \frac{\exp(\langle\boldsymbol{q}_{\xi,i}^{(t)},\boldsymbol{k}_+^{(t)}\rangle)}{\exp(\langle\boldsymbol{q}_+^{(t)},\boldsymbol{k}_+^{(t)}\rangle)+\exp(\langle\boldsymbol{q}_{\xi,i}^{(t)},\boldsymbol{k}_{\xi,i'}^{(t)}\rangle)} \cdot V_+^{(t)}\\
&+ \frac{1}{m_v} \cdot \frac{\exp(\langle\boldsymbol{q}_-^{(t)},\boldsymbol{k}_-^{(t)}\rangle)}{\exp(\langle\boldsymbol{q}_-^{(t)},\boldsymbol{k}_-^{(t)}\rangle)+\exp(\langle\boldsymbol{q}_-^{(t)},\boldsymbol{k}_{\xi,i}^{(t)}\rangle)}\cdot V_-^{(t)}\\
&+ \frac{1}{m_v} \cdot \frac{\exp(\langle\boldsymbol{q}_{\xi,i}^{(t)},\boldsymbol{k}_+^{(t)}\rangle)}{\exp(\langle\boldsymbol{q}_+^{(t)},\boldsymbol{k}_+^{(t)}\rangle)+\exp(\langle\boldsymbol{q}_{\xi,i}^{(t)},\boldsymbol{k}_{\xi,i'}^{(t)}\rangle)} \cdot V_-^{(t)}\\
&+ \frac{1}{m_v} \cdot \frac{\exp(\langle\boldsymbol{q}_+^{(t)},\boldsymbol{k}_{\xi,i}^{(t)}\rangle)}{\exp(\langle\boldsymbol{q}_+^{(t)},\boldsymbol{k}_+^{(t)}\rangle)+\exp(\langle\boldsymbol{q}_+^{(t)},\boldsymbol{k}_{\xi,i'}^{(t)}\rangle)}\cdot V_{\xi,i}^{(t)}\\
&+ \frac{1}{m_v} \cdot \frac{\exp(\langle\boldsymbol{q}_{\xi,i}^{(t)},\boldsymbol{k}_{\xi,i'}^{(t)}\rangle)}{\exp(\langle\boldsymbol{q}_{\xi,i}^{(t)},\boldsymbol{k}_+^{(t)}\rangle)+\exp(\langle\boldsymbol{q}_{\xi,i}^{(t)},\boldsymbol{k}_{\xi,i'}^{(t)}\rangle)} \cdot V_{\xi,i}^{(t)}
\end{align*}
To simplify the analysis, let's define:
\begin{align*}
A&=\frac{1}{m_v} \cdot \frac{\exp(\langle\boldsymbol{q}_+^{(t)},\boldsymbol{k}_+^{(t)}\rangle)}{\exp(\langle\boldsymbol{q}_+^{(t)},\boldsymbol{k}_+^{(t)}\rangle)+\exp(\langle\boldsymbol{q}_+^{(t)},\boldsymbol{k}_{\xi,i}^{(t)}\rangle)}\cdot V_+^{(t)}\\
B&=\frac{1}{m_v} \cdot \frac{\exp(\langle\boldsymbol{q}_+^{(t)},\boldsymbol{k}_+^{(t)}\rangle)}{\exp(\langle\boldsymbol{q}_+^{(t)},\boldsymbol{k}_+^{(t)}\rangle)+\exp(\langle\boldsymbol{q}_+^{(t)},\boldsymbol{k}_{\xi,i}^{(t)}\rangle)}\cdot V_+^{(t)}\\
C&=\frac{1}{m_v} \cdot \frac{\exp(\langle\boldsymbol{q}_-^{(t)},\boldsymbol{k}_-^{(t)}\rangle)}{\exp(\langle\boldsymbol{q}_-^{(t)},\boldsymbol{k}_-^{(t)}\rangle)+\exp(\langle\boldsymbol{q}_-^{(t)},\boldsymbol{k}_{\xi,i}^{(t)}\rangle)}\cdot V_-^{(t)}\\
D&=\frac{1}{m_v} \cdot \frac{\exp(\langle\boldsymbol{q}_{\xi,i}^{(t)},\boldsymbol{k}_+^{(t)}\rangle)}{\exp(\langle\boldsymbol{q}_+^{(t)},\boldsymbol{k}_+^{(t)}\rangle)+\exp(\langle\boldsymbol{q}_{\xi,i}^{(t)},\boldsymbol{k}_{\xi,i'}^{(t)}\rangle)} \cdot V_-^{(t)}\\
E&=\frac{1}{m_v} \cdot \frac{\exp(\langle\boldsymbol{q}_+^{(t)},\boldsymbol{k}_{\xi,i}^{(t)}\rangle)}{\exp(\langle\boldsymbol{q}_+^{(t)},\boldsymbol{k}_+^{(t)}\rangle)+\exp(\langle\boldsymbol{q}_+^{(t)},\boldsymbol{k}_{\xi,i'}^{(t)}\rangle)}\cdot V_{\xi,i}^{(t)}\\
F&=\frac{1}{m_v} \cdot \frac{\exp(\langle\boldsymbol{q}_{\xi,i}^{(t)},\boldsymbol{k}_{\xi,i'}^{(t)}\rangle)}{\exp(\langle\boldsymbol{q}_{\xi,i}^{(t)},\boldsymbol{k}_+^{(t)}\rangle)+\exp(\langle\boldsymbol{q}_{\xi,i}^{(t)},\boldsymbol{k}_{\xi,i'}^{(t)}\rangle)} \cdot V_{\xi,i}^{(t)}\\
&f(\theta, \mathbf{x}, \upsilon) =A + B + C + D+E + F
\end{align*}
\begin{align*}
|f(\theta, \mathbf{x}, \upsilon) |&\leq|A|+|B|+|C|+|D|+|E|+|F|
\end{align*}
By plugging \eqref{eq:formula20}--\eqref{eq:formula26} into the above definition, the following inequality holds:
\begin{align*}
|A|&\leq\frac{1}{m_v}\left|O\left(\frac{\sigma_p^2 d (\log(24N^2 / \delta))^3}{\| \mu \|_2^2 \| v \|_2^2 d^{\frac{1}{2}}}\right)\right|\left(O(d^{-\frac{1}{4}}) + \frac{\eta c_{15} \sigma_p^2 d \| v \|_2^2 (t - T_1)}{N}\right)\\
&=\frac{1}{m_v}\left(O\left(\frac{\sigma_p^2 d^{\frac{1}{4}} (\log(24N^2 / \delta))^3}{\| \mu \|_2^2 \| v \|_2^2}\right)+O\left(\frac{\eta c_{15} \sigma_p^4 d^{\frac{3}{2}} (\log(24N^2 / \delta))^3(t - T_1)}{\| \mu \|_2^2 N \| v \|_2^2}\right)\right)\\
|B|&\leq\frac{1}{m_v}\left(1 - O\left(\frac{\sigma_p^2 d (\log(24N^2 / \delta))^3}{\| \mu \|_2^2 \| v \|_2^2 d^{\frac{1}{2}}}\right)\right)\left(O(d^{-\frac{1}{4}}) + \frac{\eta c_{15} \sigma_p^2 d \| v \|_2^2 (t - T_1)}{N}\right)\\
&\approx\frac{1}{m_v}\left(O(d^{-\frac{1}{4}}) + \frac{\eta c_{15} \sigma_p^2 d \| v \|_2^2 (t - T_1)}{N}\right)\\
|C|&\leq\frac{1}{m_v}\left(O\left(\frac{\sigma_p^2 d^{\frac{1}{4}} (\log(24N^2 / \delta))^3}{\| \mu \|_2^2 \| v \|_2^2}\right)+O\left(\frac{\eta c_{15} \sigma_p^4 d^{\frac{3}{2}} (\log(24N^2 / \delta))^3(t - T_1)}{\| \mu \|_2^2 N \| v \|_2^2}\right)\right)\\
|D|&\leq\frac{1}{m_v}\left(O(d^{-\frac{1}{4}}) + \frac{\eta c_{15} \sigma_p^2 d \| v \|_2^2 (t - T_1)}{N}\right)\\
|E|&\leq\frac{1}{m_v}\left|O\left(\frac{(\log(24N^2 / \delta))^3}{\| v \|_2^2 d^{\frac{1}{2}}}\right)\right|\frac{\eta c_{14} \sigma_p^2 d \| v \|_2^2 (t - T_1)}{N}\\
&=\frac{1}{m_v}O\left(\frac{\eta c_{14} \sigma_p^2 d^{\frac{1}{2}}(\log(24N^2 / \delta))^3(t - T_1)}{N}\right)\\
|F|&\leq\frac{1}{m_v}\left(1 - O\left(\frac{(\log(24N^2 / \delta))^3}{\| v \|_2^2 d^{\frac{1}{2}}}\right)\right)\frac{\eta c_{14} \sigma_p^2 d \| v \|_2^2 (t - T_1)}{N}\\
&\approx\frac{1}{m_v}\frac{\eta c_{14} \sigma_p^2 d \| v \|_2^2 (t - T_1)}{N}
\end{align*}
Calculate the sum of the absolute values presented above,  we can get:
\begin{align}
f(\theta, \mathbf{X}, \upsilon) &\leq\frac{1}{m_v}\left(2O\left(\frac{\sigma_p^2 d^{\frac{1}{4}} (\log(24N^2 / \delta))^3}{\| \mu \|_2^2 \| v \|_2^2}\right)+2O\left(\frac{\eta c_{15} \sigma_p^4 d^{\frac{3}{2}} (\log(24N^2 / \delta))^3(t - T_1)}{\| \mu \|_2^2 N \| v \|_2^2}\right)+2O(d^{-\frac{1}{4}})\right.\notag\\
&\left.+2\frac{\eta c_{15} \sigma_p^2 d \| v \|_2^2 (t - T_1)}{N}+O\left(\frac{\eta c_{14} \sigma_p^2 d^{\frac{1}{2}}(\log(24N^2 / \delta))^3(t - T_1)}{N}\right)+\frac{\eta c_{14} \sigma_p^2 d \| v \|_2^2 (t - T_1)}{N}\right)\notag\\
&\leq  O\left(\frac{d^{\frac{1}{4}}(\log(N^2/\delta))^3}{m_v\|\mu\|_2^2\|v\|_2^2}\right)+O\left(\eta d^{\frac{1}{4}}(\log(N^2/\delta))^3(t - T_1)\left(\frac{1}{m_v\|\mu\|_2^2N\|v\|_2^2}+\frac{1}{N}\right)\right)+O\left(\frac{\eta d\|v\|_2^2(t - T_1)}{N}\right)\label{eq.formula40}
\end{align}
According the full expectation formula, the test loss is split into two parts: label not flipped and label flipped respectively, and the sum is obtained as follows:
\begin{align}
L_{\mathcal{D}}(\mathbf{W}^{(t)})&=\mathbb{E}\ell[y\cdot f(\theta, \mathbf{X}, \upsilon)]\notag\\
&=(1 - \alpha)E\ell(yf(\theta, \mathbf{X}, \upsilon))+\alpha E\ell(-yf(\theta, \mathbf{X}, \upsilon))\label{eq.formula41}\\
E[y]&=p(y = 1)\times1 + \alpha(y = -1)\times(-1)\notag\\
&=[p(\hat{y}=1)\times(1 - \alpha)+p(\hat{y}=-1)\times \alpha]\times1+[p(\hat{y}=1)\times \alpha + p(\hat{y}=-1)\times(1 - \alpha)]\times(-1)\notag\\
&=[p_0\times(1 - \alpha)+p_0\times \alpha]\times1+[p_0\times \alpha + p_0\times(1 - \alpha)]\times(-1)\notag\\
&=\alpha - \alpha p_0 + p_0 - \alpha p_0\notag\\
&=\alpha + p_0 - 2\alpha p_0\label{eq.formula42}
\end{align}
The above formula $E[y]$ is derived from the full expectation formula, and $p_0$ is the probability that the real label $\widehat{y}$ outputs 1 when it is not flipped. 
By plugging \eqref{eq.formula40} into \eqref{eq.formula41}, we can get:
\begin{align}
L_D(\mathbf{W}^{(t)})&=(1 - 2\alpha)\mathbb{E}[yf(\theta,\mathbf{X},v)]\notag\\
&\leq(1 - 2\alpha)\mathbb{E}\left[y\cdot\left(O\left(\frac{d^{\frac{1}{2}}(\log(N^2/\delta))^3}{m_v\|\mu\|_2^2\|v\|_2^2}\right)\right.\right.\notag\\
&\quad\left.\left.+O\left(\eta d^{\frac{1}{2}}(\log(N^2/\delta))^3(t - T_1)\left(\frac{1}{m_v\|\mu\|_2^2N\|v\|_2^2}+\frac{1}{N}\right)\right)+O\left(\frac{\eta d\|\ v\|_2^2(t - T_1)}{N}\right)\right)\right]\notag\\
&=(1 - 2\alpha)\left(\mathbb{E}\left[y\cdot O\left(\frac{d^{\frac{1}{2}}(\log(N^2/\delta))^3}{m_v\|\mu\|_2^2\|v\|_2^2}\right)\right]\right.\notag\\
&\quad\left.+\mathbb{E}\left[y\cdot O\left(\eta d^{\frac{1}{2}}(\log(N^2/\delta))^3(t - T_1)\left(\frac{1}{m_v\|\mu\|_2^2N\|v\|_2^2}+\frac{1}{N}\right)\right)\right]+\mathbb{E}\left[y\cdot O\left(\frac{\eta d\|\ v\|_2^2(t - T_1)}{N}\right)\right]\right)\notag\\
&=(1 - 2\alpha)\left(O\left(\frac{d^{\frac{1}{2}}(\log(N^2/\delta))^3}{m_v\|\mu\|_2^2\|v\|_2^2}\right)\mathbb{E}[y]\right.\notag\\
&\quad\left.+O\left(\eta d^{\frac{1}{2}}(\log(N^2/\delta))^3(t - T_1)\left(\frac{1}{m_v\|\mu\|_2^2N\|v\|_2^2}+\frac{1}{N}\right)\right)\mathbb{E}[y]+O\left(\frac{\eta d\|\ v\|_2^2(t - T_1)}{N}\right)\mathbb{E}[y]\right)\label{eq:formula43}
\end{align}
Substitute \(p_0 = \frac{1}{2}\) into \(\mathbb{E}[y]\),  we can get
\begin{align}
L_D(\mathbf{W}^{(t)})&=(1 - 2\alpha)\left(\frac{1}{2}O\left(\frac{d^{\frac{1}{2}}(\log(N^2/\delta))^3}{m_v\|\mu\|_2^2\|v\|_2^2}\right)\right.\notag\\
&\quad\left.+\frac{1}{2}O\left(\eta d^{\frac{1}{2}}(\log(N^2/\delta))^3(t - T_1)\left(\frac{1}{m_v\|\mu\|_2^2N\|v\|_2^2}+\frac{1}{N}\right)\right)+\frac{1}{2}O\left(\frac{\eta d\|\ v\|_2^2(t - T_1)}{N}\right)\right)\notag\\
&=(\frac{1}{2} - \alpha)\left(O\left(\frac{d^{\frac{1}{2}}(\log(N^2/\delta))^3}{m_v\|\mu\|_2^2\|v\|_2^2}\right)\right.\notag\\
&\quad\left.+O\left(\eta d^{\frac{1}{2}}(\log(N^2/\delta))^3(t - T_1)\left(\frac{1}{m_v\|\mu\|_2^2N\|v\|_2^2}+\frac{1}{N}\right)\right)+O\left(\frac{\eta d\|\ v\|_2^2(t - T_1)}{N}\right)\right)\label{eq:formula44}
\end{align}

\subsection{Stage III Test Loss}
\begin{theorem}[\textbf{(Third part of Theorem \ref{thm:4.2}}]
Under the same conditions as Theorem \ref{thm:4.2}, when $N^{-1} \cdot \text{SNR}^{-2} = \Omega(1)$, here exists \(T_3 = \Theta\left(\frac{N}{\eta\epsilon\sigma_p^2d\lVert\upsilon\rVert_2^2}\right)\), \(t\in(T_2,T_3]\) such that: 
\[
L_{\mathcal{D}}(\mathbf{W}^{(t)}) \geq (1 - \alpha)\cdot\log\left(\frac{1 + 2e^{\frac{1}{2}}}{1 + e^{\frac{1}{2}}}\right)+\alpha\log\left(2 + e^{-\frac{1}{2}}\right)
\]
\end{theorem}
\textit{Proof.}
According to \eqref{eq:formula27} and \eqref{eq:formula29}, the test error is divided into two parts from \eqref{eq.formula41}.
Firstly, let us analyze the first case when the label is not flipped.
Let \(z = \widehat{y}f(\theta, \mathbf{X}, \upsilon)\). Since \(t \in (T_2, T_3]\), we have \(\widehat{y}f(\theta, \mathbf{x}, \upsilon) \geq \log(1 + e^{-\frac{1}{2}})\) and \(\ell(z)=\log(1+\exp(-z))\). Then we can get
\begin{align*}
(1 - \alpha)\mathbb{E}_{y\in\widehat{y}}\ell(\widehat{y}f(\theta, \mathbf{X}, \upsilon)) &\leq (1 - \alpha)\ell(\widehat{y}f(\theta, \mathbf{x}, \upsilon)) \\
&=(1 - \alpha)\log\left(1+\exp(-\widehat{y}f(\theta, \mathbf{X}, \upsilon))\right) \\
&\leq (1 - \alpha)\log\left(1+\frac{e^{\frac{1}{2}}}{1 + e^{\frac{1}{2}}}\right)\\
&=(1 - \alpha)\log\left(\frac{1 + 2e^{\frac{1}{2}}}{1 + e^{\frac{1}{2}}}\right)
\end{align*}
Secondly, let's analyze the second case. When the label is flipped:
\begin{align*}
\ell(-\widehat{y}f(\theta, \mathbf{X}, \upsilon)) &= \log(1 + \exp(\widehat{y}f(\theta, \mathbf{X}, \upsilon)))\\
&\geq \log\left(1 + (1 + e^{-1/2})\right)\\
&= \log\left(2 + e^{-1/2}\right)\\
\alpha\mathbb{E}_{y\in - \widehat{y}}\ell(-\widehat{y}f(\theta, \mathbf{X}, \upsilon))&\geq \alpha\log\left(2 + e^{-1/2}\right)
\end{align*}
The proof can be completed by adding the results of the first and second cases.
\section{More Experiment Details on Training Dynamics}
\label{a}
\begin{figure}[t]
    \centering
    \par\vspace{0.5em}
    \subfigure[Dynamic loss analysis of  \( N \).]{
       \includegraphics[width=0.46\linewidth]{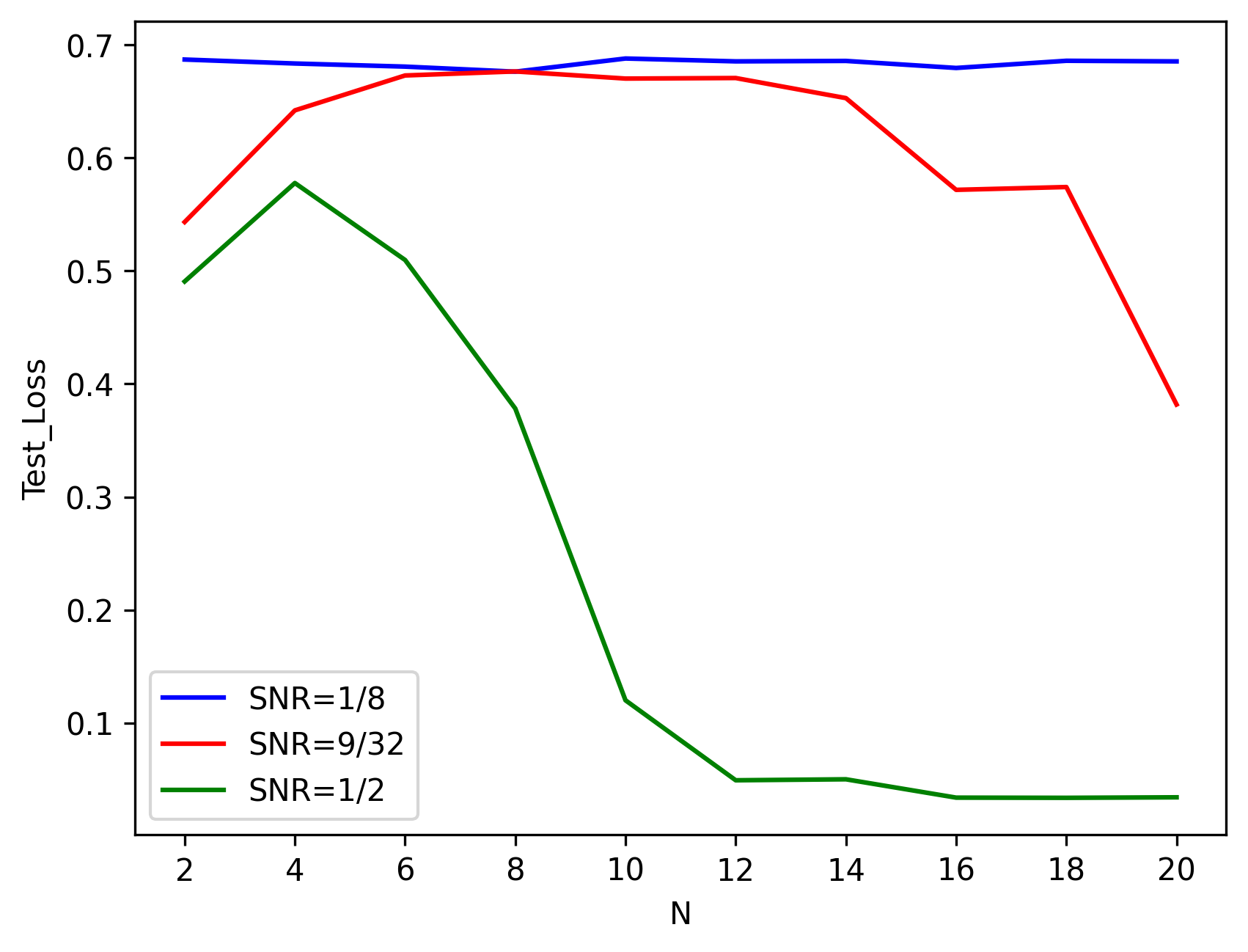}
    }
    \hfill
    \subfigure[Dynamic loss analysis of $\alpha$.]{
       \includegraphics[width=0.46\linewidth]{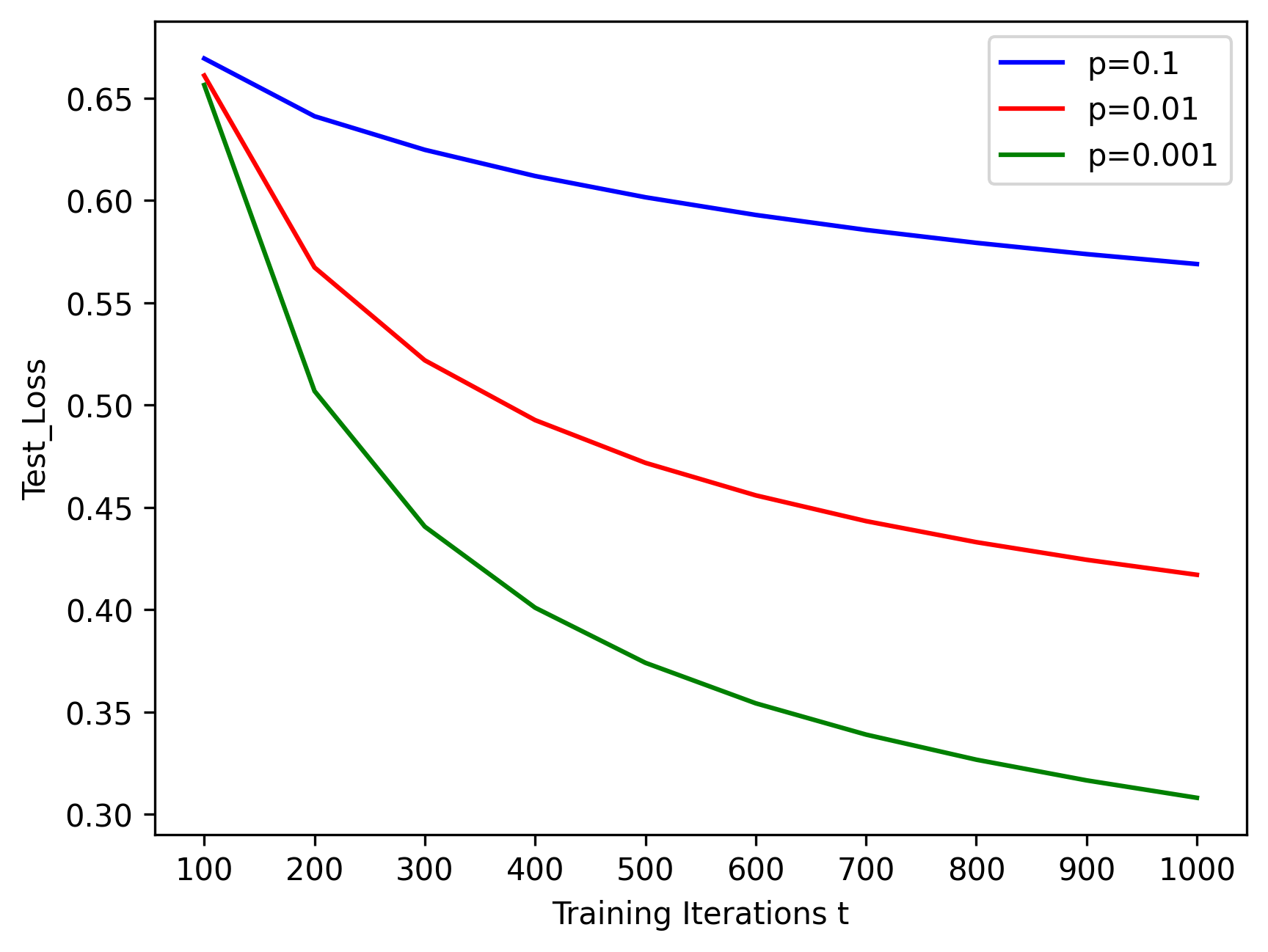}
    }
    \par\vspace{1em}
    \subfigure[Dynamic loss analysis of $\mu$.]{
       \includegraphics[width=0.46\linewidth]{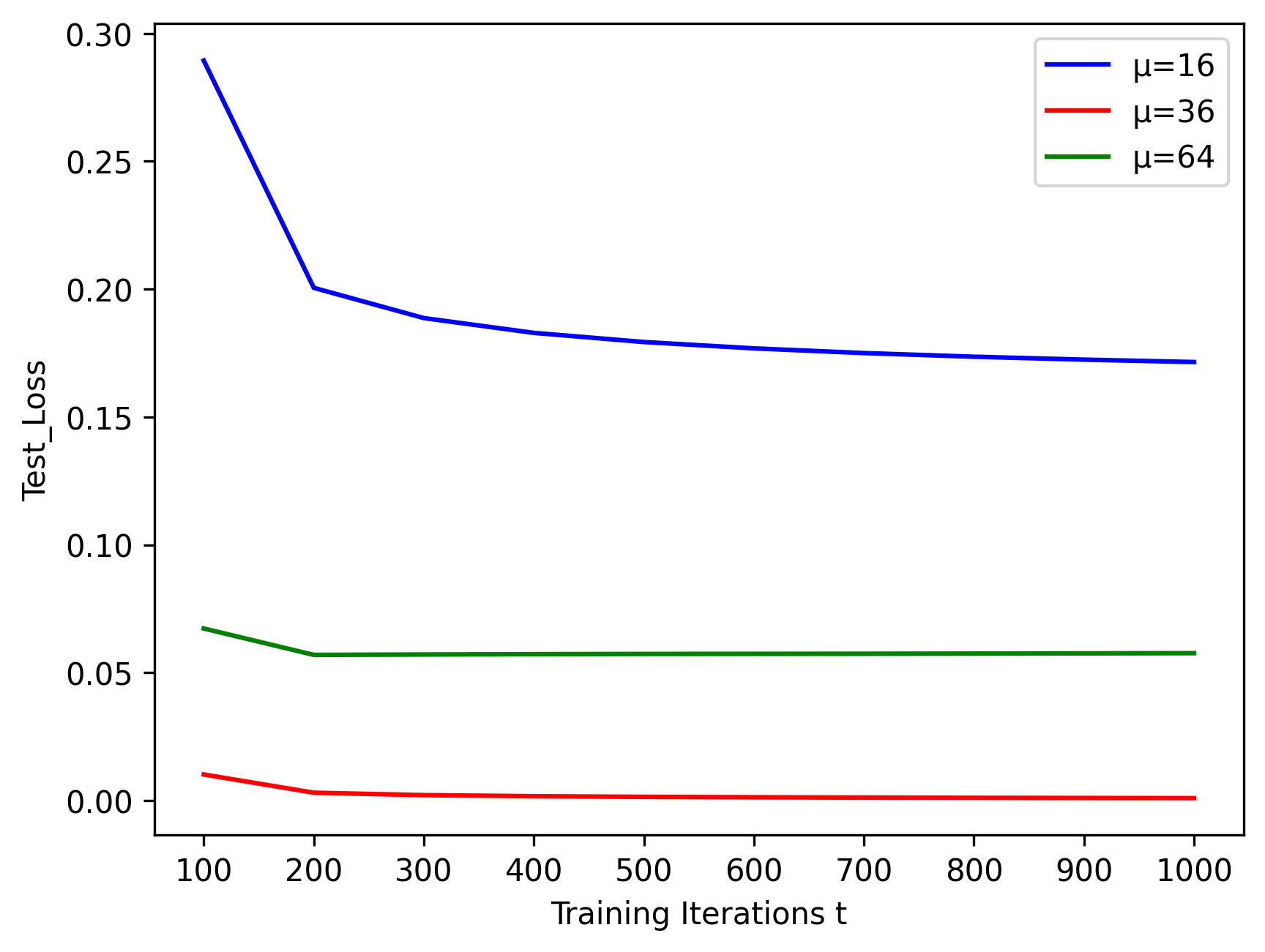}
    }
    \hfill
    \subfigure[Dynamic loss analysis of $\sigma$.]{
       \includegraphics[width=0.46\linewidth]{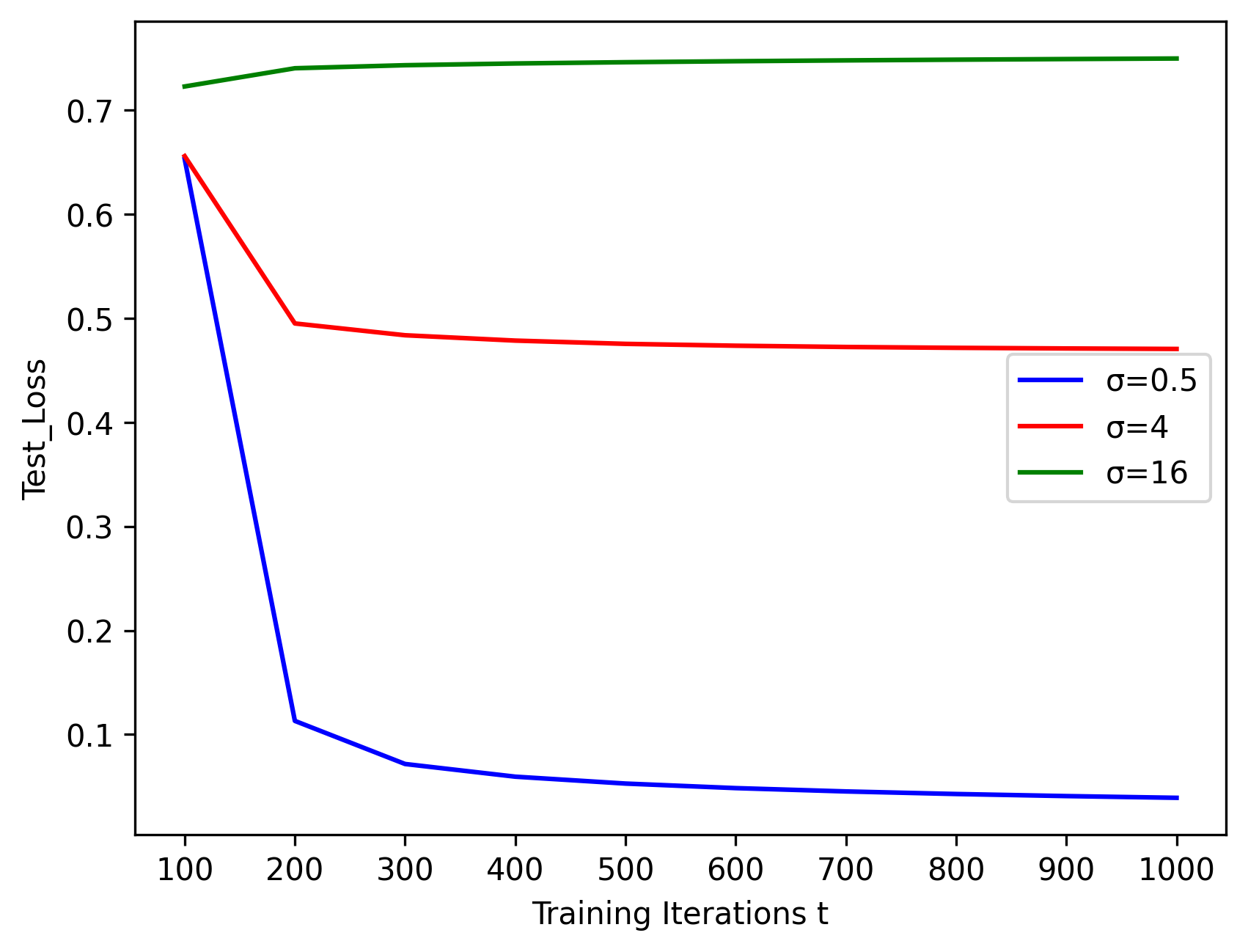}
    }
    \caption{
    This figure is a study of individual variables in \ref{heatmap} and \ref{heatmap}, where (a) explores the impact of changes in \( \text{N} \) and \( \text{SNR} \) on test loss, (b) investigates the impact of $\alpha$. on test loss, and (c) and (d) are components of \( \text{SNR} \). We further investigate the values of \(\mu\) and \( \sigma \) to analyze the changes in loss
    }
    \label{SNR_N_P}
\end{figure}

\begin{enumerate}
\item Sample size N: \cref{SNR_N_P} (a) shows that the overall test loss decreases with the change of N. However, from the \( \text{SNR} \)  curves representing different values, it can be observed that the number of changes in \( \text{N} \) has little effect on the loss under lower signal-to-noise ratio conditions, and a significant decrease only occurs when the signal-to-noise ratio is increased. This indicates that there is not a certain relationship between the influence of n and \( \text{SNR} \). As we obtained in the theoretical part \( N \cdot \text{SNR}^2 = \Omega(1) \)
\item Label flipping $\alpha$ : Label inversion means changing the label of the original signal, and in binary classification tasks, it means taking the opposite number of labels. We tried different probabilities in the experiment, and the specific results are shown in the \cref{SNR_N_P} (b). 
\item signal-to-noise ratio ( \( \text{SNR} \) ): The previous experiment proved that as \( \text{SNR} \) increases, the loss will show a decreasing trend. Here, we further explore the factors that affect \( \text{SNR} \) and their impact on the loss. \( \text{SNR} \) is determined by both the signal norm and the noise variance. We conducted experiments on $\mu$ and $\sigma$ while keeping other variables unchanged.The experimental results are shown in the figure. \cref{SNR_N_P}(c), it can be seen that the test loss increases with the increase of signal strength, while in \ref{SNR_N_P}(d), it can be seen that the test loss decreases with the increase of noise variance. This result is exactly the same as our definition of \( \text{SNR} \).
\end{enumerate}

\end{document}